%% file: main-arxiv.tex
\pdfoutput=1
\documentclass[11pt]{article}
\usepackage[final]{acl}
\usepackage{times}
\usepackage{latexsym}
\usepackage{booktabs,longtable,multirow,makecell,array,xcolor,colortbl,pifont}
\newcommand{\cmark}{\ding{51}}\newcommand{\xmark}{\ding{55}}
\usepackage[T1]{fontenc}
\usepackage[utf8]{inputenc}
\usepackage{microtype}
\input{preamble-arxiv}

\usepackage{subfigure}
\usepackage{nicefrac}
\usepackage[most]{tcolorbox}
\usepackage{fontawesome5}
\usepackage{caption}
\definecolor{GoogleBlue}{RGB}{66,133,244}
\definecolor{GoogleRed}{RGB}{234,67,53}
\definecolor{GoogleYellow}{RGB}{251,188,5}
\definecolor{GoogleGreen}{RGB}{52,168,83}
\definecolor{GoogleBlueLight}{RGB}{232,240,254}
\definecolor{GoogleRedLight}{RGB}{252,232,230}
\definecolor{GoogleYellowLight}{RGB}{254,247,224}
\definecolor{GoogleGreenLight}{RGB}{230,244,234}
\definecolor{GoogleGray}{RGB}{95,99,104}
\definecolor{GoogleGrayLight}{RGB}{241,243,244}
\usepackage{colortbl}
\usepackage{enumitem}
\setlist[itemize]{noitemsep, topsep=0pt, leftmargin=*}
\newlist{compactitem}{itemize}{1}
\setlist[compactitem]{noitemsep, topsep=0pt}
\usepackage{tcolorbox}
\newtcolorbox{promptbox}[2][]{
  enhanced,
  colback=white!98!blue!2,
  colframe=blue!70!black,
  coltitle=white,
  fonttitle=\bfseries\sffamily,
  title={\faTasks[regular]\hspace{1mm}~#2},
  left=2mm, right=2mm, top=1mm, bottom=1mm,
  arc=3mm,
  attach boxed title to top left={xshift=3mm, yshift*=-2mm},
  boxed title style={
    colback=blue!80!black,
    size=small,
    sharp corners=south,
    bottom=0.5mm, top=0.5mm,
    left=1mm, right=1mm,
    fontupper=\bfseries\sffamily,
    boxrule=0pt,
    drop shadow
  },
  boxrule=0.9pt,
  drop shadow southeast,
}
\usepackage{bold-extra}
\usepackage[T1]{fontenc}
\usepackage{array}
\newcolumntype{P}[1]{>{\centering\arraybackslash}p{#1}}
\newcolumntype{M}[1]{>{\centering\arraybackslash}m{#1}}
\definecolor{orange}{rgb}{1,0.5,0}
\definecolor{graynode}{RGB}{20,20,20}
\definecolor{crimsonred}{RGB}{220,20,60}
\definecolor{darkgraynode}{gray}{0.5}
\definecolor{lightgraynode}{gray}{0.8}
\usepackage{rotate}
\usepackage{adjustbox}
\usepackage{rotating}
\usepackage{array}
\usepackage{capt-of}
\usepackage{tabulary}
\usepackage{setspace}
\usepackage{amssymb}
\usepackage{mathtools}
\usepackage{pifont}

\definecolor{gray}{RGB}{20,20,20}
\definecolor{gray}{RGB}{0.7,0.7,0.7}
\definecolor{greencm}{RGB}{0,153,0}
\newcommand{\cm}{ {\color{greencm}\normalsize\cmark}}
\newcommand{\cmgray}{ {\color{gray}\normalsize\cmark}}
\newcommand{\xm}{ {\color{red}\normalsize\xmark}}

\definecolor{plotblue}{RGB}	{30,144,255}
\definecolor{plotgreen}{RGB}	{50,205,50}
\definecolor{plotred}{RGB}	{220,20,60}
\definecolor{myyellow}{RGB}{255,255,204}
\definecolor{myred}{RGB}{255,204,204}
\definecolor{myblue}{RGB}{0,200,255}
\definecolor{mygreen}{RGB}{80,220,80}

\newcommand*\hrulefillvar[1][0.4pt]{\leavevmode\leaders\hrule height#1\hfill\kern0pt}
\usepackage{algorithm}
\usepackage[noend]{algpseudocode}
\algrenewcommand\algorithmicrequire{\textbf{Input:}}
\algrenewcommand\algorithmicensure{\textbf{Output:}}
\DeclareMathAlphabet{\mathbcal}{OMS}{cmsy}{b}{n}
\usepackage{amsmath}
\usepackage{mathrsfs}
\usepackage{comment}

\usepackage{bm}
\usepackage{bbm}
\usepackage{amssymb}
\usepackage{enumitem}

\usepackage{amsthm}
\theoremstyle{definition}
\newtheorem{definition}{Definition}
\newtheorem{problem}{Problem}

\title{Joint and Cross-Modal Video-Audio Generation and Editing: A Unified Formulation and Design Taxonomy}

\author{%
  \parbox{0.94\textwidth}{\centering
  Abhinav~Sharma$^{1}$,
  Sai~Karthik~Navuluru$^{2}$,
  Wang~Wei$^{3}$,
  Daksh~Dangi$^{1}$,
  Xiangbo~Gao$^{4}$,
  Li~Li$^{5}$,
  Bo~Ni$^{6}$,
  Vardhan~Dongre$^{7}$,
  Junda~Wu$^{8}$,
  Xiyang~Hu$^{9}$,
  Jiuxiang~Gu$^{8}$,
  Seunghyun~Yoon$^{8}$,
  Tong~Yu$^{8}$,
  Chien~Van~Nguyen$^{10}$,
  Mohamed~Elmoghany$^{11}$,
  Nedim~Lipka$^{8}$,
  Hoda~Eldardiry$^{3}$,
  Hongjie~Chen$^{12}$,
  Tyler~Derr$^{6}$,
  Thien~Huu~Nguyen$^{10}$,
  Zhengzhong~Tu$^{4}$,
  Nesreen~K.~Ahmed$^{13}$,
  Franck~Dernoncourt$^{8}$,
  Ryan~A.~Rossi$^{8}$
  \\[0.8ex]
  {\normalsize\mdseries $^{1}$University~of~Massachusetts~Amherst \quad $^{2}$University~of~Texas~at~Dallas \quad $^{3}$Virginia~Tech \quad $^{4}$Texas~A\&M~University \quad $^{5}$University~of~Southern~California \quad $^{6}$Vanderbilt~University \quad $^{7}$University~of~Illinois~Urbana-Champaign \quad $^{8}$Adobe~Research \quad $^{9}$Arizona~State~University \quad $^{10}$University~of~Oregon \quad $^{11}$Stanford~University \quad $^{12}$Dolby~Laboratories \quad $^{13}$Cisco}%
  }%
}
\begin{document}
\pagestyle{plain}% arXiv preprint: page numbers (acl.sty final mode sets an empty page style)
\makeatletter\setlength{\@fptop}{0pt}\setlength{\@dblfptop}{0pt}\makeatother% float pages: tables start at the top instead of vertically centred
\maketitle
\thispagestyle{plain}
\begin{abstract}
Video and audio are perceived together, yet most generative models treat them in isolation. We examine methods that model the two modalities jointly, generate one from the other, or edit them in a coupled manner, organized around a single question: how is the output kept coherent across modalities in time and semantics? A unified formulation casts joint generation, cross-modal generation, and joint editing as three problems defined on a single distribution over audio-visual pairs, and a taxonomy compares methods along five design axes. To our knowledge, this is the first overview to systematically taxonomize joint audio-visual \emph{editing}, which we map as nine edit categories spanning 28 edit types. We describe methods, datasets, and metrics for each setting and close with the open problems we view as most consequential.
\end{abstract}

\section{Introduction}
\label{sec:intro}
Sound and picture are perceived as a single percept: a door closing and its impact sound coincide, and even a small offset reads as an error. Generative models of video and audio, however, developed largely in isolation, so two strong unimodal capabilities, combined naively, produce streams that do not agree. A growing body of work instead treats the pair as coupled, in three settings grouped by \emph{output}: (i) both modalities generated together; (ii) one generated from the other, as in soundtracking a silent video; and (iii) an existing pair edited so a change in one modality propagates to the other. All three share one demand---coherence across modalities in time and semantics---which we take as our organizing principle.

\paragraph{Contributions.}
Existing overviews treat video generation, audio generation, or audio-visual understanding in isolation~\cite{qin2026avi}. This work covers generation and editing of the two streams as a coupled pair. Our contributions are summarized as follows:
\begin{itemize}
    \item \textbf{First systematic taxonomy of joint audio-visual editing.} To our knowledge, this is the first overview to systematically taxonomize the editing of video and audio as a coupled pair, in which an edit specified in one modality must propagate to the other; we map this space as a taxonomy of nine categories and 28 edit types, each with representative operations and use cases (Table~\ref{tab:audiovisual-edits-taxonomy}, Section~\ref{sec:jointedit}).
    \item \textbf{A unified formulation and five-axis design taxonomy.} Section~\ref{sec:formulation} casts joint generation, cross-modal generation, and joint editing as three problems over one distribution on audio-visual pairs, and Table~\ref{tab:av-design-taxonomy} organizes the literature along five design axes.
    \item \textbf{Open problems grounded in the formulation.} Section~\ref{sec:open} states the open problems---long-horizon coherence, fine-grained control, physical plausibility, and evaluation---as instances of one underlying challenge: raising cross-modal alignment while preserving per-stream quality.
\end{itemize}

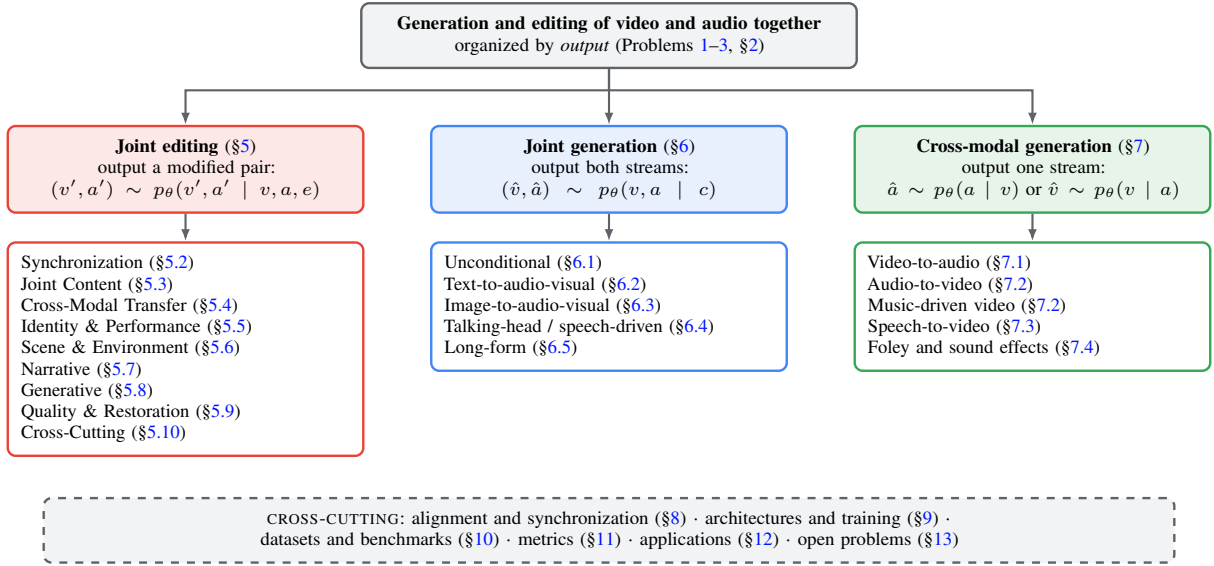
\begin{figure*}[t]
\centering
\begin{tikzpicture}[
  font=\scriptsize,
  every node/.style={line width=0.9pt},
  root/.style={draw=GoogleGray, fill=GoogleGrayLight, rounded corners=3pt, align=center, inner sep=5pt, text width=6.2cm},
  fam/.style={rounded corners=3pt, align=center, inner sep=5pt, text width=4.35cm, minimum height=1.02cm},
  lst/.style={rounded corners=3pt, fill=white, align=left, inner sep=5pt, text width=4.35cm},
  cc/.style={draw=GoogleGray, dashed, rounded corners=3pt, fill=GoogleGrayLight, align=center, inner sep=5pt, text width=14.6cm},
  edge/.style={GoogleGray, line width=0.8pt, -{Latex[length=1.6mm]}}
]
\node[root] (root) at (5.6,0) {\textbf{Generation and editing of video and audio together}\\ organized by \emph{output} (Problems~\ref{prob:jointgen}--\ref{prob:jointedit}, \S\ref{sec:formulation})};

\node[fam, draw=GoogleRed, fill=GoogleRedLight, anchor=north] (f3) at (0.0,-1.18) {\textbf{Joint editing} (\S\ref{sec:jointedit})\\ output a modified pair:\\ $(v', a') \sim p_\theta(v', a' \mid v, a, e)$};
\node[fam, draw=GoogleBlue, fill=GoogleBlueLight, anchor=north] (f1) at (5.6,-1.18) {\textbf{Joint generation} (\S\ref{sec:jointgen})\\ output both streams:\\ $(\hat v, \hat a) \sim p_\theta(v, a \mid c)$};
\node[fam, draw=GoogleGreen, fill=GoogleGreenLight, anchor=north] (f2) at (11.2,-1.18) {\textbf{Cross-modal generation} (\S\ref{sec:crossgen})\\ output one stream:\\ $\hat a \sim p_\theta(a \mid v)$ or $\hat v \sim p_\theta(v \mid a)$};

\node[lst, draw=GoogleRed, anchor=north] (l3) at (0.0,-2.72) {Synchronization (\S\ref{sec:jointedit-sync})\\ Joint Content (\S\ref{sec:jointedit-content})\\ Cross-Modal Transfer (\S\ref{sec:jointedit-transfer})\\ Identity \& Performance (\S\ref{sec:jointedit-identity})\\ Scene \& Environment (\S\ref{sec:jointedit-scene})\\ Narrative (\S\ref{sec:jointedit-narrative})\\ Generative (\S\ref{sec:jointedit-generative})\\ Quality \& Restoration (\S\ref{sec:jointedit-restoration})\\ Cross-Cutting (\S\ref{sec:jointedit-crosscutting})};
\node[lst, draw=GoogleBlue, anchor=north] (l1) at (5.6,-2.72) {Unconditional (\S\ref{sec:jointgen-unconditional})\\ Text-to-audio-visual (\S\ref{sec:jointgen-text})\\ Image-to-audio-visual (\S\ref{sec:jointgen-image})\\ Talking-head / speech-driven (\S\ref{sec:jointgen-talkinghead})\\ Long-form (\S\ref{sec:jointgen-long})};
\node[lst, draw=GoogleGreen, anchor=north] (l2) at (11.2,-2.72) {Video-to-audio (\S\ref{sec:crossgen-v2a})\\ Audio-to-video (\S\ref{sec:crossgen-a2v})\\ Music-driven video (\S\ref{sec:crossgen-a2v})\\ Speech-to-video (\S\ref{sec:crossgen-s2v})\\ Foley and sound effects (\S\ref{sec:crossgen-foley})};

\node[cc, anchor=north] (cc) at (5.6,-6.1) {\textsc{cross-cutting:}\ alignment and synchronization (\S\ref{sec:align}) $\cdot$ architectures and training (\S\ref{sec:arch}) $\cdot$ datasets and benchmarks (\S\ref{sec:datasets}) $\cdot$ metrics (\S\ref{sec:eval}) $\cdot$ applications (\S\ref{sec:applications}) $\cdot$ open problems (\S\ref{sec:open})};

\draw[edge] (root.south) -- ++(0,-0.28) -| (f3.north);
\draw[edge] (root.south) -- (f1.north);
\draw[edge] (root.south) -- ++(0,-0.28) -| (f2.north);
\draw[edge] (f3.south) -- (l3.north);
\draw[edge] (f1.south) -- (l1.north);
\draw[edge] (f2.south) -- (l2.north);
\end{tikzpicture}
\caption{\textbf{Roadmap of this work.} Methods are grouped by the \emph{output} they produce (Section~\ref{sec:formulation}): joint editing outputs a modified pair, joint generation outputs both streams, and cross-modal generation exactly one---the signals a method consumes vary freely within each family. The listed subsections cover each family; the gray strip collects the cross-cutting sections that apply to all three. Color marks the family (red editing, blue joint, green cross-modal) and is redundant with position and headings. Notation follows Table~\ref{tab:notation}.}
\label{fig:roadmap}
\end{figure*}

\paragraph{Scope.}
A method is in scope when at least one of video or audio is among its outputs and the other appears in its pipeline; single-modality generation and audio-visual understanding without a generative or editing component are excluded. The closest overlapping overview, a concurrent review of audio-visual intelligence in foundation models~\cite{qin2026avi}, treats neither editing nor our organizing device. Figure~\ref{fig:roadmap} maps the organization.

\def\rotateDeg{90}
\definecolor{googlegreen}{HTML}{0F9D58}
\definecolor{googleblue}{HTML}{4285F4}
\definecolor{googlered}{HTML}{DB4437}
\definecolor{googlepurple}{HTML}{673AB7}
\definecolor{googleorange}{HTML}{F4B400}
\newcommand\TE{\rule{0pt}{2.0ex}}
\newcommand\BE{\rule[-1.1ex]{0pt}{0pt}}
{
\newcolumntype{C}{ >{\centering\arraybackslash} m{4cm} }
\providecommand{\rotateDeg}{90}
\setlength{\tabcolsep}{1.8pt}
\providecommand{\rotDeg}{70}
\definecolor{verylightgreennew}{RGB}	{220,255,220}
\definecolor{verylightrednew}{RGB}		{255, 230, 230}
\definecolor{verylightreddarker}{HTML} {FFCBCB}
\definecolor{verylightrednew}{RGB}		{255, 230, 230}
\definecolor{verylightrednewlighter}{RGB}		{255, 229, 239}
\definecolor{lightgraynew}{rgb}{0.95,0.95,0.95}
\definecolor{newgray}{RGB}{0.3,0.3,0.3}
\providecommand{\cellsz}{0.30cm}
\providecommand{\cellszlg}{0.30cm}
\providecommand{\cellszsm}{0.30cm}
\renewcommand{\cm}{{\color{greencm}\normalsize\cmark}}
\renewcommand{\cmgray}{{\color{lightgraynew}\normalsize\cmark}}
\renewcommand{\xm}{{\color{verylightreddarker}\normalsize\xmark}}
\newcommand\BBBBB{\rule[1.6ex]{0pt}{1.6ex}}
\newcommand\BBBBBBB{\rule[-1.5ex]{0.0pt}{2.5ex}}
\newcommand\BBBnew{\rule[-2.5ex]{0pt}{0pt}}
\newcommand\BBBBBB{\rule[-1.1ex]{0pt}{0pt}}
\newcommand{\sysName}[1]{{
\BBBBBB
#1
}}
\providecommand{\cellsomewhat}{
\BBBBB
\cmgray
\cellcolor{verylightgreennew}
}
\providecommand{\cellno}{
\BBBBB
\xm
\cellcolor{verylightrednew}}
\providecommand{\cellyes}{
\BBBBB
\cm
\cellcolor{verylightgreennew}
}
\begin{table*}[t!]
\centering
\renewcommand{\arraystretch}{1.12}
\scriptsize
\caption{
\textbf{Complementary taxonomy of joint audio-video methods along five complementary design axes:}
\textcolor{googlegreen}{\sc generation strategy} (\S\ref{sec:strategy}) captures how the two modalities are produced;
\textcolor{googleblue}{\sc audio representation} (\S\ref{sec:audio-repr}) describes the latent space in which audio is generated;
\textcolor{googleorange}{\sc video representation} (\S\ref{sec:video-repr}) describes the latent space in which video is generated;
\textcolor{googlered}{\sc alignment enforcement} (\S\ref{sec:sync}) identifies the stage at which audio-video alignment is imposed;
and \textcolor{googlepurple}{\sc pretraining reuse} (\S\ref{sec:pretrain}) characterizes the source of model weights.
A check mark (\cmark) indicates that a method falls under the corresponding category within each axis.
For the partially closed system Wan~2.5, the axes whose design is not publicly disclosed (alignment enforcement and pretraining reuse) are left without a mark.
}
\label{tab:av-design-taxonomy}
\begin{tabular}
{P{2mm}
l
c@{\hspace{8pt}}
P{\cellszlg} P{\cellszlg} P{\cellszlg} P{\cellszlg} P{\cellszlg}
@{\hspace{8pt}}
P{\cellszlg} P{\cellszlg} P{\cellszlg} P{\cellszlg}
@{\hspace{8pt}}
P{\cellszlg} P{\cellszlg} P{\cellszlg} P{\cellszlg}
@{\hspace{8pt}}
P{\cellszlg} P{\cellszlg} P{\cellszlg} P{\cellszlg} P{\cellszlg}
@{\hspace{8pt}}
P{\cellszsm} P{\cellszsm} P{\cellszsm}
@{}
}
\toprule
& &
& \multicolumn{5}{c}{\textcolor{googlegreen}{\textsc{\bfseries \shortstack{Generation\\Strategy}}}}
& \multicolumn{4}{c}{\textcolor{googleblue}{\textsc{\bfseries \shortstack{Audio\\Representation}}}}
& \multicolumn{4}{c}{\textcolor{googleorange}{\textsc{\bfseries \shortstack{Video\\Representation}}}}
& \multicolumn{5}{c}{\textcolor{googlered}{\textsc{\bfseries \shortstack{Alignment\\Enforcement}}}}
& \multicolumn{3}{c}{\textcolor{googlepurple}{\textsc{\bfseries \shortstack{Pretraining\\Reuse}}}}
\\

& &
\BBBnew
& \multicolumn{5}{c}{(\textbf{Section~\ref{sec:strategy}})}
\BBBnew
& \multicolumn{4}{c}{(\textbf{Section~\ref{sec:audio-repr}})}
& \multicolumn{4}{c}{(\textbf{Section~\ref{sec:video-repr}})}
& \multicolumn{5}{c}{(\textbf{Section~\ref{sec:sync}})}
& \multicolumn{3}{c}{(\textbf{Section~\ref{sec:pretrain}})}
\\
\cmidrule(lr){4-8}\cmidrule(lr){9-12}\cmidrule(lr){13-16}\cmidrule(lr){17-21}\cmidrule(lr){22-24}

& &
&
\rotatebox{\rotateDeg}{\textbf{Single-Tower (\S\ref{sec:strategy-single})}} &
\rotatebox{\rotateDeg}{\textbf{Dual-Tower (\S\ref{sec:strategy-dual})}} &
\rotatebox{\rotateDeg}{\textbf{Cascaded (\S\ref{sec:strategy-cascade})}} &
\rotatebox{\rotateDeg}{\textbf{Unified-Token (\S\ref{sec:strategy-token})}} &
\rotatebox{\rotateDeg}{\textbf{Guidance-Based (\S\ref{sec:strategy-guidance})}} &
\rotatebox{\rotateDeg}{\textbf{Waveform (\S\ref{sec:audio-wav})}} &
\rotatebox{\rotateDeg}{\textbf{Mel-Spectrogram (\S\ref{sec:audio-mel})}} &
\rotatebox{\rotateDeg}{\textbf{Continuous Latent (\S\ref{sec:audio-latent})}} &
\rotatebox{\rotateDeg}{\textbf{Discrete Tokens (\S\ref{sec:audio-tokens})}} &
\rotatebox{\rotateDeg}{\textbf{Pixel (\S\ref{sec:video-pixel})}} &
\rotatebox{\rotateDeg}{\textbf{2D-VAE + Temporal (\S\ref{sec:video-2d})}} &
\rotatebox{\rotateDeg}{\textbf{3D-VAE (\S\ref{sec:video-3d})}} &
\rotatebox{\rotateDeg}{\textbf{Discrete Tokens (\S\ref{sec:video-tokens})}} &
\rotatebox{\rotateDeg}{\textbf{Cross-Attention (\S\ref{sec:sync-xattn})}} &
\rotatebox{\rotateDeg}{\textbf{Shared Pos. Enc. (\S\ref{sec:sync-pe})}} &
\rotatebox{\rotateDeg}{\textbf{Discriminator (\S\ref{sec:sync-disc})}} &
\rotatebox{\rotateDeg}{\textbf{Classifier Guidance (\S\ref{sec:sync-guide})}} &
\rotatebox{\rotateDeg}{\textbf{Explicit Prior (\S\ref{sec:sync-prior})}} &
\rotatebox{\rotateDeg}{\textbf{From-Scratch (\S\ref{sec:pretrain-scratch})}} &
\rotatebox{\rotateDeg}{\textbf{Single Pretrained (\S\ref{sec:pretrain-single})}} &
\rotatebox{\rotateDeg}{\textbf{Dual Pretrained (\S\ref{sec:pretrain-dual})}}
\\
\midrule

\rowcolor{lightgraynew}
\multicolumn{24}{l}{\BBBBB\BBBBBB\BBBBBBB \sc \bfseries \hspace{-2mm}  Joint Audio-Video Generation}
\\
\midrule
& \sysName{MM-Diffusion}~\cite{ruan2023mmdiffusion}
& & \cellno & \cellyes & \cellno & \cellno & \cellno
& \cellno & \cellyes & \cellno & \cellno
& \cellyes & \cellno & \cellno & \cellno
& \cellyes & \cellno & \cellno & \cellno & \cellno
& \cellyes & \cellno & \cellno
\\
& \sysName{CoDi}~\cite{tang2023codi}
& & \cellno & \cellyes & \cellno & \cellno & \cellno
& \cellno & \cellno & \cellyes & \cellno
& \cellno & \cellyes & \cellno & \cellno
& \cellyes & \cellno & \cellno & \cellno & \cellno
& \cellno & \cellno & \cellyes
\\
& \sysName{Seeing-and-Hearing}~\cite{xing2024seeing}
& & \cellno & \cellno & \cellno & \cellno & \cellyes
& \cellno & \cellno & \cellyes & \cellno
& \cellno & \cellyes & \cellno & \cellno
& \cellno & \cellno & \cellno & \cellyes & \cellno
& \cellno & \cellno & \cellyes
\\
& \sysName{AV-DiT}~\cite{wang2024avdit}
& & \cellyes & \cellno & \cellno & \cellno & \cellno
& \cellno & \cellno & \cellyes & \cellno
& \cellno & \cellyes & \cellno & \cellno
& \cellno & \cellyes & \cellno & \cellno & \cellno
& \cellno & \cellyes & \cellno
\\
& \sysName{MM-LDM}~\cite{sun2024mmldm}
& & \cellyes & \cellno & \cellno & \cellno & \cellno
& \cellno & \cellno & \cellyes & \cellno
& \cellno & \cellyes & \cellno & \cellno
& \cellyes & \cellno & \cellno & \cellno & \cellno
& \cellyes & \cellno & \cellno
\\
& \sysName{Movie Gen}~\cite{polyak2024moviegen}
& & \cellno & \cellno & \cellyes & \cellno & \cellno
& \cellno & \cellno & \cellyes & \cellno
& \cellno & \cellno & \cellyes & \cellno
& \cellyes & \cellno & \cellno & \cellno & \cellno
& \cellyes & \cellno & \cellno
\\
& \sysName{SVG}~\cite{ishii2024svg}
& & \cellno & \cellyes & \cellno & \cellno & \cellno
& \cellno & \cellno & \cellyes & \cellno
& \cellno & \cellyes & \cellno & \cellno
& \cellyes & \cellno & \cellno & \cellno & \cellno
& \cellno & \cellno & \cellyes
\\
& \sysName{MMDisCo}~\cite{hayakawa2025mmdisco}
& & \cellno & \cellno & \cellno & \cellno & \cellyes
& \cellno & \cellno & \cellyes & \cellno
& \cellno & \cellyes & \cellno & \cellno
& \cellno & \cellno & \cellyes & \cellno & \cellno
& \cellno & \cellno & \cellyes
\\
& \sysName{SyncFlow}~\cite{liu2024syncflow}
& & \cellno & \cellyes & \cellno & \cellno & \cellno
& \cellno & \cellno & \cellyes & \cellno
& \cellno & \cellno & \cellyes & \cellno
& \cellyes & \cellno & \cellno & \cellno & \cellno
& \cellyes & \cellno & \cellno
\\
& \sysName{JavisDiT}~\cite{liu2025javisdit}
& & \cellno & \cellyes & \cellno & \cellno & \cellno
& \cellno & \cellno & \cellyes & \cellno
& \cellno & \cellno & \cellyes & \cellno
& \cellno & \cellno & \cellno & \cellno & \cellyes
& \cellyes & \cellno & \cellno
\\
& \sysName{JavisDiT++}~\cite{liu2026javisditpp}
& & \cellno & \cellyes & \cellno & \cellno & \cellno
& \cellno & \cellno & \cellyes & \cellno
& \cellno & \cellno & \cellyes & \cellno
& \cellno & \cellyes & \cellno & \cellno & \cellyes
& \cellyes & \cellno & \cellno
\\
& \sysName{BridgeDiT}~\cite{guan2025bridgedit}
& & \cellno & \cellyes & \cellno & \cellno & \cellno
& \cellno & \cellno & \cellyes & \cellno
& \cellno & \cellno & \cellyes & \cellno
& \cellyes & \cellno & \cellno & \cellno & \cellno
& \cellno & \cellno & \cellyes
\\
& \sysName{ALIVE}~\cite{guo2026alive}
& & \cellno & \cellyes & \cellno & \cellno & \cellno
& \cellno & \cellno & \cellyes & \cellno
& \cellno & \cellno & \cellyes & \cellno
& \cellno & \cellyes & \cellno & \cellno & \cellno
& \cellno & \cellyes & \cellno
\\
& \sysName{Ovi}~\cite{low2025ovi}
& & \cellno & \cellyes & \cellno & \cellno & \cellno
& \cellno & \cellno & \cellyes & \cellno
& \cellno & \cellno & \cellyes & \cellno
& \cellyes & \cellno & \cellno & \cellno & \cellno
& \cellyes & \cellno & \cellno
\\
& \sysName{UniAVGen}~\cite{zhang2025uniavgen}
& & \cellno & \cellyes & \cellno & \cellno & \cellno
& \cellno & \cellno & \cellyes & \cellno
& \cellno & \cellno & \cellyes & \cellno
& \cellyes & \cellno & \cellno & \cellno & \cellno
& \cellyes & \cellno & \cellno
\\
& \sysName{Animate-and-Sound}~\cite{wang2025jointdit}
& & \cellno & \cellyes & \cellno & \cellno & \cellno
& \cellno & \cellno & \cellyes & \cellno
& \cellno & \cellyes & \cellno & \cellno
& \cellyes & \cellno & \cellno & \cellno & \cellno
& \cellno & \cellno & \cellyes
\\
& \sysName{CCL}~\cite{ma2026ccl}
& & \cellno & \cellyes & \cellno & \cellno & \cellno
& \cellno & \cellno & \cellyes & \cellno
& \cellno & \cellno & \cellyes & \cellno
& \cellyes & \cellno & \cellno & \cellno & \cellno
& \cellno & \cellno & \cellyes
\\
& \sysName{Hallo-Live}~\cite{li2026hallolive}
& & \cellno & \cellyes & \cellno & \cellno & \cellno
& \cellno & \cellno & \cellyes & \cellno
& \cellno & \cellno & \cellyes & \cellno
& \cellyes & \cellno & \cellno & \cellno & \cellno
& \cellno & \cellyes & \cellno
\\
& \sysName{UniForm}~\cite{zhao2025uniform}
& & \cellyes & \cellno & \cellno & \cellno & \cellno
& \cellno & \cellno & \cellyes & \cellno
& \cellno & \cellno & \cellyes & \cellno
& \cellno & \cellyes & \cellno & \cellno & \cellno
& \cellyes & \cellno & \cellno
\\
& \sysName{Wan 2.5}~\cite{wan25_2025}
& & \cellyes & \cellno & \cellno & \cellno & \cellno
& \cellno & \cellno & \cellyes & \cellno
& \cellno & \cellno & \cellyes & \cellno
& & & & & 
& & &
\\
& \sysName{LTX-2}~\cite{hacohen2026ltx2}
& & \cellno & \cellyes & \cellno & \cellno & \cellno
& \cellno & \cellno & \cellyes & \cellno
& \cellno & \cellno & \cellyes & \cellno
& \cellyes & \cellno & \cellno & \cellno & \cellno
& \cellyes & \cellno & \cellno
\\
& \sysName{MOVA}~\cite{mova2026}
& & \cellyes & \cellno & \cellno & \cellno & \cellno
& \cellno & \cellno & \cellyes & \cellno
& \cellno & \cellno & \cellyes & \cellno
& \cellyes & \cellno & \cellno & \cellno & \cellno
& \cellyes & \cellno & \cellno
\\
& \sysName{Apollo}~\cite{wang2026klear}
& & \cellyes & \cellno & \cellno & \cellno & \cellno
& \cellno & \cellno & \cellyes & \cellno
& \cellno & \cellno & \cellyes & \cellno
& \cellno & \cellyes & \cellno & \cellno & \cellno
& \cellyes & \cellno & \cellno
\\
& \sysName{3MDiT}~\cite{li2025threemdit}
& & \cellyes & \cellno & \cellno & \cellno & \cellno
& \cellno & \cellno & \cellyes & \cellno
& \cellno & \cellno & \cellyes & \cellno
& \cellno & \cellyes & \cellno & \cellno & \cellno
& \cellno & \cellyes & \cellno
\\
& \sysName{OmniForcing}~\cite{su2026omniforcing}
& & \cellno & \cellyes & \cellno & \cellno & \cellno
& \cellno & \cellno & \cellyes & \cellno
& \cellno & \cellno & \cellyes & \cellno
& \cellyes & \cellno & \cellno & \cellno & \cellno
& \cellno & \cellyes & \cellno
\\

\midrule
\rowcolor{lightgraynew}
\multicolumn{24}{l}{\BBBBB\BBBBBB\BBBBBBB \sc \bfseries \hspace{-2mm}  Joint Audio-Video Editing}
\\
\midrule
& \sysName{Lang.-Guid. AV Edit}~\cite{liang2024avedit}
& & \cellno & \cellyes & \cellno & \cellno & \cellno
& \cellno & \cellno & \cellyes & \cellno
& \cellno & \cellyes & \cellno & \cellno
& \cellyes & \cellno & \cellno & \cellno & \cellno
& \cellno & \cellno & \cellyes
\\
& \sysName{AV-Edit}~\cite{guo2026avedit}
& & \cellyes & \cellno & \cellno & \cellno & \cellno
& \cellno & \cellno & \cellyes & \cellno
& \cellno & \cellyes & \cellno & \cellno
& \cellyes & \cellno & \cellno & \cellno & \cellno
& \cellno & \cellyes & \cellno
\\
& \sysName{JUST-DUB-IT}~\cite{chen2026justdubit}
& & \cellno & \cellyes & \cellno & \cellno & \cellno
& \cellno & \cellno & \cellyes & \cellno
& \cellno & \cellno & \cellyes & \cellno
& \cellyes & \cellno & \cellno & \cellno & \cellno
& \cellno & \cellyes & \cellno
\\
& \sysName{EditYourself}~\cite{flynn2026edityourself}
& & \cellyes & \cellno & \cellno & \cellno & \cellno
& \cellno & \cellno & \cellyes & \cellno
& \cellno & \cellno & \cellyes & \cellno
& \cellyes & \cellno & \cellno & \cellno & \cellno
& \cellno & \cellyes & \cellno
\\
\bottomrule
\end{tabular}
\vspace{-1mm}
\end{table*}
}

\section{Problem Formulation and Notation}
\label{sec:formulation}
\paragraph{Notation.}
A video is $v \in \mathbb{R}^{T_v \times H \times W \times 3}$, where $T_v$ is the number of frames and $H, W$ are the spatial dimensions. An audio signal is $a \in \mathbb{R}^{T_a}$, where $T_a$ is the number of audio samples. A condition is $c$, with subscripts for specific modalities: $c_t$ for text, $c_i$ for an image, $c_s$ for speech, and $c_m$ for music. A generative model is $p_\theta$ with parameters $\theta$. Generated outputs are written $\hat{v}, \hat{a}$, and edited outputs $v', a'$. We write $v^{(t)}$ for the $t$-th frame and use $\phi_v$ and $\phi_a$ for encoders that map video and audio into latent or shared representation spaces, with $z_v = \phi_v(v)$ and $z_a = \phi_a(a)$.
Table~\ref{tab:notation} collects the symbols used throughout this work.

\begin{table}[t!]
\centering
\scriptsize
\renewcommand{\arraystretch}{1.2}
\caption{\textbf{Notation.} We summarize the symbols used throughout this work. Subscripts denote modality and primes denote edited outputs.}
\label{tab:notation}
\setlength{\tabcolsep}{5pt}
\begin{tabular}{@{}ll@{}}
\toprule
\textbf{Symbol} & \textbf{Meaning} \\
\midrule
$v \in \mathbb{R}^{T_v \times H \times W \times 3}$ & video, $T_v$ frames of size $H \times W$ \\
$a \in \mathbb{R}^{T_a}$ & audio signal of $T_a$ samples \\
$x = (v, a)$ & audio-visual clip \\
$\mathcal{X} = \mathcal{V} \times \mathcal{A}$ & space of audio-visual clips \\
$\mathrm{fps}, \mathrm{sr}$ & frame rate, audio sampling rate \\
$\tau$ & shared clip duration \\
$c$ & condition signal \\
$c_t, c_i, c_s, c_m$ & text, image, speech, music \\
$p_\theta$ & generative model with parameters $\theta$ \\
$\theta_v, \theta_a, \theta_\times$ & modality-specific and cross-modal parameters (\S\ref{sec:strategy}) \\
$\Gamma$ & sampling procedure (\S\ref{sec:taxonomy}) \\
$\lambda$ & guidance weight (\S\ref{sec:strategy-guidance}) \\
$\hat{v}, \hat{a}$ & generated video, audio \\
$v', a'$ & edited video, audio \\
$\phi_v, \phi_a$ & video, audio encoders \\
$z_v, z_a$ & video, audio latents (Def.~\ref{def:latent}) \\
$\psi_v, \psi_a$ & video, audio decoders \\
$\mathcal{S}(v, a)$ & alignment score (Def.~\ref{def:correspondence}) \\
$\delta$ & synchronization tolerance (Def.~\ref{def:sync}) \\
$e$ & edit instruction \\
$\mathcal{D}$ & dataset (Def.~\ref{def:dataset}) \\
$v^{(t)}$ & $t$-th video frame \\
\bottomrule
\end{tabular}
\end{table}

An audio-visual clip\label{def:avclip} is a pair $x = (v, a)$ with $v \in \mathbb{R}^{T_v \times H \times W \times 3}$ and $a \in \mathbb{R}^{T_a}$ sharing a time interval $[0, \tau]$, so frame and sample index refer to the same physical time; $\mathcal{X} = \mathcal{V} \times \mathcal{A}$ is the space of such pairs.

\begin{definition}[Audio-visual correspondence]
\label{def:correspondence}
$(v, a)$ corresponds when the streams agree \emph{semantically}---the sources visible are the sources audible---and \emph{temporally}---each acoustic event is localized to the frames of its visual cause: for an alignment score $\mathcal{S}: \mathcal{X} \to \mathbb{R}$, natural clips satisfy $\mathcal{S}(v, a) \ge \mathcal{S}(v, \tilde{a})$ for mismatched or time-shifted $\tilde{a}$.
\end{definition}

Definition~\ref{def:correspondence} is what separates the joint and cross-modal setting from two independent unimodal problems. A model that produces a high-quality $\hat{v}$ and a high-quality $\hat{a}$ but assigns them low $\mathcal{S}(\hat{v}, \hat{a})$ is perceived as broken, even when each stream is convincing on its own. The methods in Sections~\ref{sec:jointedit} through~\ref{sec:crossgen} differ primarily in how they raise $\mathcal{S}$ while keeping the per-modality quality high.

\begin{definition}[Latent representations]
\label{def:latent}
Methods typically operate on $z_v = \phi_v(v)$ and $z_a = \phi_a(a)$ with decoders $\psi_v, \psi_a$; $p_\theta$ is defined over $(z_v, z_a)$, and $\phi_v, \phi_a$ fix the coupling space.
\end{definition}

\begin{problem}[Joint audio-visual generation]
\label{prob:jointgen}
Given $c$ (possibly $\emptyset$), learn $p_\theta(v, a \mid c)$ whose samples are faithful to $c$, of high per-modality quality, and of high $\mathcal{S}(\hat{v}, \hat{a})$.
\end{problem}

The defining requirement in Problem~\ref{prob:jointgen} is that the parameterization of $p_\theta(v, a \mid c)$ encode the dependency between $v$ and $a$. A model that factorizes as $p_\theta(v \mid c)\, p_\theta(a \mid c)$ with no further coupling treats the two modalities as conditionally independent given $c$ and cannot, in general, raise $\mathcal{S}$ beyond what $c$ already determines. Joint methods therefore introduce coupling either in the architecture, through shared parameters or cross-modal attention, or in the objective, through a term that rewards correspondence.

\begin{problem}[Cross-modal generation]
\label{prob:crossgen}
Given one modality, generate the other so that the pair corresponds: video-to-audio learns $p_\theta(a \mid v)$, audio-to-video learns $p_\theta(v \mid a)$.
\end{problem}

Problem~\ref{prob:crossgen} differs from Problem~\ref{prob:jointgen} in what is given. In the joint setting both modalities are outputs of a single distribution; in the cross-modal setting one of them is the condition. The contrast is concrete: a joint text-to-audio-visual model given $c_t = $ ``footsteps on gravel'' synthesizes both the visual scene and the footstep sounds, whereas a cross-modal video-to-audio model given a silent video $v$ of a person walking on gravel synthesizes only $\hat{a}$, with the footstep sounds aligned to the frames in which the foot contacts the ground.

\begin{problem}[Joint audio-visual editing]
\label{prob:jointedit}
Given $(v, a)$ and an instruction $e$ (text, mask, style, or identity), sample $(v', a') \sim p_\theta(v', a' \mid v, a, e)$ such that the edit is applied, reflected in both modalities so $\mathcal{S}(v', a')$ stays high, and content outside the targeted region is preserved.
\end{problem}

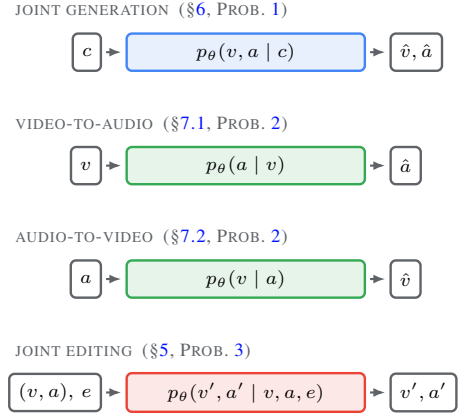
\begin{figure}[t]
\centering
\begin{tikzpicture}[
  font=\scriptsize,
  every node/.style={line width=0.9pt},
  sig/.style={draw=GoogleGray, fill=white, rounded corners=2pt, inner sep=3.5pt, minimum height=14pt},
  mdl/.style={rounded corners=2pt, inner sep=3.5pt, minimum height=14pt, minimum width=3.15cm, align=center},
  rowlab/.style={anchor=west, font=\scriptsize\scshape, text=GoogleGray, inner sep=1pt},
  arr/.style={GoogleGray, line width=0.8pt, -{Latex[length=1.6mm]}, shorten >=1.5pt, shorten <=1.5pt}
]
\node[rowlab] at (0,0.55) {joint generation\ \ (\S\ref{sec:jointgen}, Prob.~\ref{prob:jointgen})};
\node[sig, anchor=east] (i1) at (1.2,0) {$c$};
\node[mdl, draw=GoogleBlue, fill=GoogleBlueLight] (m1) at (3.1,0) {$p_\theta(v, a \mid c)$};
\node[sig, anchor=west] (o1) at (5.0,0) {$\hat v, \hat a$};
\draw[arr] (i1) -- (m1); \draw[arr] (m1) -- (o1);

\node[rowlab] at (0,-0.95) {video-to-audio\ \ (\S\ref{sec:crossgen-v2a}, Prob.~\ref{prob:crossgen})};
\node[sig, anchor=east] (i2) at (1.2,-1.5) {$v$};
\node[mdl, draw=GoogleGreen, fill=GoogleGreenLight] (m2) at (3.1,-1.5) {$p_\theta(a \mid v)$};
\node[sig, anchor=west] (o2) at (5.0,-1.5) {$\hat a$};
\draw[arr] (i2) -- (m2); \draw[arr] (m2) -- (o2);

\node[rowlab] at (0,-2.45) {audio-to-video\ \ (\S\ref{sec:crossgen-a2v}, Prob.~\ref{prob:crossgen})};
\node[sig, anchor=east] (i3) at (1.2,-3.0) {$a$};
\node[mdl, draw=GoogleGreen, fill=GoogleGreenLight] (m3) at (3.1,-3.0) {$p_\theta(v \mid a)$};
\node[sig, anchor=west] (o3) at (5.0,-3.0) {$\hat v$};
\draw[arr] (i3) -- (m3); \draw[arr] (m3) -- (o3);

\node[rowlab] at (0,-3.95) {joint editing\ \ (\S\ref{sec:jointedit}, Prob.~\ref{prob:jointedit})};
\node[sig, anchor=east] (i4) at (1.2,-4.5) {$(v, a),\, e$};
\node[mdl, draw=GoogleRed, fill=GoogleRedLight] (m4) at (3.1,-4.5) {$p_\theta(v', a' \mid v, a, e)$};
\node[sig, anchor=west] (o4) at (5.0,-4.5) {$v', a'$};
\draw[arr] (i4) -- (m4); \draw[arr] (m4) -- (o4);
\end{tikzpicture}
\caption{\textbf{The three problem settings as input-to-output mappings}, in the notation of Table~\ref{tab:notation}: what varies across rows is only which signals are given (left) and which are generated (right). Joint generation outputs both streams from a condition $c$ (which may be empty, text $c_t$, an image $c_i$, speech $c_s$, or music $c_m$); cross-modal generation outputs exactly one stream given the other; joint editing outputs a modified pair given a clip and an edit instruction $e$. Row color marks the family as in Figure~\ref{fig:roadmap}. Every setting must satisfy the correspondence requirement of Definition~\ref{def:correspondence}.}
\label{fig:problem-settings}
\end{figure}

Problem~\ref{prob:jointedit} adds two constraints absent from generation: the propagation of an edit across modalities, and the preservation of untouched content. As an example, given a clip of a person speaking and the instruction $e = $ ``change the speaker's voice to a child's voice,'' the model must produce $a'$ with the new voice and $v'$ in which the lip motion is consistent with $a'$, while leaving the background and identity intact. Editing thus inherits the correspondence requirement of generation and adds a fidelity-to-input requirement on top of it. Figure~\ref{fig:problem-settings} summarizes the three problems as mappings from inputs to outputs, Table~\ref{tab:task-formalization} instantiates them task by task, and Table~\ref{tab:notation} collects the notation.

\begin{table*}[t!]
\centering
\scriptsize
\renewcommand{\arraystretch}{1.2}
\caption{\textbf{A unified view of the tasks covered.} Each task is an instance of Problems~\ref{prob:jointgen} through~\ref{prob:jointedit}: it models the joint distribution over audio-visual pairs or one of its conditionals. \textbf{Given} lists the observed signals, \textbf{Generated} the produced signals, and \textbf{Primary Correspondence} the form of audio-visual agreement (Def.~\ref{def:correspondence}) that dominates evaluation. The table doubles as a map from a task to the section that treats it. $^{\dagger}$For talking-head generation the model re-synthesizes the speech track: $c_s$ conditions the output and $\hat a$ is emitted synchronized to $\hat v$, so speech appears as both condition and output.}
\label{tab:task-formalization}
\setlength{\tabcolsep}{6pt}
\begin{tabular}{@{}lllll l@{}}
\toprule
\textbf{Task} & \textbf{Given} & \textbf{Generated} & \textbf{Learned Distribution} & \textbf{Primary Correspondence} & \textbf{Section} \\
\midrule
Unconditional joint        & $\emptyset$      & $v, a$    & $p_\theta(v, a)$                 & semantic $+$ temporal      & \S\ref{sec:jointgen-unconditional} \\
Text-to-audio-video        & $c_t$            & $v, a$    & $p_\theta(v, a \mid c_t)$        & semantic $+$ temporal      & \S\ref{sec:jointgen-text} \\
Image-to-audio-video       & $c_i$            & $v, a$    & $p_\theta(v, a \mid c_i)$        & semantic $+$ temporal      & \S\ref{sec:jointgen-image} \\
Talking-head / speech-driven & $c_i, c_s$     & $v, a^{\dagger}$    & $p_\theta(v, a \mid c_i, c_s)$   & lip-sync                   & \S\ref{sec:jointgen-talkinghead} \\
Music-driven video         & $c_m$            & $v$       & $p_\theta(v \mid c_m)$           & beat / rhythm              & \S\ref{sec:crossgen-a2v} \\
Video-to-audio / Foley     & $v$              & $a$       & $p_\theta(a \mid v)$             & temporal (onset)           & \S\ref{sec:crossgen-v2a} \\
Audio-to-video             & $a$              & $v$       & $p_\theta(v \mid a)$             & temporal                   & \S\ref{sec:crossgen-a2v} \\
Speech-to-video            & $c_s$            & $v$       & $p_\theta(v \mid c_s)$           & lip-sync                   & \S\ref{sec:crossgen-s2v} \\
Joint editing              & $v, a, e$        & $v', a'$  & $p_\theta(v', a' \mid v, a, e)$  & preserve $+$ propagate     & \S\ref{sec:jointedit} \\
Dubbing / re-voicing       & $v, a, e$        & $v', a'$  & $p_\theta(v', a' \mid v, a, e)$  & lip-sync                   & \S\ref{sec:jointedit-sync} \\
\bottomrule
\end{tabular}
\end{table*}

\section{Background and Preliminaries}
\label{sec:background}
This section introduces the building blocks that joint and cross-modal methods inherit from the unimodal setting: the generative model families that instantiate $p_\theta$, the representations that realize Definition~\ref{def:latent} for each modality, and the conditioning mechanisms that inject $c$.

\subsection{Generative Modeling Foundations}
\label{sec:background-genmodels}
The methods we cover instantiate a generative model $p_\theta$ that approximates a target distribution over video, audio, or both. Four families recur: variational autoencoders~\cite{kingma2014vae}, generative adversarial networks~\cite{goodfellow2014gan}, autoregressive models~\cite{oord2016pixelrnn}, and diffusion and flow-matching models~\cite{ho2020ddpm,lipman2023flow}. Each corresponds to a different parameterization of $p_\theta$ and a different training objective, and each has been adapted to the joint and cross-modal setting in its own way. Recent work has converged on diffusion and flow-matching backbones for both video and audio synthesis, and a large fraction of the methods in later sections build directly on top of one. We therefore use the denoising formulation as the running example: a forward process gradually corrupts a clean latent $z_0$ into noise, and the model learns to reverse it, optionally conditioned on $c$, by predicting the noise or the velocity at each step.

\subsection{Video Representations}
\label{sec:background-videorep}
Several representational spaces realize $\phi_v$ in Definition~\ref{def:latent}, and the choice has direct consequences for the design of $p_\theta$. Pixel-space models operate on $v$ directly. Latent-space models, following the latent diffusion recipe~\cite{rombach2022ldm}, first encode $v$ into a lower-dimensional latent $z_v = \phi_v(v)$ and model the distribution over $z_v$ rather than over raw pixels, using either a 2D autoencoder applied per frame with a separate temporal module, or a 3D autoencoder that compresses space and time jointly. Token-based representations take a further step and discretize $v$ into a sequence of tokens, which is what enables autoregressive modeling. The three representations trade off fidelity, computational cost, and compatibility with the audio representation used alongside them, with the last factor mattering most in joint models that couple $v$ and $a$ in a shared space.

\subsection{Audio Representations}
\label{sec:background-audiorep}
An audio signal admits an analogous set of choices for $\phi_a$. Waveform-domain models operate on $a$ directly. Spectrogram-domain models first transform $a$ into a time-frequency representation such as a mel spectrogram and model that image-like array, relying on a neural vocoder to recover the waveform~\cite{kong2020hifigan}. Latent audio codecs~\cite{defossez2022encodec} encode $a$ into a sequence of discrete or continuous tokens and decode back to the waveform, playing a role analogous to latent video encoders. As with video, the representation chosen for audio interacts with the one chosen for video in joint models, since the two streams must be aligned in time and, for unified backbones, processed by a shared network. A recurring difficulty is the mismatch in native rate between the two modalities, since audio is sampled far more densely in time than video is, and the latent rates must be reconciled for the streams to be coupled frame by frame.

\subsection{Conditioning Mechanisms}
\label{sec:background-conditioning}
Once the representations are fixed, a conditioning signal $c$ enters the generative model through one of several mechanisms. Cross-attention injects $c$ into intermediate features of the network, giving fine-grained, position-dependent control. Adaptive normalization modulates feature statistics based on $c$, providing a lighter and coarser form of control. Concatenation appends an encoded $c$ to the input or to intermediate features. Classifier-free guidance~\cite{ho2022cfg} steers samples toward $c$ at inference time by mixing conditional and unconditional predictions. The form of $c$ together with the mechanism through which it is injected determines how tightly the output follows the condition, and in the joint setting the same mechanisms are reused to let one modality condition the other.

\subsection{Audio-Visual Correspondence in Practice}
\label{sec:background-avcorr}
Definition~\ref{def:correspondence} states correspondence as an abstract property; in practice it is operationalized through learned encoders that map a clip to a score $\mathcal{S}(v, a)$. Contrastive audio-visual encoders trained to pull matched pairs together and push mismatched pairs apart, of which ImageBind~\cite{girdhar2023imagebind} is a widely reused instance, provide such a score---temporally focused variants additionally treat time-shifted pairs as negatives~\cite{luo2023difffoley}---and several methods reuse these encoders either as a training signal or as a guidance term at inference. The remainder of this work is structured around how methods raise $\mathcal{S}$ while keeping per-modality quality high, since this is the property that distinguishes the joint and cross-modal problem from two unimodal ones.

\section{A Five-Axis Design Taxonomy}
\label{sec:taxonomy}

Methods that solve Problems~\ref{prob:jointgen} through~\ref{prob:jointedit} differ along a small number of design axes that, taken together, account for most of the variation in the literature. We propose a taxonomy that categorizes methods along five such axes, summarized in Table~\ref{tab:av-design-taxonomy}: the generation strategy, the audio representation, the video representation, the alignment-enforcement mechanism, and the reuse of pretrained weights.

Formally, a method is characterized by a tuple $(\phi, \theta, \mathcal{L}, \Gamma)$: the encoders $\phi = (\phi_v, \phi_a)$ of Definition~\ref{def:latent}, which fix the spaces $\mathcal{Z}_v, \mathcal{Z}_a$ in which generation occurs; the parameters $\theta$ of the model $p_\theta$ acting on those spaces; the training objective $\mathcal{L}$ used to train $\theta$; and the sampling procedure $\Gamma$ that draws $(\hat{z}_v, \hat{z}_a)$ from $p_\theta$. The five axes constrain different components of this tuple. The generation strategy (\S\ref{sec:strategy}) constrains how $\theta$ decomposes across the two streams; the audio and video representations (\S\ref{sec:audio-repr}, \S\ref{sec:video-repr}) constrain the codomains of $\phi_a$ and $\phi_v$; the alignment-enforcement mechanism (\S\ref{sec:sync}) constrains the stage at which correspondence pressure is applied, which may be in $\theta$, in the training objective $\mathcal{L}$, or in $\Gamma$; and pretraining reuse (\S\ref{sec:pretrain}) constrains the initialization of $\theta$. Because the axes constrain different components, a method makes a choice on each, and two choices on different axes are not alternatives to one another.

The axes are complementary rather than orthogonal. Each constrains a different component, so a method makes a choice along every axis, but the choices are not fully independent: a guidance-based strategy requires two frozen unimodal models and therefore entails dual pretraining, and pretraining reuse interacts with the representation axes in turn---every method covered here that reuses one or two pretrained backbones inherits its representation from them, and the backbones reused are without exception continuous-latent diffusion models, so the discrete-token cells can be filled only by training from scratch, the cost of Section~\ref{sec:open-scaling}, or by coupling token-based unimodal generators, a route no method we cover takes. Nor is every combination occupied---the empty and near-empty cells of Table~\ref{tab:av-design-taxonomy} are themselves informative, and we return to them in Section~\ref{sec:open-designspace}. We treat each axis in turn.

\subsection{Generation Strategy}
\label{sec:strategy}
The generation strategy is how the two modalities are produced relative to each other, that is, how the dependency required by Problem~\ref{prob:jointgen} is realized in the computation graph.

\subsubsection{Single-Tower}
\label{sec:strategy-single}
Formally $\theta_v = \theta_a = \theta_\times = \theta$: one network is applied to the concatenated sequence $[z_v ; z_a]$, so every parameter sees both streams and the dependency is carried by the parameters themselves.
A single network processes both modalities through shared parameters, with the two streams concatenated or interleaved into one sequence. Coupling is automatic, since every layer sees both modalities, at the cost of a representation that must serve both.
Single-tower designs are the basis of AV-DiT~\cite{wang2024avdit} and MM-LDM~\cite{sun2024mmldm}, and of recent open systems such as MOVA~\cite{mova2026}, Apollo (formerly Klear)~\cite{wang2026klear}, and 3MDiT~\cite{li2025threemdit}.

\subsubsection{Dual-Tower}
\label{sec:strategy-dual}
Formally $\theta = (\theta_v, \theta_a, \theta_\times)$ with $\theta_v \cap \theta_a = \emptyset$: each stream is processed by its own tower, and information is exchanged only through the cross-modal parameters $\theta_\times$, so the dependency between $v$ and $a$ is carried by $\theta_\times$ alone.
Two modality-specific towers run in parallel and exchange information through cross-modal connections such as cross-attention or bridge layers. The towers retain modality-specific inductive biases while the connections carry the dependency between $v$ and $a$.
This is the most common choice in the literature, adopted by MM-Diffusion~\cite{ruan2023mmdiffusion}, JavisDiT~\cite{liu2025javisdit}, BridgeDiT~\cite{guan2025bridgedit}, Ovi~\cite{low2025ovi}, UniAVGen~\cite{zhang2025uniavgen}, and LTX-2~\cite{hacohen2026ltx2}, among others.

\subsubsection{Cascaded}
\label{sec:strategy-cascade}
Formally $p_\theta(v, a \mid c) = p_{\theta_1}(v \mid c)\, p_{\theta_2}(a \mid v, c)$, or the symmetric factorization, with disjoint $\theta_1, \theta_2$ and sequential sampling; the coupling is carried by the conditioning path rather than by shared parameters.
The modalities are produced in sequence, with the second conditioned on the first, which reduces joint generation to a generation step followed by a cross-modal step. Movie Gen~\cite{polyak2024moviegen} is the representative instance, generating video from text and then audio from the generated video.

\subsubsection{Unified-Token}
\label{sec:strategy-token}
Formally $\phi_v, \phi_a$ are quantizers into finite vocabularies, the pair is serialized into a single sequence $s = \pi(z_v, z_a)$, and $p_\theta(s) = \prod_t p_\theta(s_t \mid s_{<t})$ or a masked variant.
Both modalities are discretized into tokens and modeled as a single sequence by an autoregressive or masked transformer, so the dependency is captured by the sequence model and the same backbone can serve multiple tasks by reordering inputs and outputs. No joint method covered here adopts this strategy. UniForm~\cite{zhao2025uniform} comes closest---it serializes the two modalities into one sequence and shares a denoiser across tasks distinguished by task tokens---but it does so over continuous VAE latents with a diffusion objective rather than over discrete vocabularies, which places it in the single-tower category. The unified-token cell is thus unoccupied among joint models, in contrast to the unimodal literature where token-based video and audio generation are established, an asymmetry we return to in Section~\ref{sec:open-designspace}.

\subsubsection{Guidance-Based}
\label{sec:strategy-guidance}
Formally the two frozen models are coupled only in the sampler $\Gamma$, for example by adjusting the score of the pair $z = (z_v, z_a)$ as $\nabla_z \log p_\theta(z \mid c) + \lambda\, \nabla_z \mathcal{S}\big(\psi_v(z_v), \psi_a(z_a)\big)$ with guidance weight $\lambda$; $\theta$ is never updated jointly.
Two pretrained unimodal models are frozen and coupled only at inference, through a guidance term such as a classifier or an alignment score that nudges the two samples toward mutual consistency. No joint training is required, which trades fidelity for flexibility.
Seeing-and-Hearing~\cite{xing2024seeing} couples two frozen generators through an ImageBind~\cite{girdhar2023imagebind} alignment score, and MMDisCo~\cite{hayakawa2025mmdisco} through a trained discriminator applied as sampling guidance.

\subsection{Audio Representation}
\label{sec:audio-repr}
The audio representation is the realization of $\phi_a$ in Definition~\ref{def:latent}, and it fixes the space in which audio is generated. We discuss each choice in turn.

\subsubsection{Continuous Latent}
\label{sec:audio-latent}
A neural audio autoencoder maps the waveform to a compact continuous latent in which a diffusion or flow model is trained. Writing the encoder as $\phi_a$ and its decoder as $\psi_a$ (Table~\ref{tab:notation}), the autoencoder is trained so that
\begin{equation}
z_a = \phi_a(a) \in \mathbb{R}^{T_z \times d}, \qquad \psi_a(z_a) \approx a,
\label{eq:audio-ae}
\end{equation}
with a reconstruction loss plus, in the variational form~\cite{kingma2014vae}, a KL regularizer on the encoder's posterior over $z_a$; here $T_z \ll T_a$ is the compressed length and $d$ the channel width. Generation then follows latent diffusion~\cite{ho2020ddpm,rombach2022ldm}: at diffusion step $u$, noised latents $z_u = \sqrt{\bar\alpha_u}\, z_a + \sqrt{1-\bar\alpha_u}\, \epsilon$ with $\epsilon \sim \mathcal{N}(0, I)$ are drawn along a noise schedule $\bar\alpha_u$, a network $\epsilon_\theta$ is trained to minimize
\begin{equation}
\mathcal{L} \;=\; \mathbb{E}_{z_a,\, u,\, \epsilon}\,\big\lVert \epsilon - \epsilon_\theta(z_u, u, c) \big\rVert_2^2,
\label{eq:latent-diffusion}
\end{equation}
and sampling denoises from pure noise to $\hat z_a$, decoded as $\hat a = \psi_a(\hat z_a)$. This is the dominant choice in recent joint models: it is used by twenty-eight of the twenty-nine methods in Table~\ref{tab:av-design-taxonomy}, spanning early dual-tower models such as CoDi~\cite{tang2023codi} through recent systems including LTX-2~\cite{hacohen2026ltx2} and MOVA~\cite{mova2026}.

\subsubsection{Discrete Tokens}
\label{sec:audio-tokens}
A neural codec instead quantizes the latent: a codebook $\mathcal{C} = \{e_1, \dots, e_K\} \subset \mathbb{R}^{d}$ replaces each latent frame $z_a^{(i)}$ by its nearest entry,
\begin{equation}
q\big(z_a^{(i)}\big) = e_{k^\ast}, \qquad k^\ast = \arg\min\nolimits_k \big\lVert z_a^{(i)} - e_k \big\rVert_2,
\label{eq:vq}
\end{equation}
trained with codebook and commitment terms $\lVert \mathrm{sg}[z_a^{(i)}] - e_{k^\ast} \rVert_2^2 + \beta\, \lVert z_a^{(i)} - \mathrm{sg}[e_{k^\ast}] \rVert_2^2$ under a straight-through gradient, where $\mathrm{sg}[\cdot]$ denotes stop-gradient~\cite{vandenoord2017vqvae}; practical audio codecs quantize residually over a stack of such codebooks, typically replacing the codebook term with an exponential-moving-average update of the selected entries~\cite{defossez2022encodec}. The resulting index sequence supports autoregressive or masked modeling and a shared treatment with tokenized video. No joint method covered here, however, generates audio as codec tokens---UniForm~\cite{zhao2025uniform} serializes the modalities into one sequence but over continuous latents (\S\ref{sec:audio-latent})---leaving this cell, like the waveform, unoccupied.

\subsubsection{Mel-Spectrogram}
\label{sec:audio-mel}
Audio is represented as a mel spectrogram, obtained from the short-time Fourier transform through a mel filterbank $M$,
\begin{equation}
m \;=\; \log\big(M\,\lvert \mathrm{STFT}(a) \rvert^{2}\big) \in \mathbb{R}^{F \times T_m},
\label{eq:mel}
\end{equation}
with $F$ mel bins and $T_m$ spectral frames, and treated as an image-like array, which makes image generative machinery directly applicable but requires a separate vocoder to recover the waveform from a generated $\hat m$.
MM-Diffusion~\cite{ruan2023mmdiffusion} is the representative instance among joint models.

\subsubsection{Waveform}
\label{sec:audio-wav}
The model operates on the raw waveform $a \in \mathbb{R}^{T_a}$ directly, preserving full fidelity at the cost of modeling a very long and densely sampled sequence. No method we cover generates audio directly in the waveform domain, a consequence of the sequence lengths involved at audio sampling rates, an unoccupied corner we return to in Section~\ref{sec:open-designspace}.

\subsection{Video Representation}
\label{sec:video-repr}
The video representation is the realization of $\phi_v$, with the same fidelity, cost, and compatibility trade-offs; the constructions mirror Equations~\ref{eq:audio-ae}--\ref{eq:vq} with $v$ in place of $a$, and we discuss each choice in turn.

\subsubsection{3D-VAE}
\label{sec:video-3d}
A 3D autoencoder compresses space and time jointly,
\begin{equation}
z_v = \phi_v(v) \in \mathbb{R}^{\,T_v/s_t \,\times\, H/s_s \,\times\, W/s_s \,\times\, d_v},
\label{eq:video-3d}
\end{equation}
with temporal stride $s_t$, spatial stride $s_s$, and channel width $d_v$, producing a spatio-temporal latent in which a single backbone models the whole clip under the objective of Equation~\ref{eq:latent-diffusion}. This is the dominant choice in recent video and joint models.
Nineteen of the methods in Table~\ref{tab:av-design-taxonomy} use it, including Movie Gen~\cite{polyak2024moviegen}, JavisDiT~\cite{liu2025javisdit}, Ovi~\cite{low2025ovi}, LTX-2~\cite{hacohen2026ltx2}, Apollo~\cite{wang2026klear}, and---through the pretrained Open-Sora autoencoder---UniForm~\cite{zhao2025uniform}.

\subsubsection{2D-VAE plus Temporal}
\label{sec:video-2d}
A per-frame 2D autoencoder compresses each frame independently, $z_v^{(t)} = \phi_{v,\mathrm{2D}}\big(v^{(t)}\big)$ for $t = 1, \dots, T_v$, and a separate temporal module models motion across the stacked latent frames $\big(z_v^{(1)}, \dots, z_v^{(T_v)}\big)$.
CoDi~\cite{tang2023codi}, AV-DiT~\cite{wang2024avdit}, SVG~\cite{ishii2024svg}, and Animate-and-Sound~\cite{wang2025jointdit} take this route.

\subsubsection{Discrete Tokens}
\label{sec:video-tokens}
Video is quantized into discrete tokens by the construction of Equation~\ref{eq:vq} applied to spatio-temporal patches, enabling autoregressive or masked modeling and a shared sequence with tokenized audio.
No joint method covered here takes this route---UniForm~\cite{zhao2025uniform} fuses the streams as continuous latent tokens (\S\ref{sec:video-3d})---mirroring the unoccupied discrete-audio cell.

\subsubsection{Pixel}
\label{sec:video-pixel}
The model operates directly on pixels $v \in \mathbb{R}^{T_v \times H \times W \times 3}$, with no learned compression of the visual stream.
MM-Diffusion~\cite{ruan2023mmdiffusion} is the sole pixel-space instance, reflecting the resolutions feasible when the work appeared.

\subsection{Alignment Enforcement}
\label{sec:sync}
Alignment enforcement is how a method raises the alignment score $\mathcal{S}$ of Definition~\ref{def:correspondence}, that is, the stage of the pipeline at which correspondence between the two streams is imposed (see Section~\ref{sec:align}).

\subsubsection{Cross-Attention}
\label{sec:sync-xattn}
Cross-attention layers let the audio stream attend to the video stream and vice versa, carrying alignment information between the two towers or branches---an architectural mechanism.
It is the dominant mechanism, used by nineteen methods including MM-Diffusion~\cite{ruan2023mmdiffusion}, BridgeDiT~\cite{guan2025bridgedit}, Ovi~\cite{low2025ovi}, and OmniForcing~\cite{su2026omniforcing}.

\subsubsection{Shared Positional Encoding}
\label{sec:sync-pe}
A shared positional or rotary encoding~\cite{su2021roformer} ties the two streams to a common time axis, so that tokens at the same physical time are forced into correspondence at the level of position---likewise an architectural mechanism.
ALIVE~\cite{guo2026alive} and JavisDiT++~\cite{liu2026javisditpp} use temporally aligned rotary encodings for frame-level correspondence; AV-DiT~\cite{wang2024avdit}, Apollo~\cite{wang2026klear}, and 3MDiT~\cite{li2025threemdit} share positional information across streams.

\subsubsection{Discriminator}
\label{sec:sync-disc}
An auxiliary joint discriminator is trained to score whether a pair is jointly real, and its gradient is applied as guidance at sampling time while the base generators stay frozen---an inference-time mechanism rather than a training objective for the generators.
MMDisCo~\cite{hayakawa2025mmdisco}, a trained discriminator applied as sampling guidance, is the only instance we cover.

\subsubsection{Classifier Guidance}
\label{sec:sync-guide}
An external classifier or alignment model guides sampling toward consistent pairs at inference, in the manner of classifier guidance for diffusion models~\cite{dhariwal2021beatgans}, without changing the generator's weights---an inference-time mechanism.
Seeing-and-Hearing~\cite{xing2024seeing} is the representative case.

\subsubsection{Explicit Prior}
\label{sec:sync-prior}
A separately estimated spatio-temporal prior is injected to align the streams, decoupling the synchronization signal from the generators---an architectural or an external mechanism, depending on whether the prior is learned jointly or estimated separately.
JavisDiT~\cite{liu2025javisdit} estimates a hierarchical spatio-temporal prior for this purpose, retained in JavisDiT++~\cite{liu2026javisditpp}.

\subsection{Pretraining Reuse}
\label{sec:pretrain}
The final axis is the source of the model's weights, which strongly affects data and compute cost (see Section~\ref{sec:open-scaling}).

\subsubsection{From-Scratch}
\label{sec:pretrain-scratch}
The model is trained from random initialization on paired audio-visual data.
Twelve methods train this way, including MM-Diffusion~\cite{ruan2023mmdiffusion}, Movie Gen~\cite{polyak2024moviegen}, JavisDiT~\cite{liu2025javisdit}, Ovi~\cite{low2025ovi}, LTX-2~\cite{hacohen2026ltx2}, and MOVA~\cite{mova2026}.

\subsubsection{Single Pretrained}
\label{sec:pretrain-single}
One pretrained backbone, typically a video model, is reused and the other modality is added on top.
ALIVE~\cite{guo2026alive}, Hallo-Live~\cite{li2026hallolive}, and OmniForcing~\cite{su2026omniforcing} build on one pretrained backbone.

\subsubsection{Dual Pretrained}
\label{sec:pretrain-dual}
Two pretrained backbones, one per modality, are reused and coupled, so that joint training only learns the connections.
CoDi~\cite{tang2023codi}, Seeing-and-Hearing~\cite{xing2024seeing}, SVG~\cite{ishii2024svg}, BridgeDiT~\cite{guan2025bridgedit}, and CCL~\cite{ma2026ccl} couple two pretrained models.

\section{Joint Audio-Visual Editing}
\label{sec:jointedit}
Joint editing outputs a modified pair, $(v', a') \sim p_\theta(v', a' \mid v, a, e)$ (Problem~\ref{prob:jointedit}), propagating the edit across modalities while preserving untouched content.

\subsection{The Space of Audio-Visual Edits}
\label{sec:jointedit-space}
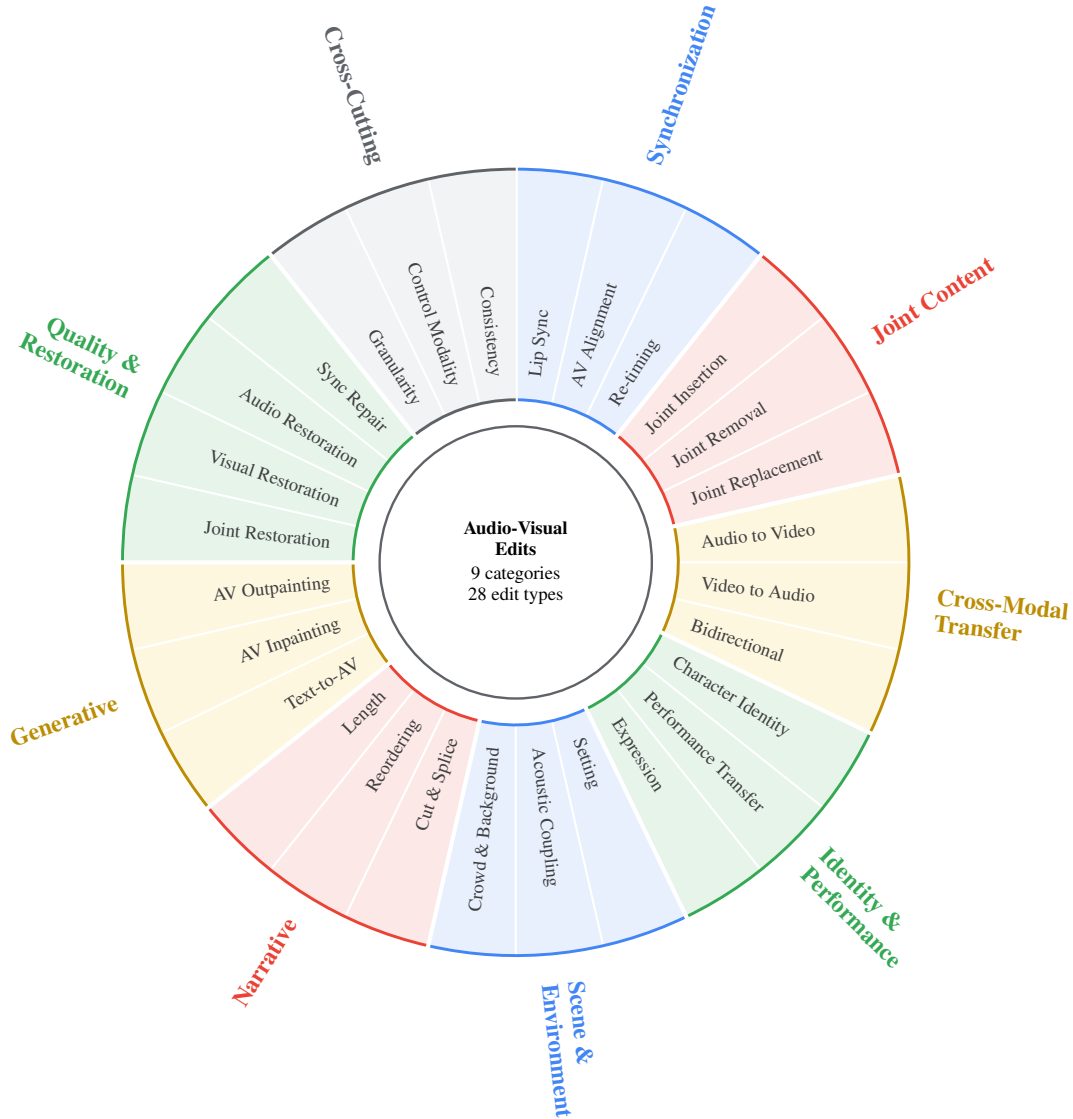
\begin{figure*}[t!]
\centering
\begin{tikzpicture}[font=\scriptsize, line width=0.5pt]
\fill[GoogleBlueLight] (90.000:2.15) -- (90.000:5.2) arc (90.000:77.143:5.2) -- (77.143:2.15) arc (77.143:90.000:2.15) -- cycle;
\draw[white, line width=0.8pt] (90.000:2.15) -- (90.000:5.2);
\fill[GoogleBlueLight] (77.143:2.15) -- (77.143:5.2) arc (77.143:64.286:5.2) -- (64.286:2.15) arc (64.286:77.143:2.15) -- cycle;
\draw[white, line width=0.8pt] (77.143:2.15) -- (77.143:5.2);
\fill[GoogleBlueLight] (64.286:2.15) -- (64.286:5.2) arc (64.286:51.429:5.2) -- (51.429:2.15) arc (51.429:64.286:2.15) -- cycle;
\draw[white, line width=0.8pt] (64.286:2.15) -- (64.286:5.2);
\draw[GoogleBlue, line width=1.1pt] (90.000:5.2) arc (90.000:51.429:5.2);
\draw[GoogleBlue, line width=1.1pt] (90.000:2.15) arc (90.000:51.429:2.15);
\draw[white, line width=1.6pt] (90.000:2.1) -- (90.000:5.25);
\fill[GoogleRedLight] (51.429:2.15) -- (51.429:5.2) arc (51.429:38.571:5.2) -- (38.571:2.15) arc (38.571:51.429:2.15) -- cycle;
\draw[white, line width=0.8pt] (51.429:2.15) -- (51.429:5.2);
\fill[GoogleRedLight] (38.571:2.15) -- (38.571:5.2) arc (38.571:25.714:5.2) -- (25.714:2.15) arc (25.714:38.571:2.15) -- cycle;
\draw[white, line width=0.8pt] (38.571:2.15) -- (38.571:5.2);
\fill[GoogleRedLight] (25.714:2.15) -- (25.714:5.2) arc (25.714:12.857:5.2) -- (12.857:2.15) arc (12.857:25.714:2.15) -- cycle;
\draw[white, line width=0.8pt] (25.714:2.15) -- (25.714:5.2);
\draw[GoogleRed, line width=1.1pt] (51.429:5.2) arc (51.429:12.857:5.2);
\draw[GoogleRed, line width=1.1pt] (51.429:2.15) arc (51.429:12.857:2.15);
\draw[white, line width=1.6pt] (51.429:2.1) -- (51.429:5.25);
\fill[GoogleYellowLight] (12.857:2.15) -- (12.857:5.2) arc (12.857:0.000:5.2) -- (0.000:2.15) arc (0.000:12.857:2.15) -- cycle;
\draw[white, line width=0.8pt] (12.857:2.15) -- (12.857:5.2);
\fill[GoogleYellowLight] (0.000:2.15) -- (0.000:5.2) arc (0.000:-12.857:5.2) -- (-12.857:2.15) arc (-12.857:0.000:2.15) -- cycle;
\draw[white, line width=0.8pt] (0.000:2.15) -- (0.000:5.2);
\fill[GoogleYellowLight] (-12.857:2.15) -- (-12.857:5.2) arc (-12.857:-25.714:5.2) -- (-25.714:2.15) arc (-25.714:-12.857:2.15) -- cycle;
\draw[white, line width=0.8pt] (-12.857:2.15) -- (-12.857:5.2);
\draw[GoogleYellow!75!black, line width=1.1pt] (12.857:5.2) arc (12.857:-25.714:5.2);
\draw[GoogleYellow!75!black, line width=1.1pt] (12.857:2.15) arc (12.857:-25.714:2.15);
\draw[white, line width=1.6pt] (12.857:2.1) -- (12.857:5.25);
\fill[GoogleGreenLight] (-25.714:2.15) -- (-25.714:5.2) arc (-25.714:-38.571:5.2) -- (-38.571:2.15) arc (-38.571:-25.714:2.15) -- cycle;
\draw[white, line width=0.8pt] (-25.714:2.15) -- (-25.714:5.2);
\fill[GoogleGreenLight] (-38.571:2.15) -- (-38.571:5.2) arc (-38.571:-51.429:5.2) -- (-51.429:2.15) arc (-51.429:-38.571:2.15) -- cycle;
\draw[white, line width=0.8pt] (-38.571:2.15) -- (-38.571:5.2);
\fill[GoogleGreenLight] (-51.429:2.15) -- (-51.429:5.2) arc (-51.429:-64.286:5.2) -- (-64.286:2.15) arc (-64.286:-51.429:2.15) -- cycle;
\draw[white, line width=0.8pt] (-51.429:2.15) -- (-51.429:5.2);
\draw[GoogleGreen, line width=1.1pt] (-25.714:5.2) arc (-25.714:-64.286:5.2);
\draw[GoogleGreen, line width=1.1pt] (-25.714:2.15) arc (-25.714:-64.286:2.15);
\draw[white, line width=1.6pt] (-25.714:2.1) -- (-25.714:5.25);
\fill[GoogleBlueLight] (-64.286:2.15) -- (-64.286:5.2) arc (-64.286:-77.143:5.2) -- (-77.143:2.15) arc (-77.143:-64.286:2.15) -- cycle;
\draw[white, line width=0.8pt] (-64.286:2.15) -- (-64.286:5.2);
\fill[GoogleBlueLight] (-77.143:2.15) -- (-77.143:5.2) arc (-77.143:-90.000:5.2) -- (-90.000:2.15) arc (-90.000:-77.143:2.15) -- cycle;
\draw[white, line width=0.8pt] (-77.143:2.15) -- (-77.143:5.2);
\fill[GoogleBlueLight] (-90.000:2.15) -- (-90.000:5.2) arc (-90.000:-102.857:5.2) -- (-102.857:2.15) arc (-102.857:-90.000:2.15) -- cycle;
\draw[white, line width=0.8pt] (-90.000:2.15) -- (-90.000:5.2);
\draw[GoogleBlue, line width=1.1pt] (-64.286:5.2) arc (-64.286:-102.857:5.2);
\draw[GoogleBlue, line width=1.1pt] (-64.286:2.15) arc (-64.286:-102.857:2.15);
\draw[white, line width=1.6pt] (-64.286:2.1) -- (-64.286:5.25);
\fill[GoogleRedLight] (-102.857:2.15) -- (-102.857:5.2) arc (-102.857:-115.714:5.2) -- (-115.714:2.15) arc (-115.714:-102.857:2.15) -- cycle;
\draw[white, line width=0.8pt] (-102.857:2.15) -- (-102.857:5.2);
\fill[GoogleRedLight] (-115.714:2.15) -- (-115.714:5.2) arc (-115.714:-128.571:5.2) -- (-128.571:2.15) arc (-128.571:-115.714:2.15) -- cycle;
\draw[white, line width=0.8pt] (-115.714:2.15) -- (-115.714:5.2);
\fill[GoogleRedLight] (-128.571:2.15) -- (-128.571:5.2) arc (-128.571:-141.429:5.2) -- (-141.429:2.15) arc (-141.429:-128.571:2.15) -- cycle;
\draw[white, line width=0.8pt] (-128.571:2.15) -- (-128.571:5.2);
\draw[GoogleRed, line width=1.1pt] (-102.857:5.2) arc (-102.857:-141.429:5.2);
\draw[GoogleRed, line width=1.1pt] (-102.857:2.15) arc (-102.857:-141.429:2.15);
\draw[white, line width=1.6pt] (-102.857:2.1) -- (-102.857:5.25);
\fill[GoogleYellowLight] (-141.429:2.15) -- (-141.429:5.2) arc (-141.429:-154.286:5.2) -- (-154.286:2.15) arc (-154.286:-141.429:2.15) -- cycle;
\draw[white, line width=0.8pt] (-141.429:2.15) -- (-141.429:5.2);
\fill[GoogleYellowLight] (-154.286:2.15) -- (-154.286:5.2) arc (-154.286:-167.143:5.2) -- (-167.143:2.15) arc (-167.143:-154.286:2.15) -- cycle;
\draw[white, line width=0.8pt] (-154.286:2.15) -- (-154.286:5.2);
\fill[GoogleYellowLight] (-167.143:2.15) -- (-167.143:5.2) arc (-167.143:-180.000:5.2) -- (-180.000:2.15) arc (-180.000:-167.143:2.15) -- cycle;
\draw[white, line width=0.8pt] (-167.143:2.15) -- (-167.143:5.2);
\draw[GoogleYellow!75!black, line width=1.1pt] (-141.429:5.2) arc (-141.429:-180.000:5.2);
\draw[GoogleYellow!75!black, line width=1.1pt] (-141.429:2.15) arc (-141.429:-180.000:2.15);
\draw[white, line width=1.6pt] (-141.429:2.1) -- (-141.429:5.25);
\fill[GoogleGreenLight] (-180.000:2.15) -- (-180.000:5.2) arc (-180.000:-192.857:5.2) -- (-192.857:2.15) arc (-192.857:-180.000:2.15) -- cycle;
\draw[white, line width=0.8pt] (-180.000:2.15) -- (-180.000:5.2);
\fill[GoogleGreenLight] (-192.857:2.15) -- (-192.857:5.2) arc (-192.857:-205.714:5.2) -- (-205.714:2.15) arc (-205.714:-192.857:2.15) -- cycle;
\draw[white, line width=0.8pt] (-192.857:2.15) -- (-192.857:5.2);
\fill[GoogleGreenLight] (-205.714:2.15) -- (-205.714:5.2) arc (-205.714:-218.571:5.2) -- (-218.571:2.15) arc (-218.571:-205.714:2.15) -- cycle;
\draw[white, line width=0.8pt] (-205.714:2.15) -- (-205.714:5.2);
\fill[GoogleGreenLight] (-218.571:2.15) -- (-218.571:5.2) arc (-218.571:-231.429:5.2) -- (-231.429:2.15) arc (-231.429:-218.571:2.15) -- cycle;
\draw[white, line width=0.8pt] (-218.571:2.15) -- (-218.571:5.2);
\draw[GoogleGreen, line width=1.1pt] (-180.000:5.2) arc (-180.000:-231.429:5.2);
\draw[GoogleGreen, line width=1.1pt] (-180.000:2.15) arc (-180.000:-231.429:2.15);
\draw[white, line width=1.6pt] (-180.000:2.1) -- (-180.000:5.25);
\fill[GoogleGrayLight] (-231.429:2.15) -- (-231.429:5.2) arc (-231.429:-244.286:5.2) -- (-244.286:2.15) arc (-244.286:-231.429:2.15) -- cycle;
\draw[white, line width=0.8pt] (-231.429:2.15) -- (-231.429:5.2);
\fill[GoogleGrayLight] (-244.286:2.15) -- (-244.286:5.2) arc (-244.286:-257.143:5.2) -- (-257.143:2.15) arc (-257.143:-244.286:2.15) -- cycle;
\draw[white, line width=0.8pt] (-244.286:2.15) -- (-244.286:5.2);
\fill[GoogleGrayLight] (-257.143:2.15) -- (-257.143:5.2) arc (-257.143:-270.000:5.2) -- (-270.000:2.15) arc (-270.000:-257.143:2.15) -- cycle;
\draw[white, line width=0.8pt] (-257.143:2.15) -- (-257.143:5.2);
\draw[GoogleGray, line width=1.1pt] (-231.429:5.2) arc (-231.429:-270.000:5.2);
\draw[GoogleGray, line width=1.1pt] (-231.429:2.15) arc (-231.429:-270.000:2.15);
\draw[white, line width=1.6pt] (-231.429:2.1) -- (-231.429:5.25);
\node[rotate=83.57, anchor=west, text=black!75] at (83.571:2.33) {Lip Sync};
\node[rotate=70.71, anchor=west, text=black!75] at (70.714:2.33) {AV Alignment};
\node[rotate=57.86, anchor=west, text=black!75] at (57.857:2.33) {Re-timing};
\node[rotate=45.00, anchor=west, text=black!75] at (45.000:2.33) {Joint Insertion};
\node[rotate=32.14, anchor=west, text=black!75] at (32.143:2.33) {Joint Removal};
\node[rotate=19.29, anchor=west, text=black!75] at (19.286:2.33) {Joint Replacement};
\node[rotate=6.43, anchor=west, text=black!75] at (6.429:2.33) {Audio to Video};
\node[rotate=-6.43, anchor=west, text=black!75] at (-6.429:2.33) {Video to Audio};
\node[rotate=-19.29, anchor=west, text=black!75] at (-19.286:2.33) {Bidirectional};
\node[rotate=-32.14, anchor=west, text=black!75] at (-32.143:2.33) {Character Identity};
\node[rotate=-45.00, anchor=west, text=black!75] at (-45.000:2.33) {Performance Transfer};
\node[rotate=-57.86, anchor=west, text=black!75] at (-57.857:2.33) {Expression};
\node[rotate=-70.71, anchor=west, text=black!75] at (-70.714:2.33) {Setting};
\node[rotate=-83.57, anchor=west, text=black!75] at (-83.571:2.33) {Acoustic Coupling};
\node[rotate=83.57, anchor=east, text=black!75] at (-96.429:2.33) {Crowd \& Background};
\node[rotate=70.71, anchor=east, text=black!75] at (-109.286:2.33) {Cut \& Splice};
\node[rotate=57.86, anchor=east, text=black!75] at (-122.143:2.33) {Reordering};
\node[rotate=45.00, anchor=east, text=black!75] at (-135.000:2.33) {Length};
\node[rotate=32.14, anchor=east, text=black!75] at (-147.857:2.33) {Text-to-AV};
\node[rotate=19.29, anchor=east, text=black!75] at (-160.714:2.33) {AV Inpainting};
\node[rotate=6.43, anchor=east, text=black!75] at (-173.571:2.33) {AV Outpainting};
\node[rotate=-6.43, anchor=east, text=black!75] at (-186.429:2.33) {Joint Restoration};
\node[rotate=-19.29, anchor=east, text=black!75] at (-199.286:2.33) {Visual Restoration};
\node[rotate=-32.14, anchor=east, text=black!75] at (-212.143:2.33) {Audio Restoration};
\node[rotate=-45.00, anchor=east, text=black!75] at (-225.000:2.33) {Sync Repair};
\node[rotate=-57.86, anchor=east, text=black!75] at (-237.857:2.33) {Granularity};
\node[rotate=-70.71, anchor=east, text=black!75] at (-250.714:2.33) {Control Modality};
\node[rotate=-83.57, anchor=east, text=black!75] at (-263.571:2.33) {Consistency};
\node[rotate=70.71, anchor=west, align=left, font=\small\bfseries, text=GoogleBlue] at (70.714:5.45) {Synchronization};
\node[rotate=32.14, anchor=west, align=left, font=\small\bfseries, text=GoogleRed] at (32.143:5.45) {Joint Content};
\node[rotate=-6.43, anchor=west, align=left, font=\small\bfseries, text=GoogleYellow!75!black] at (-6.429:5.45) {Cross-Modal\\Transfer};
\node[rotate=-45.00, anchor=west, align=left, font=\small\bfseries, text=GoogleGreen] at (-45.000:5.45) {Identity \&\\Performance};
\node[rotate=-83.57, anchor=west, align=left, font=\small\bfseries, text=GoogleBlue] at (-83.571:5.45) {Scene \&\\Environment};
\node[rotate=57.86, anchor=east, align=right, font=\small\bfseries, text=GoogleRed] at (-122.143:5.45) {Narrative};
\node[rotate=19.29, anchor=east, align=right, font=\small\bfseries, text=GoogleYellow!75!black] at (-160.714:5.45) {Generative};
\node[rotate=-25.71, anchor=east, align=right, font=\small\bfseries, text=GoogleGreen] at (-205.714:5.45) {Quality \&\\Restoration};
\node[rotate=-70.71, anchor=east, align=right, font=\small\bfseries, text=GoogleGray] at (-250.714:5.45) {Cross-Cutting};
\draw[GoogleGray, fill=white, line width=0.9pt] (0,0) circle (1.7999999999999998);
\node[align=center] at (0,0) {\textbf{Audio-Visual}\\\textbf{Edits}\\[1pt] 9 categories\\ 28 edit types};
\end{tikzpicture}
\caption{\textbf{The edit-type taxonomy as a wheel.} The nine categories of Table~\ref{tab:audiovisual-edits-taxonomy} with their edit types arranged radially, in the same order and colors as the table; sector size is proportional to the number of edit types. Representative operations and example use cases for each type are enumerated in Table~\ref{tab:audiovisual-edits-taxonomy}.}
\label{fig:edit-wheel}
\end{figure*}

This section mirrors Table~\ref{tab:audiovisual-edits-taxonomy} one to one: nine categories of edit types with representative operations and use cases (rendered radially in Figure~\ref{fig:edit-wheel}). One-stream edits retain the other stream unchanged. Development concentrates in joint content, synchronization, and text-driven control; the empty cells are the research agenda.

\begin{table*}[t!]
\centering
\caption{\textbf{A taxonomy of audio-visual edits:} nine categories broken into 28 edit types, each with representative edits and an example use case. Each type is additionally annotated by its dominant modality coupling: \textbf{A+V} denotes a genuinely joint edit that requires reasoning over both modalities; \textbf{V$\rightarrow$A} denotes a video-driven edit with an audio consequence (or audio derived from video); \textbf{A$\rightarrow$V} denotes the reverse; \textbf{V} and \textbf{A} denote edits that are primarily single-modality, with the other modality passive or unchanged. Categories are color-coded for clarity.}
\label{tab:audiovisual-edits-taxonomy}
\renewcommand{\arraystretch}{1.2}
\scriptsize
\setlength{\tabcolsep}{6pt}
\begin{tabular}{@{}p{2.0cm} p{2.8cm} c p{5.0cm} >{\RaggedRight\arraybackslash}p{3.4cm}@{}}
\toprule
\textbf{Category} & \textbf{Edit/Gen. Type}
& \textbf{Modality} & \textbf{Representative Edits} & \textbf{Example Use Case} \\
\midrule
\multirow{3}{=}{\textcolor{GoogleBlue}{\textbf{Synchronization}}}
 & \textcolor{GoogleBlue}{Lip Sync}     & A+V & Lip sync correction, dubbing alignment, viseme generation & Aligning dubbed audio to mouth motion \\
 & \textcolor{GoogleBlue}{AV Alignment} & A+V & Foley alignment, beat alignment, event sync & Matching footsteps to visual steps \\
 & \textcolor{GoogleBlue}{Re-timing}    & A+V & Joint time stretch, slow motion, speed ramping & Slowing a scene with pitch-preserved audio \\
\midrule
\multirow{3}{=}{\textcolor{GoogleRed}{\textbf{Joint Content}}}
 & \textcolor{GoogleRed}{Joint Insertion}   & A+V & Object with sound, character with voice, ambience addition & Adding a passing car with engine noise \\
 & \textcolor{GoogleRed}{Joint Removal}     & A+V & Object removal with sound suppression, character removal & Removing a person and their voice \\
 & \textcolor{GoogleRed}{Joint Replacement} & A+V & Object swap, scene swap, character swap & Replacing a dog with a cat (visual + sound) \\
\midrule
\multirow{3}{=}{\textcolor{GoogleYellow!75!black}{\textbf{Cross-Modal Transfer}}}
 & \textcolor{GoogleYellow!75!black}{Audio to Video} & A$\rightarrow$V & Talking head from speech, music-driven video & Animating a portrait from voice \\
 & \textcolor{GoogleYellow!75!black}{Video to Audio} & V$\rightarrow$A & Foley from video, ambience from scene, music from mood & Generating sound effects from silent video \\
 & \textcolor{GoogleYellow!75!black}{Bidirectional}  & A+V & AV style transfer, joint enhancement, coupled denoising & Stylizing both modalities to a reference \\
\midrule
\multirow{3}{=}{\textcolor{GoogleGreen}{\textbf{Identity \& Perf.}}}
 & \textcolor{GoogleGreen}{Character Identity}   & A+V & Face swap with voice swap, aging, full replacement & Replacing an actor in face and voice \\
 & \textcolor{GoogleGreen}{Performance Transfer} & A+V & Facial reenactment, gesture transfer, puppetry & Driving an avatar with a real performance \\
 & \textcolor{GoogleGreen}{Expression}           & A+V & Emotion editing across face and voice, energy, persona & Making a sad scene appear joyful \\
\midrule
\multirow{3}{=}{\textcolor{GoogleBlue}{\textbf{Scene \& Environment}}}
 & \textcolor{GoogleBlue}{Setting}             & A+V & Location transfer, time-of-day, weather & Changing day to night with night ambience \\
 & \textcolor{GoogleBlue}{Acoustic Coupling}   & V$\rightarrow$A & Reverb to visual space, occlusion, acoustic shadows & Matching reverb to a depicted cathedral \\
 & \textcolor{GoogleBlue}{Crowd \& Background} & A+V & Crowd density with noise, traffic with engines & Adding a crowd with crowd noise \\
\midrule
\multirow{3}{=}{\textcolor{GoogleRed}{\textbf{Narrative}}}
 & \textcolor{GoogleRed}{Cut \& Splice} & A+V & AV cut detection, J-cuts, L-cuts, montage & Editing dialogue with overlapping audio \\
 & \textcolor{GoogleRed}{Reordering}    & A+V & Scene reordering, dialogue reordering, chronology & Rearranging scenes while preserving audio \\
 & \textcolor{GoogleRed}{Length}        & A+V & Summarization, expansion, highlight extraction & Producing a 30s highlight from a long clip \\
\midrule
\multirow{3}{=}{\textcolor{GoogleYellow!75!black}{\textbf{Generative}}}
 & \textcolor{GoogleYellow!75!black}{Text-to-AV}     & A+V & Text-to-video with audio, music video, talking head & Generating a music video from a prompt \\
 & \textcolor{GoogleYellow!75!black}{AV Inpainting}  & A+V & Masked region inpainting, occlusion, gap filling & Filling a missing segment in a clip \\
 & \textcolor{GoogleYellow!75!black}{AV Outpainting} & A+V & Temporal extension, spatial extension, continuation & Extending a clip beyond its original duration \\
\midrule
\multirow{4}{=}{\textcolor{GoogleGreen}{\textbf{Quality \& Restoration}}}
 & \textcolor{GoogleGreen}{Joint Restoration} & A+V & Joint denoising, archival restoration & Restoring old film with audio \\
 & \textcolor{GoogleGreen}{Visual Restoration} & V & Super-resolution, deblurring, color correction & Upscaling old footage \\
 & \textcolor{GoogleGreen}{Audio Restoration}  & A & Denoising, dereverberation, click removal & Cleaning a noisy dialogue track \\
 & \textcolor{GoogleGreen}{Sync Repair}        & A+V & Drift correction, lip sync repair, AV offset & Fixing audio that drifts out of sync \\
\midrule
\multirow{3}{=}{\textcolor{GoogleGray}{\textbf{Cross-Cutting}}}
 & \textcolor{GoogleGray}{Granularity}      & A+V & Frame, shot, scene, clip, with paired audio scales & Editing one frame vs the entire clip \\
 & \textcolor{GoogleGray}{Control Modality} & A+V & Text, reference clip, storyboard, parametric, trajectory & Prompting via natural language \\
 & \textcolor{GoogleGray}{Consistency}      & A+V & Cross-modal, temporal, identity, stylistic & Maintaining identity across an edit \\
\bottomrule
\end{tabular}
\end{table*}

\subsection{Synchronization}
\label{sec:jointedit-sync}
Synchronization edits alter timing rather than content: lip-sync correction, alignment, re-timing. Dubbing is the developed instance: EdiDub~\cite{manela2025edidub} re-synchronizes lips by content-aware mouth-region editing, JUST-DUB-IT~\cite{chen2026justdubit} jointly generates translated audio and synchronized facial motion via a low-rank adapter, and EditYourself~\cite{flynn2026edityourself} targets identity-preserving re-voicing; alignment and re-timing have no dedicated methods.

\subsection{Joint Content}
\label{sec:jointedit-content}
Insertion, removal, and replacement in both streams is the prototypical joint edit---a passing car arrives with its engine sound. AV-Edit~\cite{guo2026avedit} gates a masked-autoencoder-plus-DiT pipeline by audio-visual correlation; Object-AVEdit~\cite{fu2025objectavedit} reaches the same operations by inversion and regeneration. Still object-scoped: scene swaps and ambience-level insertions remain undemonstrated.

\subsection{Cross-Modal Transfer}
\label{sec:jointedit-transfer}
Deriving one stream from the other---re-voicing a portrait, re-soundtracking an edited video, stylizing both to a reference---runs the problems of Section~\ref{sec:crossgen} inside an editing pipeline; editing-specific is \emph{propagation}, an edit in one modality inducing the consistent change in the other automatically, and the bidirectional type is unrealized.

\subsection{Identity \& Performance}
\label{sec:jointedit-identity}
Coupled face-and-voice swaps, performance transfer, and joint emotion edits are reached today only through dubbing (JUST-DUB-IT~\cite{chen2026justdubit}, EditYourself~\cite{flynn2026edityourself}); the unimodal ingredients are mature, missing only the coupling that keeps a swapped face and converted voice the same person (cf.\ \S\ref{sec:open-ethics}).

\subsection{Scene \& Environment}
\label{sec:jointedit-scene}
Relocation, time-of-day and weather shifts, crowd changes, and acoustic coupling (reverberation inferred from depicted geometry) require audio edits proportional to the visual change; no method covered here targets these, the nearest machinery being ambience generation (\S\ref{sec:crossgen-v2a}).

\subsection{Narrative}
\label{sec:jointedit-narrative}
Cut-level edits---J- and L-cut splicing, reordering, summarization---are what professional editors do most, and no model covered here supports them: current methods edit within a shot, while narrative editing reasons across shots and modalities at once---the clearest open opportunity the taxonomy exposes.

\subsection{Generative}
\label{sec:jointedit-generative}
Joint inpainting and outpainting sit on the editing-generation boundary: the machinery is that of Section~\ref{sec:jointgen}, but fidelity to input binds outside the synthesized region; neither has a dedicated method in the literature we cover.

\subsection{Quality \& Restoration}
\label{sec:jointedit-restoration}
Joint denoising, per-stream restoration, and drift repair raise fidelity while changing nothing else; independent restoration can break correspondence, and sync repair after such processing has no learned joint treatment---well posed, demanded, unclaimed.

\subsection{Cross-Cutting}
\label{sec:jointedit-crosscutting}
Granularity, control modality, and consistency cut across all categories; control differentiates current methods. The most general instruction is text: language-guided joint editing~\cite{liang2024avedit} adapts a joint model to a single example so a textual edit propagates, and AvED~\cite{lin2026aved} obtains it zero-shot by delta-denoising both streams under frozen unimodal models.

\section{Joint Audio-Visual Generation}
\label{sec:jointgen}
Joint generation outputs both streams, $(\hat{v}, \hat{a}) \sim p_\theta(v, a \mid c)$ (Problem~\ref{prob:jointgen}); \emph{joint} names what is generated, not how. Table~\ref{tab:av-taxonomy} organizes the methods we cover, with the cross-modal and editing families, grouped by output.

\begin{sidewaystable*}
\centering
\caption{\textbf{Taxonomy of methods for joint audio-video generation and editing.}
Approaches are categorized by their \emph{output}: the first group outputs both modalities together, V$+$A (Sec.~\ref{sec:jointgen}); the second outputs a single modality conditioned on the other, A from V or V from A (Sec.~\ref{sec:crossgen}); and the third outputs a modified version of an existing pair, V$'+$A$'$ (Sec.~\ref{sec:jointedit}).
\textbf{Task}: Uncond. $=$ unconditional joint generation, T2AV $=$ text-to-audio-video, I2AV $=$ image-to-audio-video, Any2AV $=$ any-modality input, V2A $=$ video-to-audio, A2V $=$ audio-to-video, AVE $=$ joint audio-video editing, Dub $=$ joint dubbing.
\textbf{Inputs}: T (text), V (video), A (audio), I (image), A-ref (reference audio), instr.\ (instruction), trans.\ (translated transcript).
\textbf{Output}: V$+$A jointly generated, V$'+$A$'$ jointly edited.
\textbf{Backbone}: UNet, DiT (diffusion transformer~\cite{peebles2023dit}), Flow (rectified flow / flow matching).
\textbf{Train}: ZS (zero-shot), OS (one-shot), FT (fine-tune), SC (from-scratch), Adp (adapter-only), LoRA.
\textbf{Code}: \cmark\ open-source, \xmark\ not released. -- marks entries not applicable or not publicly disclosed (Google V2A, Wan 2.5).}
\label{tab:av-taxonomy}
\renewcommand{\arraystretch}{1.2}
\setlength{\tabcolsep}{3.5pt}
\scriptsize
\begin{tabular}{@{}lHllllllllc@{}}
\toprule
\textbf{Method} & \textbf{Year/Venue} & \textbf{Task} & \textbf{Inputs} & \textbf{Output} & \textbf{Backbone} & \textbf{Architecture} & \textbf{Conditioning} & \textbf{Alignment} & \textbf{Train} & \textbf{Code} \\
\midrule
\multicolumn{11}{@{}l}{\textit{\textbf{Joint Audio-Video Generation} (Sec.~\ref{sec:jointgen})}\quad---\quad output both streams: \ $(\hat{v}, \hat{a}) \sim p_\theta(v, a \mid c)$} \\
\midrule
MM-Diffusion~\cite{ruan2023mmdiffusion}     & CVPR 2023      & Uncond. & --         & V$+$A & Coupled UNet           & Sequential dual UNet            & Random-shift cross-attn          & Joint denoising                   & SC     & \cmark \\
CoDi~\cite{tang2023codi}                    & NeurIPS 2023   & Any2AV & T,V,A,I    & V$+$A & Latent UNet            & Composable diffusion            & Bridging encoders                & Cross-modal latents               & Adp     & \cmark \\
Seeing-and-Hearing~\cite{xing2024seeing}    & CVPR 2024      & T2AV   & T,V,A      & V$+$A & Frozen UNets           & Two single-modal models         & ImageBind aligner                & Latent classifier guidance        & ZS     & \cmark \\
AV-DiT~\cite{wang2024avdit}                 & NeurIPS-W 24   & T2AV   & T          & V$+$A & Shared DiT             & Single DiT, two heads           & Lightweight adapters             & Shared self-attn                  & FT     & \xmark \\
MM-LDM~\cite{sun2024mmldm}                  & ACM MM 2024    & T2AV   & T          & V$+$A & Latent UNet            & Hierarchical latent             & Hierarchical multi-modal         & Shared latent                     & SC     & \xmark \\
Movie Gen~\cite{polyak2024moviegen}         & arXiv 2024     & T2AV   & T,I        & V$+$A & DiT (Flow)             & Cascaded T2V$\to$V2A            & Cascaded conditioning            & Cascaded                          & SC     & \xmark \\
SVG~\cite{ishii2024svg}                     & arXiv 2024     & T2AV   & T          & V$+$A & Two pretrained DiTs    & Adapted dual-tower              & Lightweight bridging             & Cross-modal exchange              & FT     & \xmark \\
MMDisCo~\cite{hayakawa2025mmdisco}          & ICLR 2025      & T2AV   & T          & V$+$A & Two UNets              & Frozen $+$ joint discriminator  & Discriminator guidance           & Adversarial alignment             & FT     & \cmark \\
SyncFlow~\cite{liu2024syncflow}             & arXiv 2024     & T2AV   & T          & V$+$A & Dual DiT (Flow)        & d-DiT, decoupled multi-stage    & Text on both branches            & Joint fine-tune                   & SC     & \xmark \\
JavisDiT~\cite{liu2025javisdit}             & ICLR 2026      & T2AV   & T          & V$+$A & Joint DiT              & AV-DiT with ST cross-attn       & HiST-Sypo prior                  & Hierarchical ST attn              & SC     & \cmark \\
JavisDiT++~\cite{liu2026javisditpp}         & ICLR 2026      & T2AV   & T          & V$+$A & Joint DiT (MS-MoE)     & Dual-branch with MS-MoE         & HiST prior $+$ TA-RoPE           & Frame-level TA-RoPE, AV-DPO       & SC     & \cmark \\
BridgeDiT~\cite{guan2025bridgedit}          & arXiv 2025     & T2AV   & T          & V$+$A & DiT                    & Dual-tower with bridge          & Decoupled $T_V/T_A$ captions     & Bidirectional bridge              & FT     & \cmark \\
ALIVE~\cite{guo2026alive}                   & arXiv 2026     & T2AV   & T,I        & V$+$A & DiT                    & Dual$+$single stream            & TA-CrossAttn $+$ UniTemp-RoPE    & Strict temporal RoPE              & FT     & \xmark \\
Ovi~\cite{low2025ovi}                       & arXiv 2025     & T2AV   & T          & V$+$A & Twin DiT               & Twin backbones, cross fusion    & Cross-modal fusion               & Symmetric fusion                  & SC     & \cmark \\
UniAVGen~\cite{zhang2025uniavgen}           & arXiv 2025     & T2AV   & T          & V$+$A & Joint DiT              & Dual-branch parallel DiT        & Asym.\ cross-modal interaction   & Face-aware modulation, MA-CFG     & SC     & \cmark \\
Animate-and-Sound~\cite{wang2025jointdit}   & CVPR 2025      & I2AV   & I          & V$+$A & Dual-tower DiT         & Decompose $+$ expert blocks     & Image-conditioned                & Mutual influence                  & FT     & \xmark \\
CCL~\cite{ma2026ccl}                        & arXiv 2026     & T2AV   & T          & V$+$A & Dual-stream DiT        & Cross-modal context learning    & Decoupled cross-modal context    & Context alignment                 & FT     & \xmark \\
Hallo-Live~\cite{li2026hallolive}           & arXiv 2026     & I2AV   & I,A        & V$+$A & Dual-stream DiT        & Async.\ dual-stream, streaming  & Future-expanding attn            & Streaming lip-sync, HP-DMD        & FT     & \cmark \\
UniForm~\cite{zhao2025uniform}              & arXiv 2025     & Any2AV & T,V,A      & V$+$A & Multi-task DiT         & Shared denoiser, task tokens    & Task tokens                      & Shared latent                     & SC     & \xmark \\
Wan 2.5~\cite{wan25_2025}                   & Alibaba 2025   & T2AV   & T,I,A      & V$+$A & DiT                    & Multilingual joint              & T5 $+$ audio $+$ image           & --                                & --     & \xmark \\
LTX-2~\cite{hacohen2026ltx2}                & arXiv 2026     & T2AV   & T          & V$+$A & Asym.\ dual DiT (14B$+$5B) & Bidir.\ cross-attn          & Modality-CFG, AdaLN              & Bidir.\ cross-attn                & SC     & \cmark \\
MOVA~\cite{mova2026}                        & arXiv 2026     & T2AV   & T          & V$+$A & DiT                    & Open joint model                & Multi-track conditioning         & End-to-end joint                  & SC     & \cmark \\
Apollo~\cite{wang2026klear}                  & arXiv 2026     & T2AV   & T          & V$+$A & Single-tower MM-DiT    & Omni-Full Attention             & Progressive multi-task           & Tight AV alignment                & SC     & \xmark \\
3MDiT~\cite{li2025threemdit}                & arXiv 2025     & T2AV   & T          & V$+$A & Tri-modal DiT          & Isomorphic A/V branches         & Trimodal omni-blocks             & Dynamic text $+$ AV co-evolve     & SC/FT  & \xmark \\
OmniForcing~\cite{su2026omniforcing}        & arXiv 2026     & T2AV   & T          & V$+$A & Streaming DiT          & Causal AR distilled from LTX-2  & Distilled bidirectional          & Streaming sync                    & FT     & \cmark \\
\bottomrule
\end{tabular}
\end{sidewaystable*}

\begin{sidewaystable*}
\ContinuedFloat
\centering
\caption{\textbf{Taxonomy of methods for joint audio-video generation and editing} (continued).}
\renewcommand{\arraystretch}{1.2}
\setlength{\tabcolsep}{3.5pt}
\scriptsize
\begin{tabular}{@{}lHllllllllc@{}}
\toprule
\textbf{Method} & \textbf{Year/Venue} & \textbf{Task} & \textbf{Inputs} & \textbf{Output} & \textbf{Backbone} & \textbf{Architecture} & \textbf{Conditioning} & \textbf{Alignment} & \textbf{Train} & \textbf{Code} \\
\midrule
\multicolumn{11}{@{}l}{\textit{\textbf{Cross-Modal Generation} (Sec.~\ref{sec:crossgen})}\quad---\quad output one stream given the other: \ $\hat{a} \sim p_\theta(a \mid v)$ \ or \ $\hat{v} \sim p_\theta(v \mid a)$} \\
\midrule
Diff-Foley~\cite{luo2023difffoley}            & NeurIPS 2023   & V2A     & V             & A             & Latent UNet              & CAVP $+$ latent diffusion            & --                         & Contrastive AV pretraining         & SC  & \cmark \\
Foley Analogies~\cite{iyer2023foleyanalogies} & CVPR 2023      & V2A     & V,A-ref       & A             & Latent diffusion         & Reference-conditioned Foley          & Audio reference            & Onset transfer from exemplar       & SC  & \cmark \\
V2A-Mapper~\cite{lin2024v2amapper}            & AAAI 2024      & V2A     & V             & A             & Frozen foundation        & Lightweight vision-audio mapper      & --                         & Foundation-model embedding match   & Adp  & \xmark \\
Video-Foley~\cite{lee2024videofoley}          & TASLP 2025     & V2A     & V,T,A-ref     & A             & Latent UNet              & RMS two-stage control                & Text $+$ audio ref         & RMS envelope conditioning          & Adp  & \cmark \\
FoleyCrafter~\cite{zhang2024foleycrafter}     & IJCV 2026      & V2A     & V,T           & A             & Latent UNet              & Semantic adapter $+$ temporal ctrl.  & Text                       & Temporal controller                & Adp  & \cmark \\
Frieren~\cite{wang2024frieren}                & NeurIPS 2024   & V2A     & V             & A             & Flow (RF)                & Rectified flow matching              & --                         & Onset-aligned flow                 & FT  & \cmark \\
MaskVAT~\cite{garoufis2024maskvat}            & ECCV 2024      & V2A     & V             & A             & Masked transformer       & Masked generative transformer        & --                         & Enhanced synchronicity             & SC  & \xmark \\
STA-V2A~\cite{chen2024stav2a}                 & arXiv 2024     & V2A     & V,T           & A             & Latent UNet              & Local $+$ global visual features     & Text                       & Semantic $+$ temporal alignment    & FT  & \cmark \\
VATT~\cite{liu2024vatt}                       & NeurIPS 2024   & V2A     & V,T           & A             & Latent UNet              & Caption-mediated generation          & Text                       & Caption-level semantic match       & SC/LoRA  & \cmark \\
Google V2A~\cite{deepmind2024v2a}             & DeepMind 2024  & V2A     & V,T           & A             & Latent diffusion         & Prompt-conditioned diffusion         & Text                       & Onset conditioning                 & --  & \xmark \\
MMAudio~\cite{cheng2025mmaudio}               & CVPR 2025      & V2A     & V,T           & A             & Flow (RF)                & Multimodal joint training            & Text                       & Dedicated synchronization module   & SC  & \cmark \\
Mel-QCD~\cite{wang2025melqcd}                 & CVPR 2025     & V2A     & V,T           & A             & Latent UNet              & Mel decomposition $+$ ControlNet     & Text                       & Mel quantization-continuum         & Adp  & \cmark \\
VAFlow~\cite{wang2025vaflow}                  & ICCV 2025      & V2A     & V             & A             & Flow (RF)                & Cross-modality flow matching         & --                         & Cross-modal flow coupling          & SC  & \xmark \\
Foley-Flow~\cite{mo2025foleyflow}             & CVPR 2025      & V2A     & V             & A             & Flow (RF)                & Masked AV align $+$ dynamic flow     & --                         & Masked audio-visual alignment      & SC  & \xmark \\
MultiFoley~\cite{chen2025multifoley}          & CVPR 2025      & V2A     & V,T,A-ref     & A             & DiT                      & Multi-conditional training           & Text $+$ audio ref         & Onset alignment                    & SC  & \xmark \\
TARO~\cite{zhang2025taro}                     & ICCV 2025      & V2A     & V             & A             & DiT                      & Timestep-adaptive repr.\ alignment   & --                         & Onset-aware conditioning           & SC  & \cmark \\
ThinkSound~\cite{liu2025thinksound}           & NeurIPS 2025   & V2A     & V,T,mask      & A             & DiT                      & MLLM chain-of-thought                & Text $+$ click/mask        & Reasoned event placement           & FT  & \cmark \\
Hear-Your-Click~\cite{guo2025hearyourclick}   & arXiv 2025     & V2A     & V,T,click     & A             & DiT                      & Object-centric conditioning          & Click/mask                 & Object-level onset                 & FT  & \cmark \\
SelVA~\cite{anonymous2025selva}               & CVPR 2026      & V2A     & V,T           & A             & DiT                      & Text-conditioned selective V2A       & Text                       & Selective source onset             & FT  & \cmark \\
SoundReactor~\cite{saito2025soundreactor}     & arXiv 2025     & V2A     & V             & A             & Causal AR $+$ diff.\ head & Frame-level online generation       & --                         & Streaming frame-level sync         & SC  & \xmark \\
Foley-Omni~\cite{tao2026foleyomni}            & arXiv 2026     & V2A     & V,T           & A             & DiT                      & Unified task $\to$ full soundtrack   & Text                       & Multi-track soundtrack alignment   & SC  & \cmark \\
AV-Link~\cite{hajiali2025avlink}              & ICCV 2025      & V2A/A2V & V \emph{or} A & A \emph{or} V & Flow (frozen)            & Frozen backbones $+$ feature links   & Text $+$ audio ref         & Temporally-aligned diff.\ features  & Adp  & \xmark \\
\midrule
\multicolumn{11}{@{}l}{\textit{\textbf{Joint Audio-Video Editing} (Sec.~\ref{sec:jointedit})}\quad---\quad output a modified pair: \ $(v', a') \sim p_\theta(v', a' \mid v, a, e)$} \\
\midrule
Lang.-Guided AV Edit~\cite{liang2024avedit} & ACCV 2024      & AVE    & V,A,T          & V$'+$A$'$ & Joint AV diffusion & One-shot LoRA adaptation        & Text $+$ paired AV               & Cross-modal sem.\ enhancement     & OS     & \xmark \\
AvED~\cite{lin2026aved}                     & WACV 2026      & AVE    & V,A,T          & V$'+$A$'$ & Frozen latent UNets & Cross-modal delta denoising     & Text prompts                     & Patch-level AV delta align.       & ZS     & \cmark \\
EdiDub~\cite{manela2025edidub}              & arXiv 2025     & Dub    & V,A,mask       & V$'$      & 3D UNet (diff.)    & Two-stage content-aware edit    & Quantized HuBERT audio           & AdaIN audio modulation            & SC     & \xmark \\
Object-AVEdit~\cite{fu2025objectavedit}     & arXiv 2025     & AVE    & V,A,T          & V$'+$A$'$ & Mochi-1 $+$ audio DiT & Inversion-regeneration       & Source/target prompts            & --                                & SC/ZS  & \xmark \\
AV-Edit~\cite{guo2026avedit}                & AAAI 2026      & AVE    & V,A,T-instr.   & A$'$      & MM-DiT             & CAV-MAE-Edit $+$ MM-DiT         & AV semantic control              & AV correlation gating             & FT     & \xmark \\
JUST-DUB-IT~\cite{chen2026justdubit}        & arXiv 2026     & Dub    & V,A,T-trans.   & V$'+$A$'$ & Joint AV diffusion & LoRA on AV foundation           & Audio $+$ video joint cond.      & Joint AV prior                    & LoRA   & \cmark \\
EditYourself~\cite{flynn2026edityourself}   & arXiv 2026     & AVE    & V,A,script     & V$'$      & DiT                & Audio-conditioned V2V           & Audio $+$ region masks           & Identity-preserving lip-sync      & FT     & \xmark \\
\bottomrule
\end{tabular}
\end{sidewaystable*}

\subsection{Unconditional Joint Generation}
\label{sec:jointgen-unconditional}
The case $c = \emptyset$ exposes the dependency most directly; coupled diffusion over a paired latent, as in MM-Diffusion~\cite{ruan2023mmdiffusion}, is the canonical instance.

\subsection{Text-to-Audio-Visual Generation}
\label{sec:jointgen-text}
Text-to-audio-visual generation samples $(\hat{v}, \hat{a}) \sim p_\theta(v, a \mid c_t)$ depicting the prompt in both streams. Methods span the taxonomy: dual-tower DiTs coupled through cross-attention (JavisDiT~\cite{liu2025javisdit}, extended with modality-specific experts and aligned rotary encodings~\cite{liu2026javisditpp}); cross-modal context learning~\cite{ma2026ccl}; twin-backbone fusion (Ovi~\cite{low2025ovi}) and asymmetric interaction (UniAVGen~\cite{zhang2025uniavgen}) for lip sync and timbre; open systems scaling joint training (MOVA~\cite{mova2026}, Apollo~\cite{wang2026klear}); and partially documented systems (Wan~2.5~\cite{wan25_2025}), left unmarked on undisclosed axes in Table~\ref{tab:av-design-taxonomy}.

\subsection{Image-to-Audio-Visual Generation}
\label{sec:jointgen-image}
Conditioning on an image pins identity and layout, shifting the difficulty to motion and sound consistent with a fixed first frame; sharing an expert block between video and audio branches, as in Animate-and-Sound~\cite{wang2025jointdit}, is representative.

\subsection{Talking-Head and Speech-Driven Generation}
\label{sec:jointgen-talkinghead}
Talking-head generation outputs a speaking face with its speech track from an identity image and driving speech; correspondence reduces to single-frame lip synchronization. Hallo-Live~\cite{li2026hallolive} reaches streaming avatars by attending to a short horizon of future phonetic cues.

\subsection{Long-Form Joint Generation}
\label{sec:jointgen-long}
Long-form generation adds coherence over minutes---errors accumulate, and identity, scene, and the audio-visual relationship must not drift; streaming formulations such as OmniForcing~\cite{su2026omniforcing} generate in causal blocks while distilling from a bidirectional teacher.

\section{Cross-Modal Generation}
\label{sec:crossgen}
Cross-modal generation outputs exactly one modality given the other (Problem~\ref{prob:crossgen}): the input is observed and fixed, and the output must be made consistent with it.

\subsection{Video-to-Audio Generation}
\label{sec:crossgen-v2a}
Video-to-audio, $\hat{a} \sim p_\theta(a \mid v)$, is by far the most developed cross-modal setting: onsets must land at the exact frames of visual contact. One durable strategy learns correspondence before generating (Diff-Foley's contrastive pretraining~\cite{luo2023difffoley}, STA-V2A~\cite{chen2024stav2a}, VATT's caption route~\cite{liu2024vatt}, V2A-Mapper's frozen-model bridge~\cite{lin2024v2amapper}); a second adopts flow matching for faster sampling and tighter synchronization (Frieren~\cite{wang2024frieren}, VAFlow~\cite{wang2025vaflow}, Foley-Flow~\cite{mo2025foleyflow}, MMAudio~\cite{cheng2025mmaudio}), with masked and causal token models alongside (MaskVAT~\cite{garoufis2024maskvat}, SoundReactor~\cite{saito2025soundreactor}); a third attaches control to a fixed generator (FoleyCrafter~\cite{zhang2024foleycrafter}, Video-Foley~\cite{lee2024videofoley}, Mel-QCD~\cite{wang2025melqcd}, MultiFoley~\cite{chen2025multifoley}, TARO~\cite{zhang2025taro}). Control has lately moved to instructions (ThinkSound~\cite{liu2025thinksound}, Hear-Your-Click~\cite{guo2025hearyourclick}, SelVA~\cite{anonymous2025selva}); Foley-Omni~\cite{tao2026foleyomni} folds speech, effects, and music into one generator, and closed systems such as Google's V2A~\cite{deepmind2024v2a} disclose little.

\subsection{Audio-to-Video Generation}
\label{sec:crossgen-a2v}
Audio-to-video, $\hat{v} \sim p_\theta(v \mid a)$, is structurally harder: one track licenses many videos. Dedicated attempts are isolated (\citet{yariv2024tempotokens} adapt a frozen text-to-video model and introduce AV-Align); otherwise the direction is supported only incidentally, by MM-Diffusion's joint distribution~\cite{ruan2023mmdiffusion}, Seeing-and-Hearing's direction-indifferent aligner~\cite{xing2024seeing}, UniAVGen's task list~\cite{zhang2025uniavgen}, and AV-Link's bidirectional linking~\cite{hajiali2025avlink}; even video-to-music (\S\ref{sec:align-rhythm}) runs almost entirely the other way (\S\ref{sec:open-designspace}).

\subsection{Speech-to-Video Generation}
\label{sec:crossgen-s2v}
Speech-to-video outputs only the visual stream from speech and an identity image; correspondence reduces to the viseme-phoneme match. Portrait animation dominates---speech-to-gesture~\cite{ginosar2019gesture}, animators predicting 3D coefficients~\cite{zhang2023sadtalker} or denoising video directly~\cite{tian2024emo,xu2024hallo}, extended to full figures~\cite{corona2025vlogger,lin2025omnihuman}---while beyond the portrait the mapping is radically one-to-many and largely untouched (\S\ref{sec:open-designspace}).

\subsection{Foley and Sound-Effect Generation from Video}
\label{sec:crossgen-foley}
Foley narrows video-to-audio to the diegetic sound of visible actions, where timing is tightest---a footstep a few frames off is wrong, not degraded---and where practice demands control over which sources sound, when, and how loud: exemplar transfer~\cite{iyer2023foleyanalogies}, envelope control~\cite{lee2024videofoley}, multi-signal conditioning~\cite{chen2025multifoley}, and user selection~\cite{guo2025hearyourclick,anonymous2025selva}.

\section{Alignment and Synchronization}
\label{sec:align}
Alignment is the property that ties together every setting in this work, and we treat it once here rather than repeating it in each section. It is the operational form of Definition~\ref{def:correspondence}: a pair $(v, a)$ is aligned when embeddings from a contrastively trained encoder pair $f_v:\mathcal{V}\to\mathbb{R}^{d_e}$ and $f_a:\mathcal{A}\to\mathbb{R}^{d_e}$---distinct from the generative encoders $(\phi_v,\phi_a)$ of Definition~\ref{def:latent}---are close under cosine similarity, either globally or per time step.

\begin{definition}[Temporal synchronization]
\label{def:sync}
A pair $(v, a)$ is temporally synchronized at tolerance $\delta$ when, for each time index $t$, the visual event at frame $v^{(t)}$ corresponds to an audio event within $[t - \delta, t + \delta]$ of $a$; lip sync (mouth shape vs.\ phoneme) and beat alignment (motion peak vs.\ beat) are its special cases.
\end{definition}

Definition~\ref{def:sync} makes precise the temporal component of correspondence (Definition~\ref{def:correspondence}); the score $\mathcal{S}$ itself is operationalized by the learned encoders above, and the subsections below organize methods by the kind of event they align and the tolerance they target.

\subsection{Temporal Synchronization}
\label{sec:align-temporal}

General temporal synchronization places arbitrary acoustic events at the frames of their visual cause, the requirement underlying video-to-audio and joint generation alike, and the tolerance $\delta$ of Definition~\ref{def:sync} that a method achieves is the primary measure of its temporal quality. Methods reach it in three broadly different places. Some supply alignment through a representation learned in advance, as in the contrastive audio-visual pretraining of Diff-Foley~\cite{luo2023difffoley}, so that the generator inherits correspondence rather than enforcing it. Others add machinery dedicated to timing: MMAudio~\cite{cheng2025mmaudio} attaches an explicit synchronization module, TARO~\cite{zhang2025taro} conditions on onsets while adapting representation alignment across timesteps, and MaskVAT~\cite{garoufis2024maskvat} targets synchronicity directly in a masked token model. A third group builds it into the architecture, either through cross-attention between the two streams---the dominant choice, adopted by nineteen of the methods in Table~\ref{tab:av-design-taxonomy}---or through a shared temporal encoding that forces tokens at the same physical time into correspondence, as in JavisDiT++~\cite{liu2026javisditpp} and ALIVE~\cite{guo2026alive}. Approaches that impose alignment only at inference, whether by discriminator~\cite{hayakawa2025mmdisco} or by classifier guidance~\cite{xing2024seeing}, are now the exception, which is itself evidence that synchronization has migrated from a post-hoc correction into the model.

\subsection{Semantic Alignment}
\label{sec:align-semantic}

Semantic alignment is the weaker, global property that the sources in $v$ and $a$ match in identity even when their timing is loose. It is necessary but not sufficient for correspondence: a clip whose sources agree but whose events are misplaced still feels out of step. In practice it is operationalized through contrastive audio-visual embeddings, which several methods reuse directly as a guidance term---Seeing-and-Hearing~\cite{xing2024seeing} steers two frozen unimodal generators toward agreement using an ImageBind~\cite{girdhar2023imagebind} aligner---or as a training signal. An alternative is to route the alignment through language: VATT~\cite{liu2024vatt} captions the video and generates audio from the caption, which makes the semantic link explicit and controllable at the cost of the temporal precision that a direct visual conditioning path preserves. Recent benchmarks suggest this is where current models are weakest, with AVGen-Bench~\cite{zhou2026avgenbench} reporting a gap between strong audio-visual aesthetics and unreliable semantic grounding.

\subsection{Lip-Sync and Phoneme-Level Alignment}
\label{sec:align-lipsync}

Lip synchronization is the most demanding instance of Definition~\ref{def:sync}: the tolerance is on the order of a single frame, and viewers detect phoneme-to-viseme mismatch far more readily than any other misalignment. It is the binding constraint in talking-head generation, speech-to-video, and dubbing, and methods in those families are organized around it rather than merely evaluated on it. UniAVGen~\cite{zhang2025uniavgen} introduces face-aware modulation for exactly this purpose; Hallo-Live~\cite{li2026hallolive} lets each generated video block attend to a short horizon of future phonetic cues so that streaming generation does not sacrifice lip accuracy; and in the editing setting, JUST-DUB-IT~\cite{chen2026justdubit} and EditYourself~\cite{flynn2026edityourself} must satisfy the same constraint while preserving speaker identity and the untouched regions of the source clip. Because human sensitivity here is unusually sharp, this is also the sub-problem with the most established automatic metrics, and the one where they agree best with human judgment.

\subsection{Rhythmic and Beat-Level Alignment}
\label{sec:align-rhythm}
For music, alignment is rhythmic rather than phonetic: motion peaks should coincide with musical beats. The relevant event is periodic, which makes the alignment both easier to measure and easier to violate in a way that is immediately noticeable. Most work in this setting runs from video to music rather than the reverse, generating a soundtrack whose beat structure follows observed motion. Early approaches tie note onsets to body movement in instrument performance~\cite{gan2020foleymusic,su2020audeo} and to human motion more generally~\cite{gan2021rhythmic}, while later work targets background music for arbitrary video with explicit rhythmic control~\cite{di2021bgmtransformer,di2023vbgm} and extends the horizon over which rhythm must remain coherent~\cite{zhu2023soundtracker}. Dance video, where the motion is already organized around a beat, is the most constrained instance~\cite{zhu2022quantizedgan}, and the paired dance-and-music corpora built for it~\cite{li2021aist} are the standard evaluation setting. The reverse direction, generating video whose motion follows a given piece of music, remains comparatively unexplored, an instance of the broader asymmetry discussed in Section~\ref{sec:crossgen-a2v}.

\subsection{Consistency Across Long Sequences}
\label{sec:align-longconsistency}

Over long horizons, alignment must be maintained as well as achieved, since small per-step errors accumulate into visible and audible drift. This makes long-form coherence a distinct problem rather than an extension of short-clip synchronization, and it connects this section to the long-form generation problem of Section~\ref{sec:jointgen-long}. Two families of solution have emerged. Streaming and causal formulations generate in blocks while distilling from a bidirectional teacher, as in OmniForcing~\cite{su2026omniforcing}, or attach a diffusion head to a causal transformer to produce audio frame by frame under an online latency budget, as in SoundReactor~\cite{saito2025soundreactor}. Alternatively, methods extend the generation window directly, whether through the long-form conditioning of MultiFoley~\cite{chen2025multifoley} and Movie Gen's audio branch~\cite{polyak2024moviegen} or through architectures aimed at unbounded generation~\cite{ergasti2025rflav}. Both remain evaluated on horizons far shorter than the minutes-long content the applications of Section~\ref{sec:applications} assume, which we return to in Section~\ref{sec:open}.

\section{Architectures and Training Strategies}
\label{sec:arch}
Having defined the tasks, we turn to the model families used to instantiate $p_\theta$. A cascaded model factorizes the joint distribution as $p_\theta(v, a \mid c) = p_{\theta_1}(v \mid c)\, p_{\theta_2}(a \mid v, c)$, or the reverse, producing one modality first and the other conditioned on it (\S\ref{sec:strategy-cascade}); a two-tower model instead parameterizes $p_\theta(v, a \mid c)$ jointly, denoising both streams in parallel with the coupling carried by $\theta_\times$ (\S\ref{sec:strategy-dual}). A unified backbone learns $p_\theta(v, a \mid c)$ directly with a single network that processes both modalities through shared parameters. Within any of these, diffusion and flow-matching approaches parameterize $p_\theta$ through a denoising process applied to $v$, $a$, or both, while autoregressive approaches tokenize the two modalities and model them as a single sequence. Architecture and task interact in predictable ways: cascaded models are common in cross-modal generation, where one modality is observed and the other is conditioned on it, whereas unified backbones are more common in joint generation, where both modalities are outputs of a shared distribution.
Autoregressive discrete-token modeling, natural for streaming, remains unoccupied (\S\ref{sec:strategy}).

\subsection{Two-Tower and Cascaded Models}
\label{sec:arch-cascaded}
Two-tower and cascaded models keep the modalities in separate networks and couple them through cross-modal connections or through a generation order. They can reuse strong unimodal backbones and add only the coupling, which makes them data-efficient at the cost of a coordination burden between the towers.

\subsection{Unified Multimodal Backbones}
\label{sec:arch-unified}
Unified backbones process both modalities with shared parameters, either as one fused sequence or as a single network with modality-specific heads. They capture the dependency between $v$ and $a$ most directly and are the basis of most recent joint models, at the cost of a representation that must serve both modalities at once.

\subsection{Diffusion-Based Approaches}
\label{sec:arch-diffusion}
Diffusion and flow-matching approaches dominate both modalities. Joint variants apply the denoising process to a paired latent, with the coupling realized through shared layers or cross-attention, and they inherit the controllability of classifier-free guidance directly.

\subsection{Autoregressive and Token-Based Approaches}
\label{sec:arch-autoregressive}
Autoregressive and masked token models treat the two modalities as one sequence of discrete tokens, which makes streaming and variable-length generation natural and supports a single backbone across multiple tasks through input reordering.

\subsection{Training Objectives and Losses}
\label{sec:arch-objectives}
Beyond the per-modality reconstruction or denoising loss, joint methods add objectives that target correspondence directly, including contrastive alignment losses, adversarial joint-realism losses, and preference objectives that reward synchronization. The choice of objective is the training-time counterpart of the alignment-enforcement axis of Section~\ref{sec:sync}.

\section{Datasets and Benchmarks}
\label{sec:datasets}
Progress in the area is driven by the data used to train and evaluate the methods of the previous sections.

\begin{definition}[Audio-visual dataset]
\label{def:dataset}
A generation dataset is a collection $\mathcal{D} = \{(v_n, a_n, c_n)\}_{n=1}^{N}$ of paired video, audio, and optional conditioning signals, where $c_n$ is typically a caption describing both modalities for joint generation and is empty for cross-modal generation. An editing dataset takes the richer form $\mathcal{D} = \{(v_n, a_n, e_n, v'_n, a'_n)\}_{n=1}^{N}$, where $e_n$ is an edit instruction and $(v'_n, a'_n)$ is the target pair.
\end{definition}

\begin{table*}[t!]
\centering
\scriptsize
\renewcommand{\arraystretch}{1.2}
\caption{\textbf{Representative datasets for joint and cross-modal audio-visual generation.} We group datasets by domain and report approximate scale and whether text captions are available (\textbf{Cap.}). The rightmost column lists the task each dataset most directly supports. Scales are approximate and refer to the commonly used release.}
\label{tab:datasets}
\setlength{\tabcolsep}{6pt}
\begin{tabular}{@{}llllcl@{}}
\toprule
\textbf{Dataset} & \textbf{Year} & \textbf{Domain} & \textbf{Approx.\ Scale} & \textbf{Cap.} & \textbf{Primary Task} \\
\midrule
AudioSet~\cite{gemmeke2017audioset}       & 2017 & in-the-wild events        & $\sim$2M clips, 10s each   & \xmark & AV pretraining, V2A \\
VGGSound~\cite{chen2020vggsound}          & 2020 & in-the-wild events        & $\sim$200k clips, 10s each & \xmark & V2A, joint generation \\
Kinetics~\cite{kay2017kinetics}           & 2017 & human actions             & $\sim$650k clips           & \xmark & AV pretraining \\
Greatest Hits~\cite{owens2016visually}    & 2016 & object impacts            & $\sim$1k videos            & \xmark & Foley / impacts \\
MUSIC~\cite{zhao2018soundofpixels}        & 2018 & instrument solos/duets    & 714 videos                 & \xmark & music V2A, separation \\
URMP~\cite{li2018urmp}                     & 2019 & classical ensembles       & 44 multi-track pieces      & \xmark & music V2A, separation \\
AIST++~\cite{li2021aist}                   & 2021 & dance with music          & 1408 seq., 1.1M frames     & \xmark & music-to-motion/video \\
AVSpeech~\cite{ephrat2018avspeech}        & 2018 & talking faces             & thousands of hours         & \xmark & speech-driven, separation \\
VoxCeleb2~\cite{chung2018voxceleb2}       & 2018 & talking faces             & $>$1M utterances           & \xmark & talking-head, identity \\
TAVGBench~\cite{mao2024tavgbench}         & 2024 & in-the-wild audible video & $\sim$1.7M clips, 11.8k h  & \cmark & T2AV training and evaluation \\
MMTrail~\cite{chi2024mmtrail}             & 2024 & trailers with music       & $>$20M clips (2M mm-captioned) & \cmark & music-video generation \\
JavisBench~\cite{liu2025javisdit}         & 2025 & open-domain sounding video& 10{,}140 captioned clips   & \cmark & T2AV evaluation \\
\bottomrule
\end{tabular}
\end{table*}

\begin{table*}[t!]
\centering
\scriptsize
\renewcommand{\arraystretch}{1.2}
\caption{\textbf{Recent benchmarks for joint audio-visual generation.} Each benchmark fixes a prompt set and an evaluation protocol; the columns name the task targeted and the property tested.}
\label{tab:benchmarks}
\setlength{\tabcolsep}{6pt}
\begin{tabular}{@{}lll@{}}
\toprule
\textbf{Benchmark} & \textbf{Task} & \textbf{Property Tested} \\
\midrule
JavisBench~\cite{liu2025javisdit}    & T2AV       & quality and synchronization in diverse scenes \\
SAVGBench~\cite{shimada2024savgbench}& joint gen. & spatial alignment between first-order-ambisonics audio and video \\
AVGen-Bench~\cite{zhou2026avgenbench}& T2AV       & aesthetics vs.\ semantic reliability (text, speech, physics, music) \\
AV-Phys Bench~\cite{cui2026avphys}   & joint gen. & physical commonsense across steady and transition scenes \\
\bottomrule
\end{tabular}
\end{table*}

\subsection{Audio-Visual Generation Datasets}
\label{sec:datasets-gen}
Generation datasets pair video with audio and, increasingly, with captions that describe both streams; Table~\ref{tab:datasets} lists representative instances. The scale, domain, and caption quality of $\mathcal{D}$ bound what a model can learn about correspondence, and recent collections emphasize captions that describe the audio-visual relationship rather than either stream alone.

\subsection{Audio-Visual Generation Benchmarks}
\label{sec:datasets-edit}
Benchmarks fix a prompt set and an evaluation protocol so that methods can be compared on the same footing; Table~\ref{tab:benchmarks} summarizes recent ones. Recent task-driven benchmarks such as AVGen-Bench~\cite{zhou2026avgenbench} evaluate text-to-audio-video generation at multiple granularities and expose a gap between strong audio-visual aesthetics and weak semantic reliability, while physically grounded benchmarks such as AV-Phys Bench~\cite{cui2026avphys} probe whether joint models respect the physics linking a visual event to its sound. No shared benchmark for joint audio-visual editing exists: each editing method covered here evaluates on data it assembled or repurposed itself, and none of these sets provides the $(v, a, e, v', a')$ supervision that Definition~\ref{def:dataset} defines for an editing dataset---a gap that makes editing results mutually incomparable today.

\section{Evaluation Metrics}
\label{sec:eval}
Assessing the outputs of joint and cross-modal models requires measures that capture both per-modality quality and cross-modal consistency. For a generated video $\hat{v}$, a quality metric $Q_v(\hat{v})$, or $Q_v(\hat{v}, v)$ when a reference is available, scores visual fidelity. For a generated audio $\hat{a}$, a metric $Q_a(\hat{a})$ scores audio fidelity. For a generated pair, an alignment metric $A(\hat{v}, \hat{a})$ scores cross-modal consistency, which neither $Q_v$ nor $Q_a$ alone captures. Edited pairs require two further measures: a faithfulness metric that assesses whether the edit instruction $e$ was applied, and a preservation metric that assesses whether content outside the edit region was left intact. Table~\ref{tab:metrics} organizes the metrics in use by what they measure.

\begin{table*}[t!]
\centering
\scriptsize
\renewcommand{\arraystretch}{1.2}
\caption{\textbf{Evaluation metrics organized by what they measure.} Per-modality quality metrics score one stream in isolation; condition-alignment metrics score agreement with the input $c$; cross-modal alignment metrics realize the score $\mathcal{S}$ of Def.~\ref{def:correspondence}; and editing metrics score the two competing requirements of Problem~\ref{prob:jointedit}. \textbf{Modality} indicates the streams compared (V video, A audio, T text), and \textbf{Better} the preferred direction.}
\label{tab:metrics}
\setlength{\tabcolsep}{6pt}
\begin{tabular}{@{}lllc@{}}
\toprule
\textbf{Metric} & \textbf{Measures} & \textbf{Modality} & \textbf{Better} \\
\midrule
\multicolumn{4}{@{}l}{\textit{\textbf{Per-modality quality}}} \\
FID~\cite{heusel2017fid}, FVD~\cite{unterthiner2018fvd} & visual fidelity (distribution distance)   & V     & $\downarrow$ \\
Inception Score~\cite{salimans2016improved} & visual quality and diversity              & V     & $\uparrow$ \\
FAD~\cite{kilgour2019fad}                & audio fidelity (distribution distance)    & A     & $\downarrow$ \\
KL (audio classifier)    & audio semantic match                      & A     & $\downarrow$ \\
\midrule
\multicolumn{4}{@{}l}{\textit{\textbf{Condition alignment}}} \\
CLIPScore~\cite{radford2021clip,hessel2021clipscore}     & text-to-visual agreement                  & V/T  & $\uparrow$ \\
CLAP score~\cite{wu2023clap}             & text-to-audio agreement                   & A/T  & $\uparrow$ \\
\midrule
\multicolumn{4}{@{}l}{\textit{\textbf{Cross-modal alignment} ($\mathcal{S}$, Def.~\ref{def:correspondence})}} \\
ImageBind AV score~\cite{girdhar2023imagebind} & audio-visual semantic agreement           & V/A  & $\uparrow$ \\
AV-Align~\cite{yariv2024tempotokens} / onset accuracy& temporal event synchronization            & V/A  & $\uparrow$ \\
LSE-C / LSE-D (SyncNet)~\cite{chung2016outoftime} & lip-sync confidence / distance            & V/A  & $\uparrow$ / $\downarrow$ \\
Beat alignment           & rhythmic synchronization                  & V/A  & $\uparrow$ \\
\midrule
\multicolumn{4}{@{}l}{\textit{\textbf{Editing} (Problem~\ref{prob:jointedit})}} \\
Directional faithfulness & whether the edit $e$ was applied          & V / A & $\uparrow$ \\
Masked PSNR / LPIPS~\cite{zhang2018lpips} & preservation outside the edit region      & V     & $\uparrow$ / $\downarrow$ \\
\bottomrule
\end{tabular}
\end{table*}

\subsection{Video Quality and Fidelity}
\label{sec:eval-videoquality}
Visual quality metrics score the realism and prompt-faithfulness of $\hat{v}$, typically through distances between feature distributions of generated and real clips. They are necessary but insufficient, since a model can score well while ignoring the audio entirely.

\subsection{Audio Quality and Fidelity}
\label{sec:eval-audioquality}
Audio quality metrics score the realism and prompt-faithfulness of $\hat{a}$, again through distributional distances or learned predictors. As with video, a high audio score does not imply correspondence with the visual stream.

\subsection{Cross-Modal Alignment Metrics}
\label{sec:eval-alignment}
Alignment metrics realize the score $\mathcal{S}$ of Definition~\ref{def:correspondence}, measuring either semantic agreement through contrastive audio-visual embeddings or temporal agreement through onset and beat distances. They are the metrics that distinguish joint and cross-modal evaluation from unimodal evaluation.

\subsection{Editing Faithfulness and Preservation}
\label{sec:eval-editing}
Editing metrics pair a faithfulness measure, which checks that the change named by $e$ was applied, with a preservation measure, which checks that untouched regions are unchanged. The two are in tension, and reporting one without the other is misleading.

\subsection{Human Evaluation Protocols}
\label{sec:eval-human}
Because correspondence is ultimately a perceptual property, human evaluation remains the reference standard, typically through forced-choice comparisons on quality and synchronization. Automatic metrics are validated by their agreement with these judgments.

\section{Applications}
\label{sec:applications}
Joint and cross-modal models are deployed across a range of settings, each placing its own constraints on $p_\theta$. Every application can be characterized by four elements: the input modalities, the output modalities, the underlying task of generation or editing, and the operational constraints such as latency, controllability, and identity preservation.
Content creation, dubbing and accessibility, real-time avatars, and personalization each constrain $p_\theta$ differently---controllability, identity-preserving propagation, a hard latency budget, and subject consistency, respectively.

\subsection{Film, Animation, and Content Creation}
\label{sec:applications-content}
In content creation the priority is controllability and quality, and the task spans both joint generation of new clips and editing of existing footage, often with a soundtrack composed of speech, effects, and music together.

\subsection{Dubbing, Translation, and Accessibility}
\label{sec:applications-accessibility}
Dubbing and translation are editing applications in which audio and video must change together while identity is preserved, which makes them the clearest instance of the propagation requirement of Problem~\ref{prob:jointedit}. JUST-DUB-IT~\cite{chen2026justdubit} generates translated speech and matching facial motion jointly while holding speaker identity fixed, and EditYourself~\cite{flynn2026edityourself} addresses the related re-voicing case. Accessibility uses the same machinery in the other direction, adding or adapting content for different audiences, and shares the preservation constraint: source-preserving methods such as MMAudioSep~\cite{takahashi2025mmaudiosep} and the audio-follows-video-edit setting of CoherentAVEdit~\cite{ishii2025coherent} are the research counterparts. What distinguishes this family operationally is that a wrong edit is worse than no edit, since the input is real footage a user already has.

\subsection{Virtual Avatars and Telepresence}
\label{sec:applications-avatars}
Avatars and telepresence impose the only hard latency constraint in this work: generation must keep pace with speech, which rules out the bidirectional sampling that every other application takes for granted. Streaming formulations are therefore central. Hallo-Live~\cite{li2026hallolive} generates avatar video in causal blocks that attend to a short horizon of future phonetic cues, OmniForcing~\cite{su2026omniforcing} distills a causal student from a bidirectional teacher to reach real-time joint generation, and SoundReactor~\cite{saito2025soundreactor} is the only method in Table~\ref{tab:v2a-control-taxonomy} marked as online, producing audio frame by frame. The constraint compounds with lip synchronization, whose tolerance is roughly a single frame, so this setting demands the tightest alignment under the least favourable sampling budget.

\subsection{Personalization}
\label{sec:applications-personalization}
Personalization conditions $p_\theta$ on a specific identity, voice, or style, so that generated or edited content matches a target while remaining coherent across modalities. It cuts across the other three applications rather than standing apart from them: the identity preservation that dubbing requires, the speaker consistency that avatars require, and the style control that content creation requires are the same constraint applied at different points. UniAVGen~\cite{zhang2025uniavgen} addresses it during generation through face-aware modulation that holds appearance and timbre consistent, while the editing methods of Section~\ref{sec:jointedit} address it as a preservation requirement on an existing subject. Personalization is also where the ethical exposure of Section~\ref{sec:open-ethics} is sharpest, since the capability that makes a legitimate avatar convincing is the capability that makes an impersonation convincing.

\section{Open Problems and Future Directions}
\label{sec:open}
\paragraph{Scaling joint models.}
\label{sec:open-scaling}
Reaching the best unimodal quality on both streams without multiplicative data and compute cost is open; pretraining reuse is the main lever.

\paragraph{Long-horizon coherence.}
\label{sec:open-longhorizon}
Maintaining correspondence over minutes is unsolved: errors accumulate and identity and scene drift---a problem distinct from short-clip synchronization.

\paragraph{Fine-grained cross-modal control.}
\label{sec:open-control}
Controlling which source sounds, when, and how loud---per event, not per clip---is the controllability counterpart of raising $\mathcal{S}$ at a fine temporal scale.

\paragraph{Physical plausibility.}
\label{sec:open-physics}
A clip can be synchronized yet physically wrong---sound mismatching the material or force of its visual event---and joint models often fail physical-commonsense benchmarks~\cite{cui2026avphys}.

\paragraph{Under-explored design space.}
\label{sec:open-designspace}
The empty cells of Table~\ref{tab:av-design-taxonomy} mark directions untried rather than failed: no joint method covered here generates either stream as discrete tokens (\S\ref{sec:strategy}), and audio-to-video generation has only isolated dedicated attempts (\S\ref{sec:crossgen-a2v}, \S\ref{sec:crossgen-s2v}); each gap has a structural cause and is a reason to study the problem.

\paragraph{Evaluation gaps.}
\label{sec:open-evalgaps}
Metrics measure per-modality quality well and correspondence poorly; one tracking human judgment of correspondence as reliably as quality metrics do is missing~\cite{zhou2026avgenbench}.

\paragraph{Ethics, safety, and watermarking.}
\label{sec:open-ethics}
Synchronized speech-and-face generation lowers the barrier to impersonation, making provenance, watermarking, and detection integral; defenses must treat both streams together.

\section{Conclusion}
\label{sec:conclusion}
We examined generation and editing of video and audio as three problems over one distribution on audio-visual pairs, organized by a five-axis design taxonomy; the trajectory toward unified backbones reduces the remaining open problems to raising cross-modal alignment while keeping per-stream quality high.
The empty cells of our taxonomies are as informative as the occupied ones: token-based joint generation, dedicated audio-to-video methods, and learned narrative and restoration editing are absent or nearly so for structural reasons, each a concrete opening for future work.
\label{page:endmain}

\section*{Limitations}
This work makes several scoping decisions that bound what it can claim. First, we cover only methods in which at least one of video or audio is an output and the other appears in the pipeline as an input or output; single-modality generation (for example, text-to-video without sound or text-to-audio alone), and audio-visual representation learning, retrieval, and understanding without a generative or editing component, are out of scope except as background. Second, we exclude fully closed commercial systems from the taxonomy of Section~\ref{sec:taxonomy}: systems such as Veo~3~\cite{deepmind2025veo3} and Sora~2~\cite{openai2025sora2} generate synchronized audio-visual content at or beyond the state of the art, but they disclose neither their representations nor their synchronization mechanisms, so placing them on the design axes would amount to guessing; we instead list user-facing capabilities of commercial systems in Appendix~\ref{app:other-taxonomies}, and mark partially documented systems (for example, Wan~2.5) only on the axes their reports support. The reader should therefore treat the taxonomy as a map of the \emph{documented} literature, and remember that some of the strongest current systems are absent from it by construction. Third, the field is moving quickly: our coverage reflects the literature through mid-2026, several of the systems covered here are described only in preprints or technical reports whose details may change, and empty regions of our taxonomy may fill rapidly. Finally, the five design axes are complementary rather than mutually exclusive, and assigning a method to a category occasionally requires judgment where papers are ambiguous; the per-method tables record our reading, and the cited sources remain authoritative.

% \section*{Acknowledgments}

\bibliography{main}
\bibliographystyle{acl_natbib}

\appendix

\section{Other Taxonomies}
\label{app:other-taxonomies}
This appendix collects the per-method taxonomies that support the main text. Table~\ref{tab:avgen-unified-task-taxonomy} organizes generation and editing methods by task, conditioning signal, and audio target; Table~\ref{tab:v2a-control-taxonomy} details controls, deployment constraints, and modeling mechanisms for the video-to-audio and Foley family; Table~\ref{tab:avgen-modeling-taxonomy} summarizes the dominant modeling mechanism per method; Tables~\ref{tab:video-editing-taxonomy-a} and~\ref{tab:video-editing-taxonomy-b} cover the video-editing methods most relevant to audio-video pipelines; Table~\ref{tab:commercial-av-systems} lists user-facing capabilities of commercial and foundation systems; and Table~\ref{tab:code-availability} records per-method code availability with repository links, each verified individually in July 2026.

\providecommand{\sysName}[1]{\textsc{#1}}
\providecommand{\rotateDeg}{90}
\providecommand{\cellszxs}{0.30cm}
\providecommand{\cellszsm}{0.30cm}
\providecommand{\cellszmd}{0.30cm}
\providecommand{\cellszlg}{0.30cm}
\providecommand{\rot}[1]{\rotatebox{\rotateDeg}{\textbf{#1}}}
\providecommand{\midhline}{\noalign{\hrule height 0.6pt}}
\definecolor{googlegreen}{HTML}{0F9D58}
\definecolor{googleblue}{HTML}{4285F4}
\definecolor{googlered}{HTML}{DB4437}
\definecolor{googlepurple}{HTML}{8E24AA}
\definecolor{googleyellow}{HTML}{F4B400}
\renewcommand\TE{\rule{0pt}{2.0ex}}% renew: also defined before Table 1
\renewcommand\BE{\rule[-1.1ex]{0pt}{0pt}}% renew: also defined before Table 1
{
\newcolumntype{C}{ >{\centering\arraybackslash} m{4cm} }
\providecommand{\rotateDeg}{90}
\setlength{\tabcolsep}{1.8pt}
\providecommand{\rotDeg}{70}
\definecolor{verylightgreennew}{RGB}	{220,255,220}
\definecolor{verylightrednew}{RGB}		{255, 230, 230}
\definecolor{verylightreddarker}{HTML} {FFCBCB}
\definecolor{verylightrednew}{RGB}		{255, 230, 230}
\definecolor{verylightrednewlighter}{RGB}		{255, 229, 239}
\definecolor{lightgraynew}{rgb}{0.95,0.95,0.95}
\definecolor{newgray}{RGB}{0.3,0.3,0.3}
\providecommand{\cellsz}{0.30cm}
\providecommand{\cellszlg}{0.30cm}
\providecommand{\cellszsm}{0.30cm}
\renewcommand{\cm}{{\color{greencm}\normalsize\cmark}}
\renewcommand{\cmgray}{{\color{lightgraynew}\normalsize\cmark}}
\renewcommand{\xm}{{\color{verylightreddarker}\normalsize\xmark}}
\newcommand\BBBBB{\rule[1.0ex]{0pt}{1.0ex}}
\newcommand\BBBnew{\rule[-2.5ex]{0pt}{0pt}}
\newcommand\BBBBBB{\rule[-1.0ex]{0pt}{0pt}}
\renewcommand{\sysName}[1]{{% renew: \providecommand above already defines it
\BBBBBB
#1
}}
\providecommand{\cellsomewhat}{
\BBBBB
\cmgray
\cellcolor{verylightgreennew}
}
\providecommand{\cellno}{
\BBBBB
\xm
\cellcolor{verylightrednew}}
\providecommand{\cellyes}{
\BBBBB
\cm
\cellcolor{verylightgreennew}
}
\begin{table*}[t!]
\centering
\renewcommand{\arraystretch}{1.12}
\scriptsize
\caption{
\textbf{Unified taxonomy of video-audio generation and editing methods.}
\textcolor{googlegreen}{\sc tasks} distinguish whether a method generates audio-video from text (T2AV), synthesizes audio for a given video (V2A/Foley), synthesizes video from audio (A2V), generates video and audio jointly (Joint), or supports editing.
\textcolor{googleblue}{\sc conditioning signals} capture the user/model inputs used to steer generation.
\textcolor{googlered}{\sc audio targets} indicate what acoustic layers are explicitly modeled. A check mark indicates that the method supports the corresponding capability.
}
\label{tab:avgen-unified-task-taxonomy}
\begin{tabular}{H l @{\hspace{8pt}} P{\cellszmd}P{\cellszmd}P{\cellszmd}P{\cellszmd}P{\cellszmd} @{\hspace{8pt}} P{\cellszmd}P{\cellszmd}P{\cellszmd}P{\cellszmd}P{\cellszmd}P{\cellszmd} @{\hspace{8pt}} P{\cellszmd}P{\cellszmd}P{\cellszmd}P{\cellszmd} @{\hspace{8pt}} l}
\toprule
& & \multicolumn{5}{c}{\textcolor{googlegreen}{\textsc{\bfseries Tasks}}}
& \multicolumn{6}{c}{\textcolor{googleblue}{\textsc{\bfseries Conditioning Signals}}}
& \multicolumn{4}{c}{\textcolor{googlered}{\textsc{\bfseries Audio Targets}}} & \\
\cmidrule(lr){3-7}\cmidrule(lr){8-13}\cmidrule(lr){14-17}
\textbf{Year} & \textbf{Method} &
\rot{T2AV} & \rot{V2A/Foley} & \rot{A2V} & \rot{Joint} & \rot{Editing} &
\rot{Text} & \rot{Video} & \rot{Audio} & \rot{Image/Ref.} & \rot{Spatial Ctrl.} & \rot{Temporal Ctrl.} &
\rot{SFX/Foley} & \rot{Speech} & \rot{Music} & \rot{Ambience} & \textbf{Primary Domain} \\
\midrule
2016 & \sysName{Visually Indicated Sounds}~\citep{owens2016visually} & \cellno & \cellyes & \cellno & \cellno & \cellno & \cellno & \cellyes & \cellno & \cellno & \cellno & \cellyes & \cellyes & \cellno & \cellno & \cellno & Foley / impacts \\
2018 & \sysName{Visual to Sound}~\citep{zhou2018visualsound} & \cellno & \cellyes & \cellno & \cellno & \cellno & \cellno & \cellyes & \cellno & \cellno & \cellno & \cellyes & \cellyes & \cellno & \cellno & \cellno & in-the-wild SFX \\
2020 & \sysName{Visually Aligned Sound}~\citep{chen2020vas} & \cellno & \cellyes & \cellno & \cellno & \cellno & \cellno & \cellyes & \cellno & \cellno & \cellno & \cellyes & \cellyes & \cellno & \cellno & \cellno & general V2A \\
2020 & \sysName{Foley Music}~\citep{gan2020foleymusic} & \cellno & \cellyes & \cellno & \cellno & \cellno & \cellno & \cellyes & \cellno & \cellno & \cellno & \cellyes & \cellno & \cellno & \cellyes & \cellno & video-to-music \\
2021 & \sysName{Rhythmic Soundtracks}~\citep{gan2021rhythmic} & \cellno & \cellyes & \cellno & \cellno & \cellno & \cellno & \cellyes & \cellno & \cellno & \cellno & \cellyes & \cellno & \cellno & \cellyes & \cellno & human movement \\
2021 & \sysName{Controllable BGM Transformer}~\citep{di2021bgmtransformer} & \cellno & \cellyes & \cellno & \cellno & \cellno & \cellno & \cellyes & \cellno & \cellno & \cellno & \cellyes & \cellno & \cellno & \cellyes & \cellno & background music \\
2022 & \sysName{Dance2Music Q-GAN}~\citep{zhu2022quantizedgan} & \cellno & \cellyes & \cellno & \cellno & \cellno & \cellno & \cellyes & \cellno & \cellno & \cellno & \cellyes & \cellno & \cellno & \cellyes & \cellno & dance music \\
2023 & \sysName{MM-Diffusion}~\citep{ruan2023mmdiffusion} & \cellno & \cellyes & \cellyes & \cellyes & \cellno & \cellno & \cellyes & \cellyes & \cellno & \cellno & \cellyes & \cellyes & \cellno & \cellyes & \cellyes & sounding video \\
2023 & \sysName{Diff-Foley}~\citep{luo2023difffoley} & \cellno & \cellyes & \cellno & \cellno & \cellno & \cellno & \cellyes & \cellno & \cellno & \cellno & \cellyes & \cellyes & \cellno & \cellno & \cellno & Foley / SFX \\
2023 & \sysName{Foley Analogies}~\citep{iyer2023foleyanalogies} & \cellno & \cellyes & \cellno & \cellno & \cellyes & \cellno & \cellyes & \cellyes & \cellno & \cellno & \cellyes & \cellyes & \cellno & \cellno & \cellno & reference-guided Foley \\
2023 & \sysName{Video BGM Generation}~\citep{di2023vbgm} & \cellno & \cellyes & \cellno & \cellno & \cellno & \cellno & \cellyes & \cellno & \cellno & \cellno & \cellyes & \cellno & \cellno & \cellyes & \cellno & background music \\
2023 & \sysName{Long-Term Rhythmic Soundtracker}~\citep{zhu2023soundtracker} & \cellno & \cellyes & \cellno & \cellno & \cellno & \cellno & \cellyes & \cellno & \cellno & \cellno & \cellyes & \cellno & \cellno & \cellyes & \cellno & long rhythmic music \\
2024 & \sysName{Seeing-and-Hearing}~\citep{xing2024seeing} & \cellyes & \cellyes & \cellyes & \cellyes & \cellno & \cellyes & \cellyes & \cellyes & \cellyes & \cellno & \cellyes & \cellyes & \cellno & \cellyes & \cellyes & open-domain AV \\
2024 & \sysName{V2A-Mapper}~\citep{lin2024v2amapper} & \cellno & \cellyes & \cellno & \cellno & \cellno & \cellno & \cellyes & \cellno & \cellno & \cellno & \cellyes & \cellyes & \cellno & \cellno & \cellyes & foundation mapper \\
2024 & \sysName{SonicVisionLM}~\citep{liu2024sonicvisionlm} & \cellno & \cellyes & \cellno & \cellno & \cellno & \cellyes & \cellyes & \cellno & \cellno & \cellno & \cellyes & \cellyes & \cellno & \cellno & \cellyes & VLM-guided audio \\
2025 & \sysName{Video-Foley}~\citep{lee2024videofoley} & \cellno & \cellyes & \cellno & \cellno & \cellyes & \cellyes & \cellyes & \cellyes & \cellno & \cellno & \cellyes & \cellyes & \cellno & \cellno & \cellno & RMS-conditioned Foley \\
2026 & \sysName{FoleyCrafter}~\citep{zhang2024foleycrafter} & \cellno & \cellyes & \cellno & \cellno & \cellyes & \cellyes & \cellyes & \cellno & \cellno & \cellno & \cellyes & \cellyes & \cellno & \cellno & \cellyes & text-controlled Foley \\
2024 & \sysName{Frieren}~\citep{wang2024frieren} & \cellno & \cellyes & \cellno & \cellno & \cellno & \cellno & \cellyes & \cellno & \cellno & \cellno & \cellyes & \cellyes & \cellno & \cellno & \cellyes & efficient V2A \\
2024 & \sysName{MaskVAT}~\citep{garoufis2024maskvat} & \cellno & \cellyes & \cellno & \cellno & \cellno & \cellno & \cellyes & \cellno & \cellno & \cellno & \cellyes & \cellyes & \cellno & \cellno & \cellyes & masked V2A \\
2024 & \sysName{STA-V2A}~\citep{chen2024stav2a} & \cellno & \cellyes & \cellno & \cellno & \cellno & \cellyes & \cellyes & \cellno & \cellno & \cellno & \cellyes & \cellyes & \cellno & \cellno & \cellyes & semantic/temporal V2A \\
2024 & \sysName{VATT}~\citep{liu2024vatt} & \cellno & \cellyes & \cellno & \cellno & \cellyes & \cellyes & \cellyes & \cellno & \cellno & \cellno & \cellyes & \cellyes & \cellno & \cellno & \cellyes & caption-mediated V2A \\
2024 & \sysName{MM-LDM}~\citep{sun2024mmldm} & \cellyes & \cellyes & \cellno & \cellyes & \cellno & \cellyes & \cellyes & \cellyes & \cellno & \cellno & \cellyes & \cellyes & \cellno & \cellyes & \cellyes & sounding video \\
2024 & \sysName{Movie Gen}~\citep{polyak2024moviegen} & \cellyes & \cellyes & \cellno & \cellyes & \cellyes & \cellyes & \cellyes & \cellyes & \cellyes & \cellyes & \cellyes & \cellyes & \cellyes & \cellyes & \cellyes & media foundation model \\
2024 & \sysName{Google V2A}~\citep{deepmind2024v2a} & \cellno & \cellyes & \cellno & \cellno & \cellyes & \cellyes & \cellyes & \cellno & \cellno & \cellno & \cellyes & \cellyes & \cellyes & \cellyes & \cellyes & video soundtracks \\
2025 & \sysName{MMAudio}~\citep{cheng2025mmaudio} & \cellno & \cellyes & \cellno & \cellno & \cellyes & \cellyes & \cellyes & \cellno & \cellno & \cellno & \cellyes & \cellyes & \cellno & \cellyes & \cellyes & fast V2A / T2A \\
2025 & \sysName{Mel-QCD}~\citep{wang2025melqcd} & \cellno & \cellyes & \cellno & \cellno & \cellno & \cellyes & \cellyes & \cellno & \cellno & \cellno & \cellyes & \cellyes & \cellno & \cellno & \cellyes & mel-control V2A \\
2025 & \sysName{VAFlow}~\citep{wang2025vaflow} & \cellno & \cellyes & \cellno & \cellno & \cellno & \cellno & \cellyes & \cellno & \cellno & \cellno & \cellyes & \cellyes & \cellno & \cellno & \cellyes & flow matching V2A \\
2025 & \sysName{Foley-Flow}~\citep{mo2025foleyflow} & \cellno & \cellyes & \cellno & \cellno & \cellno & \cellno & \cellyes & \cellno & \cellno & \cellno & \cellyes & \cellyes & \cellno & \cellno & \cellyes & masked AV flow \\
2025 & \sysName{MultiFoley}~\citep{chen2025multifoley} & \cellno & \cellyes & \cellno & \cellno & \cellyes & \cellyes & \cellyes & \cellyes & \cellno & \cellno & \cellyes & \cellyes & \cellno & \cellno & \cellyes & professional Foley \\
2025 & \sysName{VinTAGe}~\citep{tian2025vintage} & \cellno & \cellyes & \cellno & \cellno & \cellyes & \cellyes & \cellyes & \cellno & \cellno & \cellno & \cellyes & \cellyes & \cellno & \cellyes & \cellyes & holistic audio \\
2025 & \sysName{TARO}~\citep{zhang2025taro} & \cellno & \cellyes & \cellno & \cellno & \cellno & \cellno & \cellyes & \cellno & \cellno & \cellno & \cellyes & \cellyes & \cellno & \cellno & \cellyes & onset-aware V2A \\
2025 & \sysName{ThinkSound}~\citep{liu2025thinksound} & \cellno & \cellyes & \cellno & \cellno & \cellyes & \cellyes & \cellyes & \cellyes & \cellno & \cellyes & \cellyes & \cellyes & \cellno & \cellno & \cellyes & interactive audio edit \\
2025 & \sysName{Hear-Your-Click}~\citep{guo2025hearyourclick} & \cellno & \cellyes & \cellno & \cellno & \cellyes & \cellyes & \cellyes & \cellno & \cellno & \cellyes & \cellyes & \cellyes & \cellno & \cellno & \cellyes & click/object V2A \\
2025 & \sysName{SelVA}~\citep{anonymous2025selva} & \cellno & \cellyes & \cellno & \cellno & \cellyes & \cellyes & \cellyes & \cellno & \cellno & \cellyes & \cellyes & \cellyes & \cellno & \cellno & \cellno & selective V2A \\
2025 & \sysName{MMAudioSep}~\citep{takahashi2025mmaudiosep} & \cellno & \cellno & \cellno & \cellno & \cellyes & \cellyes & \cellyes & \cellyes & \cellno & \cellyes & \cellyes & \cellyes & \cellno & \cellyes & \cellyes & video/text queried separation \\
2025 & \sysName{CoherentAVEdit}~\citep{ishii2025coherent} & \cellno & \cellyes & \cellno & \cellno & \cellyes & \cellyes & \cellyes & \cellyes & \cellno & \cellno & \cellyes & \cellyes & \cellno & \cellyes & \cellyes & audio follows video edit \\
2025 & \sysName{AV-Link}~\citep{hajiali2025avlink} & \cellno & \cellyes & \cellyes & \cellno & \cellyes & \cellyes & \cellyes & \cellyes & \cellno & \cellno & \cellyes & \cellyes & \cellno & \cellyes & \cellyes & bidirectional AV \\
2025 & \sysName{JavisDiT}~\citep{liu2025javisdit} & \cellyes & \cellno & \cellno & \cellyes & \cellno & \cellyes & \cellno & \cellno & \cellno & \cellno & \cellyes & \cellyes & \cellno & \cellyes & \cellyes & prompt-to-AV \\
2025 & \sysName{UniAVGen}~\citep{zhang2025uniavgen} & \cellyes & \cellyes & \cellyes & \cellyes & \cellyes & \cellyes & \cellyes & \cellyes & \cellyes & \cellno & \cellyes & \cellyes & \cellyes & \cellno & \cellyes & unified AV \\
2025 & \sysName{RFLAV}~\citep{ergasti2025rflav} & \cellyes & \cellno & \cellno & \cellyes & \cellno & \cellyes & \cellno & \cellno & \cellno & \cellno & \cellyes & \cellyes & \cellno & \cellyes & \cellyes & long/infinite AV \\
2025 & \sysName{SoundReactor}~\citep{saito2025soundreactor} & \cellno & \cellyes & \cellno & \cellno & \cellno & \cellno & \cellyes & \cellno & \cellno & \cellno & \cellyes & \cellyes & \cellno & \cellyes & \cellyes & online/game V2A \\
2025 & \sysName{Sora 2}~\citep{openai2025sora2} & \cellyes & \cellno & \cellno & \cellyes & \cellyes & \cellyes & \cellyes & \cellyes & \cellyes & \cellno & \cellyes & \cellyes & \cellyes & \cellyes & \cellyes & commercial video+audio \\
2025 & \sysName{Veo 3/3.1}~\citep{deepmind2025veo3} & \cellyes & \cellyes & \cellno & \cellyes & \cellyes & \cellyes & \cellyes & \cellyes & \cellyes & \cellno & \cellyes & \cellyes & \cellyes & \cellyes & \cellyes & commercial video+audio \\
2025 & \sysName{Adobe Firefly Audio}~\citep{adobe2025fireflyaudio} & \cellno & \cellyes & \cellno & \cellno & \cellyes & \cellyes & \cellyes & \cellyes & \cellno & \cellno & \cellyes & \cellyes & \cellyes & \cellyes & \cellyes & commercial sound design \\
2026 & \sysName{Seedance 2.0}~\citep{bytedance2026seedance2} & \cellyes & \cellno & \cellno & \cellyes & \cellyes & \cellyes & \cellyes & \cellyes & \cellyes & \cellno & \cellyes & \cellyes & \cellyes & \cellyes & \cellyes & commercial multimodal video \\
2026 & \sysName{Foley-Omni}~\citep{tao2026foleyomni} & \cellno & \cellyes & \cellno & \cellno & \cellyes & \cellyes & \cellyes & \cellno & \cellno & \cellno & \cellyes & \cellyes & \cellyes & \cellyes & \cellyes & full video soundtrack \\
2026 & \sysName{MOVA}~\citep{mova2026} & \cellyes & \cellno & \cellno & \cellyes & \cellno & \cellyes & \cellno & \cellno & \cellno & \cellno & \cellyes & \cellyes & \cellyes & \cellyes & \cellyes & scalable synchronized AV \\
\bottomrule
\end{tabular}
\vspace{-1mm}
\end{table*}

\begin{table*}[t!]
\centering
\renewcommand{\arraystretch}{1.12}
\scriptsize
\caption{
\textbf{Fine-grained taxonomy of video-to-audio, Foley, and audio-following-video-edit methods.}
\textcolor{googleblue}{\sc controls} summarize how a user or upstream system steers the generated soundtrack.
\textcolor{googlegreen}{\sc deployment / editing} constraints distinguish long-form, stereo/spatial, online, and source-preserving settings.
\textcolor{googlered}{\sc modeling mechanisms} summarize the dominant generator or alignment mechanism.
}
\label{tab:v2a-control-taxonomy}
\begin{tabular}{l @{\hspace{8pt}} P{\cellszmd}P{\cellszmd}P{\cellszmd}P{\cellszmd} @{\hspace{8pt}} P{\cellszmd}P{\cellszmd}P{\cellszmd}P{\cellszmd} @{\hspace{8pt}} P{\cellszmd}P{\cellszmd}P{\cellszmd}P{\cellszmd}P{\cellszmd}P{\cellszmd}P{\cellszmd}}
\toprule
& \multicolumn{4}{c}{\textcolor{googleblue}{\textsc{\bfseries Controls}}}
& \multicolumn{4}{c}{\textcolor{googlegreen}{\textsc{\bfseries Deployment / Editing}}}
& \multicolumn{7}{c}{\textcolor{googlered}{\textsc{\bfseries Modeling Mechanism}}} \\
\cmidrule(lr){2-5}\cmidrule(lr){6-9}\cmidrule(lr){10-16}
\textbf{Method} &
\rot{Text} & \rot{Audio Ref.} & \rot{Click/Mask} & \rot{Onset/Rhythm} &
\rot{Long-form} & \rot{Stereo/Spatial} & \rot{Online} & \rot{Preserve Src.} &
\rot{LDM} & \rot{Flow/RF} & \rot{DiT} & \rot{AR/Masked} & \rot{ControlNet} & \rot{Guidance/Adapter} & \rot{MLLM/CoT} \\
\midrule
\sysName{Diff-Foley}~\citep{luo2023difffoley} & \cellno & \cellno & \cellno & \cellyes & \cellno & \cellno & \cellno & \cellno & \cellyes & \cellno & \cellno & \cellno & \cellno & \cellno & \cellno \\
\sysName{Foley Analogies}~\citep{iyer2023foleyanalogies} & \cellno & \cellyes & \cellno & \cellyes & \cellno & \cellno & \cellno & \cellno & \cellno & \cellno & \cellno & \cellno & \cellno & \cellyes & \cellno \\
\sysName{Seeing-and-Hearing}~\citep{xing2024seeing} & \cellyes & \cellyes & \cellno & \cellno & \cellno & \cellno & \cellno & \cellno & \cellyes & \cellno & \cellno & \cellno & \cellno & \cellyes & \cellno \\
\sysName{V2A-Mapper}~\citep{lin2024v2amapper} & \cellno & \cellno & \cellno & \cellno & \cellno & \cellno & \cellno & \cellno & \cellno & \cellno & \cellno & \cellno & \cellno & \cellyes & \cellno \\
\sysName{Video-Foley}~\citep{lee2024videofoley} & \cellyes & \cellyes & \cellno & \cellyes & \cellno & \cellno & \cellno & \cellno & \cellyes & \cellno & \cellno & \cellno & \cellyes & \cellno & \cellno \\
\sysName{FoleyCrafter}~\citep{zhang2024foleycrafter} & \cellyes & \cellno & \cellno & \cellyes & \cellno & \cellno & \cellno & \cellno & \cellyes & \cellno & \cellno & \cellno & \cellyes & \cellno & \cellno \\
\sysName{Frieren}~\citep{wang2024frieren} & \cellno & \cellno & \cellno & \cellyes & \cellno & \cellno & \cellno & \cellno & \cellno & \cellyes & \cellno & \cellno & \cellno & \cellno & \cellno \\
\sysName{MaskVAT}~\citep{garoufis2024maskvat} & \cellno & \cellno & \cellno & \cellyes & \cellno & \cellno & \cellno & \cellno & \cellno & \cellno & \cellno & \cellyes & \cellno & \cellno & \cellno \\
\sysName{STA-V2A}~\citep{chen2024stav2a} & \cellyes & \cellno & \cellno & \cellyes & \cellno & \cellno & \cellno & \cellno & \cellyes & \cellno & \cellno & \cellno & \cellno & \cellno & \cellno \\
\sysName{VATT}~\citep{liu2024vatt} & \cellyes & \cellno & \cellno & \cellno & \cellno & \cellno & \cellno & \cellno & \cellyes & \cellno & \cellno & \cellno & \cellno & \cellno & \cellyes \\
\sysName{Google V2A}~\citep{deepmind2024v2a} & \cellyes & \cellno & \cellno & \cellyes & \cellno & \cellno & \cellno & \cellno & \cellyes & \cellno & \cellno & \cellno & \cellno & \cellyes & \cellno \\
\sysName{Movie Gen Audio}~\citep{polyak2024moviegen} & \cellyes & \cellno & \cellno & \cellyes & \cellyes & \cellno & \cellno & \cellno & \cellno & \cellno & \cellyes & \cellno & \cellno & \cellno & \cellno \\
\sysName{MMAudio}~\citep{cheng2025mmaudio} & \cellyes & \cellno & \cellno & \cellyes & \cellno & \cellno & \cellno & \cellno & \cellno & \cellyes & \cellno & \cellno & \cellno & \cellno & \cellno \\
\sysName{Mel-QCD}~\citep{wang2025melqcd} & \cellyes & \cellno & \cellno & \cellyes & \cellno & \cellno & \cellno & \cellno & \cellyes & \cellno & \cellno & \cellno & \cellyes & \cellno & \cellno \\
\sysName{VAFlow}~\citep{wang2025vaflow} & \cellno & \cellno & \cellno & \cellyes & \cellno & \cellno & \cellno & \cellno & \cellno & \cellyes & \cellno & \cellno & \cellno & \cellno & \cellno \\
\sysName{Foley-Flow}~\citep{mo2025foleyflow} & \cellno & \cellno & \cellno & \cellyes & \cellno & \cellno & \cellno & \cellno & \cellno & \cellyes & \cellno & \cellno & \cellno & \cellno & \cellno \\
\sysName{MultiFoley}~\citep{chen2025multifoley} & \cellyes & \cellyes & \cellno & \cellyes & \cellyes & \cellno & \cellno & \cellno & \cellno & \cellno & \cellyes & \cellno & \cellno & \cellno & \cellno \\
\sysName{ThinkSound}~\citep{liu2025thinksound} & \cellyes & \cellyes & \cellyes & \cellyes & \cellno & \cellno & \cellno & \cellno & \cellno & \cellno & \cellyes & \cellno & \cellno & \cellno & \cellyes \\
\sysName{Hear-Your-Click}~\citep{guo2025hearyourclick} & \cellyes & \cellno & \cellyes & \cellyes & \cellno & \cellno & \cellno & \cellno & \cellno & \cellno & \cellyes & \cellno & \cellno & \cellno & \cellyes \\
\sysName{SelVA}~\citep{anonymous2025selva} & \cellyes & \cellno & \cellno & \cellyes & \cellno & \cellno & \cellno & \cellno & \cellno & \cellno & \cellyes & \cellno & \cellno & \cellno & \cellno \\
\sysName{MMAudioSep}~\citep{takahashi2025mmaudiosep} & \cellyes & \cellyes & \cellyes & \cellyes & \cellno & \cellyes & \cellno & \cellyes & \cellno & \cellyes & \cellno & \cellno & \cellno & \cellno & \cellno \\
\sysName{CoherentAVEdit}~\citep{ishii2025coherent} & \cellyes & \cellyes & \cellno & \cellyes & \cellno & \cellno & \cellno & \cellyes & \cellno & \cellyes & \cellno & \cellno & \cellno & \cellno & \cellno \\
\sysName{AV-Link}~\citep{hajiali2025avlink} & \cellyes & \cellyes & \cellno & \cellyes & \cellno & \cellno & \cellno & \cellno & \cellno & \cellyes & \cellno & \cellno & \cellno & \cellno & \cellno \\
\sysName{SoundReactor}~\citep{saito2025soundreactor} & \cellno & \cellno & \cellno & \cellyes & \cellyes & \cellyes & \cellyes & \cellno & \cellno & \cellno & \cellyes & \cellyes & \cellno & \cellno & \cellno \\
\bottomrule
\end{tabular}
\vspace{-1mm}
\end{table*}
\begin{table*}[t!]
\centering
\renewcommand{\arraystretch}{1.12}
\scriptsize
\caption{
\textbf{Modeling-mechanism taxonomy for representative audio-video generation and editing methods.}
The table bridges the task-level taxonomies and method sections: it identifies whether a method primarily uses GAN/VQ, autoregression, latent diffusion, flow/rectified-flow, masked modeling, AV alignment losses, joint denoising, frozen-model guidance/adapters, or MLLM/agent-style reasoning.
}
\label{tab:avgen-modeling-taxonomy}
\begin{tabular}{l @{\hspace{8pt}} P{\cellszmd}P{\cellszmd}P{\cellszmd}P{\cellszmd}P{\cellszmd}P{\cellszmd}P{\cellszmd}P{\cellszmd}P{\cellszmd} l}
\toprule
\textbf{Method} &
\rot{GAN/VQ} & \rot{AR/Transformer} & \rot{LDM} & \rot{Flow/RF} & \rot{Masked} & \rot{AV Align.} & \rot{Joint Denoise} & \rot{Frozen/Guidance} & \rot{MLLM/Agent} & \textbf{Main Mechanism} \\
\midrule
\sysName{MM-Diffusion}~\citep{ruan2023mmdiffusion} & \cellno & \cellno & \cellno & \cellno & \cellno & \cellno & \cellyes & \cellno & \cellno & joint multi-modal U-Net \\
\sysName{Diff-Foley}~\citep{luo2023difffoley} & \cellno & \cellno & \cellyes & \cellno & \cellno & \cellyes & \cellno & \cellno & \cellno & CAVP + latent diffusion \\
\sysName{Seeing-and-Hearing}~\citep{xing2024seeing} & \cellno & \cellno & \cellyes & \cellno & \cellno & \cellno & \cellno & \cellyes & \cellno & diffusion latent aligner \\
\sysName{V2A-Mapper}~\citep{lin2024v2amapper} & \cellno & \cellno & \cellno & \cellno & \cellno & \cellno & \cellno & \cellyes & \cellno & foundation-model mapper \\
\sysName{Video-Foley}~\citep{lee2024videofoley} & \cellno & \cellno & \cellyes & \cellno & \cellno & \cellno & \cellno & \cellyes & \cellno & RMS two-stage control \\
\sysName{FoleyCrafter}~\citep{zhang2024foleycrafter} & \cellno & \cellno & \cellyes & \cellno & \cellno & \cellno & \cellno & \cellyes & \cellno & semantic adapter + temporal controller \\
\sysName{Frieren}~\citep{wang2024frieren} & \cellno & \cellno & \cellno & \cellyes & \cellno & \cellyes & \cellno & \cellno & \cellno & rectified flow matching \\
\sysName{MaskVAT}~\citep{garoufis2024maskvat} & \cellno & \cellyes & \cellno & \cellno & \cellyes & \cellyes & \cellno & \cellno & \cellno & masked generative transformer \\
\sysName{STA-V2A}~\citep{chen2024stav2a} & \cellno & \cellno & \cellyes & \cellno & \cellno & \cellyes & \cellno & \cellno & \cellno & local/global visual features \\
\sysName{VATT}~\citep{liu2024vatt} & \cellno & \cellno & \cellyes & \cellno & \cellno & \cellno & \cellno & \cellyes & \cellyes & caption-mediated generation \\
\sysName{Movie Gen}~\citep{polyak2024moviegen} & \cellno & \cellyes & \cellno & \cellyes & \cellno & \cellno & \cellyes & \cellno & \cellno & scaled media transformers \\
\sysName{Google V2A}~\citep{deepmind2024v2a} & \cellno & \cellno & \cellyes & \cellno & \cellno & \cellno & \cellno & \cellyes & \cellno & prompt-conditioned diffusion \\
\sysName{MMAudio}~\citep{cheng2025mmaudio} & \cellno & \cellno & \cellno & \cellyes & \cellno & \cellyes & \cellno & \cellno & \cellno & flow matching + synchronization module \\
\sysName{Mel-QCD}~\citep{wang2025melqcd} & \cellno & \cellno & \cellyes & \cellno & \cellno & \cellyes & \cellno & \cellno & \cellno & mel decomposition + ControlNet \\
\sysName{VAFlow}~\citep{wang2025vaflow} & \cellno & \cellno & \cellno & \cellyes & \cellno & \cellyes & \cellno & \cellno & \cellno & cross-modal flow matching \\
\sysName{Foley-Flow}~\citep{mo2025foleyflow} & \cellno & \cellno & \cellno & \cellyes & \cellyes & \cellyes & \cellno & \cellno & \cellno & masked AV alignment + dynamic flow \\
\sysName{MultiFoley}~\citep{chen2025multifoley} & \cellno & \cellyes & \cellno & \cellno & \cellno & \cellyes & \cellno & \cellno & \cellno & DiT + multi-conditional training \\
\sysName{ThinkSound}~\citep{liu2025thinksound} & \cellno & \cellyes & \cellno & \cellno & \cellno & \cellno & \cellno & \cellyes & \cellyes & MLLM CoT + interactive editing \\
\sysName{AV-Link}~\citep{hajiali2025avlink} & \cellno & \cellno & \cellno & \cellyes & \cellno & \cellno & \cellno & \cellyes & \cellno & frozen flow-model feature links \\
\sysName{JavisDiT}~\citep{liu2025javisdit} & \cellno & \cellyes & \cellno & \cellno & \cellno & \cellyes & \cellyes & \cellno & \cellno & joint DiT + spatio-temporal prior \\
\sysName{UniAVGen}~\citep{zhang2025uniavgen} & \cellno & \cellyes & \cellno & \cellno & \cellno & \cellyes & \cellyes & \cellno & \cellno & dual-branch DiT cross-modal interaction \\
\sysName{SoundReactor}~\citep{saito2025soundreactor} & \cellno & \cellyes & \cellno & \cellno & \cellno & \cellyes & \cellno & \cellno & \cellno & causal transformer + diffusion head \\
\bottomrule
\end{tabular}
\vspace{-1mm}
\end{table*}

\begin{table*}[t!]
\centering
\renewcommand{\arraystretch}{1.12}
\scriptsize
\caption{
\textbf{Video editing methods most relevant to audio-video generation pipelines: temporal-adaptation, training-modification, and conditioning-branch families.}
These methods edit the visual stream; audio-video systems such as \sysName{CoherentAVEdit} can subsequently regenerate or adapt the soundtrack to match the edited result. The columns separate the user control signal, edit target, and implementation strategy.
}
\label{tab:video-editing-taxonomy-a}
\begin{tabular}{P{6mm} >{\RaggedRight\arraybackslash}p{50mm} p{30mm} @{\hspace{8pt}} P{\cellszsm}P{\cellszsm}P{\cellszsm}P{\cellszsm}P{\cellszsm}P{\cellszsm} @{\hspace{8pt}} P{\cellszsm}P{\cellszsm}P{\cellszsm} @{\hspace{8pt}} P{\cellszsm}P{\cellszsm}P{\cellszsm}P{\cellszsm}P{\cellszsm}P{\cellszsm}}
\toprule
\textbf{Year} & \textbf{Method} & \textbf{Family} &
\rot{Instr.} & \rot{Text} & \rot{Mask/Box} & \rot{Point/Traj.} & \rot{Pose} & \rot{Image/Style} &
\rot{Motion} & \rot{Appearance} & \rot{Inpaint} &
\rot{Tuning-free} & \rot{Fine-tune} & \rot{Attention} & \rot{Latent} & \rot{Cond. Branch} & \rot{Canonical} \\
\midrule
2024 & \sysName{VIA}~\citep{gu2024via} & Temporal adaptation & \cellno & \cellyes & \cellyes & \cellno & \cellno & \cellno & \cellyes & \cellyes & \cellno & \cellno & \cellyes & \cellno & \cellno & \cellyes & \cellno \\
2024 & \sysName{Slicedit}~\citep{cohen2024sliceedit} & Temporal adaptation & \cellno & \cellyes & \cellno & \cellno & \cellno & \cellno & \cellno & \cellyes & \cellno & \cellyes & \cellno & \cellno & \cellyes & \cellno & \cellno \\
2024 & \sysName{Factorized Diffusion Distillation}~\citep{singer2024fdd} & Temporal adaptation & \cellno & \cellyes & \cellno & \cellno & \cellno & \cellno & \cellyes & \cellyes & \cellno & \cellno & \cellyes & \cellno & \cellno & \cellno & \cellno \\
2024 & \sysName{MaskINT}~\citep{ma2024maskint} & Temporal adaptation & \cellyes & \cellyes & \cellyes & \cellno & \cellno & \cellno & \cellno & \cellyes & \cellyes & \cellno & \cellyes & \cellno & \cellno & \cellyes & \cellno \\
2023 & \sysName{Fairy}~\citep{wu2023fairy} & Temporal adaptation & \cellyes & \cellyes & \cellno & \cellno & \cellno & \cellno & \cellno & \cellyes & \cellno & \cellno & \cellyes & \cellno & \cellno & \cellyes & \cellno \\
2024 & \sysName{VidToMe}~\citep{li2024vidtome} & Temporal adaptation & \cellno & \cellyes & \cellno & \cellno & \cellno & \cellno & \cellno & \cellyes & \cellno & \cellyes & \cellno & \cellno & \cellyes & \cellno & \cellno \\
2024 & \sysName{SimDA}~\citep{xing2024simda} & Temporal adaptation & \cellno & \cellyes & \cellno & \cellno & \cellno & \cellno & \cellyes & \cellyes & \cellno & \cellno & \cellyes & \cellno & \cellno & \cellyes & \cellno \\
2023 & \sysName{Text-to-Image Diffusion Video Editing}~\citep{zhang2023consistent} & Temporal adaptation & \cellno & \cellyes & \cellno & \cellno & \cellno & \cellno & \cellno & \cellyes & \cellno & \cellno & \cellyes & \cellyes & \cellno & \cellno & \cellno \\
2023 & \sysName{Tune-A-Video}~\citep{wu2023tuneavideo} & Temporal adaptation & \cellno & \cellyes & \cellno & \cellno & \cellno & \cellno & \cellyes & \cellyes & \cellno & \cellno & \cellyes & \cellno & \cellno & \cellno & \cellno \\
2026 & \sysName{TrajectoryMover}~\citep{chhatre2026trajectorymover} & Conditioning branch & \cellyes & \cellyes & \cellno & \cellyes & \cellno & \cellno & \cellyes & \cellyes & \cellno & \cellno & \cellyes & \cellno & \cellno & \cellyes & \cellno \\
2025 & \sysName{VACE}~\citep{jiang2025vace} & Conditioning branch & \cellyes & \cellyes & \cellyes & \cellyes & \cellyes & \cellyes & \cellyes & \cellyes & \cellyes & \cellno & \cellyes & \cellno & \cellno & \cellyes & \cellno \\
2025 & \sysName{VideoPainter}~\citep{bian2025videopainter} & Conditioning branch & \cellyes & \cellyes & \cellyes & \cellno & \cellno & \cellno & \cellno & \cellyes & \cellyes & \cellno & \cellyes & \cellno & \cellno & \cellyes & \cellno \\
2024 & \sysName{StableV2V}~\citep{liu2024stablev2v} & Conditioning branch & \cellno & \cellyes & \cellno & \cellno & \cellno & \cellno & \cellno & \cellyes & \cellno & \cellno & \cellyes & \cellno & \cellno & \cellyes & \cellno \\
2024 & \sysName{EVA}~\citep{yang2024eva} & Conditioning branch & \cellyes & \cellyes & \cellyes & \cellno & \cellno & \cellno & \cellno & \cellyes & \cellno & \cellyes & \cellno & \cellno & \cellno & \cellyes & \cellno \\
2024 & \sysName{Diffutoon}~\citep{duan2024diffutoon} & Conditioning branch & \cellno & \cellyes & \cellno & \cellno & \cellno & \cellyes & \cellno & \cellyes & \cellno & \cellno & \cellyes & \cellno & \cellno & \cellyes & \cellno \\
2024 & \sysName{FlowVid}~\citep{liang2024flowvid} & Conditioning branch & \cellno & \cellyes & \cellno & \cellno & \cellno & \cellno & \cellyes & \cellyes & \cellno & \cellyes & \cellno & \cellno & \cellno & \cellyes & \cellno \\
2024 & \sysName{AVID}~\citep{zhang2024avid} & Conditioning branch & \cellyes & \cellyes & \cellyes & \cellno & \cellno & \cellno & \cellno & \cellno & \cellyes & \cellno & \cellyes & \cellno & \cellno & \cellyes & \cellno \\
2023 & \sysName{Motion-Conditioned Image Animation}~\citep{yan2023motionconditioned} & Conditioning branch & \cellno & \cellno & \cellno & \cellno & \cellno & \cellyes & \cellyes & \cellno & \cellno & \cellno & \cellyes & \cellno & \cellno & \cellyes & \cellno \\
2024 & \sysName{LAMP}~\citep{wu2024lamp} & Conditioning branch & \cellno & \cellyes & \cellno & \cellno & \cellno & \cellno & \cellyes & \cellno & \cellno & \cellno & \cellyes & \cellno & \cellno & \cellyes & \cellno \\
2024 & \sysName{Ground-A-Video}~\citep{jeong2024groundavideo} & Conditioning branch & \cellyes & \cellyes & \cellyes & \cellno & \cellno & \cellno & \cellno & \cellyes & \cellno & \cellyes & \cellno & \cellno & \cellno & \cellyes & \cellno \\
2023 & \sysName{CCEdit}~\citep{feng2023ccedit} & Conditioning branch & \cellyes & \cellyes & \cellyes & \cellno & \cellno & \cellno & \cellno & \cellyes & \cellyes & \cellno & \cellyes & \cellno & \cellno & \cellyes & \cellno \\
2023 & \sysName{MagicEdit}~\citep{liew2023magicedit} & Conditioning branch & \cellyes & \cellyes & \cellyes & \cellno & \cellno & \cellno & \cellno & \cellyes & \cellyes & \cellno & \cellyes & \cellno & \cellno & \cellyes & \cellno \\
2023 & \sysName{VideoControlNet}~\citep{hu2023videocontrolnet} & Conditioning branch & \cellno & \cellyes & \cellno & \cellno & \cellno & \cellno & \cellyes & \cellyes & \cellno & \cellno & \cellyes & \cellno & \cellno & \cellyes & \cellno \\
2023 & \sysName{VideoComposer}~\citep{wang2023videocomposer} & Conditioning branch & \cellno & \cellyes & \cellno & \cellno & \cellno & \cellyes & \cellyes & \cellyes & \cellno & \cellno & \cellyes & \cellno & \cellno & \cellyes & \cellno \\
2023 & \sysName{Structure/Content Guided Video Synthesis}~\citep{esser2023structurecontent} & Conditioning branch & \cellno & \cellyes & \cellno & \cellno & \cellno & \cellyes & \cellyes & \cellyes & \cellno & \cellno & \cellyes & \cellno & \cellno & \cellyes & \cellno \\
2024 & \sysName{Movie Gen}~\citep{polyak2024moviegen} & Training modification & \cellyes & \cellyes & \cellyes & \cellno & \cellno & \cellyes & \cellyes & \cellyes & \cellyes & \cellno & \cellyes & \cellno & \cellno & \cellyes & \cellno \\
2024 & \sysName{EffiVED}~\citep{zhang2024effived} & Training modification & \cellyes & \cellyes & \cellno & \cellno & \cellno & \cellno & \cellno & \cellyes & \cellno & \cellno & \cellyes & \cellno & \cellno & \cellyes & \cellno \\
2024 & \sysName{Customize-A-Video}~\citep{ren2024customizeavideo} & Training modification & \cellno & \cellyes & \cellno & \cellno & \cellno & \cellno & \cellyes & \cellno & \cellno & \cellno & \cellyes & \cellno & \cellno & \cellyes & \cellno \\
2024 & \sysName{VASE}~\citep{peruzzo2024vase} & Training modification & \cellyes & \cellyes & \cellyes & \cellno & \cellno & \cellno & \cellno & \cellyes & \cellno & \cellno & \cellyes & \cellno & \cellno & \cellyes & \cellno \\
2023 & \sysName{Customizing Motion}~\citep{materzynska2023customizingmotion} & Training modification & \cellno & \cellyes & \cellno & \cellno & \cellno & \cellno & \cellyes & \cellno & \cellno & \cellno & \cellyes & \cellno & \cellno & \cellyes & \cellno \\
2024 & \sysName{SAVE}~\citep{song2024save} & Training modification & \cellyes & \cellyes & \cellyes & \cellno & \cellno & \cellno & \cellno & \cellyes & \cellno & \cellno & \cellyes & \cellno & \cellno & \cellyes & \cellno \\
2024 & \sysName{VMC}~\citep{jeong2024vmc} & Training modification & \cellno & \cellyes & \cellno & \cellno & \cellno & \cellno & \cellyes & \cellno & \cellno & \cellno & \cellyes & \cellno & \cellno & \cellyes & \cellno \\
2024 & \sysName{DreamVideo}~\citep{wei2024dreamvideo} & Training modification & \cellno & \cellyes & \cellno & \cellno & \cellno & \cellyes & \cellyes & \cellno & \cellno & \cellno & \cellyes & \cellno & \cellno & \cellyes & \cellno \\
2024 & \sysName{Consistent V2V Transfer}~\citep{cheng2024consistentv2v} & Training modification & \cellno & \cellyes & \cellno & \cellno & \cellno & \cellno & \cellno & \cellyes & \cellno & \cellno & \cellyes & \cellno & \cellno & \cellyes & \cellno \\
2023 & \sysName{VIDiff}~\citep{xing2023vidiff} & Training modification & \cellyes & \cellyes & \cellno & \cellno & \cellno & \cellno & \cellno & \cellyes & \cellno & \cellno & \cellyes & \cellno & \cellno & \cellyes & \cellno \\
2024 & \sysName{MotionDirector}~\citep{zhao2024motiondirector} & Training modification & \cellno & \cellyes & \cellno & \cellno & \cellno & \cellno & \cellyes & \cellno & \cellno & \cellno & \cellyes & \cellno & \cellno & \cellyes & \cellno \\
2024 & \sysName{InstructVid2Vid}~\citep{qin2024instructvid2vid} & Training modification & \cellyes & \cellyes & \cellno & \cellno & \cellno & \cellno & \cellno & \cellyes & \cellno & \cellno & \cellyes & \cellno & \cellno & \cellyes & \cellno \\
2023 & \sysName{Dreamix}~\citep{molad2023dreamix} & Training modification & \cellyes & \cellyes & \cellno & \cellno & \cellno & \cellyes & \cellyes & \cellyes & \cellno & \cellno & \cellyes & \cellno & \cellno & \cellyes & \cellno \\
\bottomrule
\end{tabular}
\vspace{-1mm}
\end{table*}

\begin{table*}[t!]
\centering
\renewcommand{\arraystretch}{1.12}
\scriptsize
\caption{
\textbf{Video editing methods most relevant to audio-video generation pipelines: attention\slash latent\slash canonical\slash interactive families.}
This continuation covers the attention-injection, motion-feature-injection, latent-manipulation, canonical-representation, point/pose-conditioning, and human(-object)-animation families.
}
\label{tab:video-editing-taxonomy-b}
\begin{tabular}{P{6mm} >{\RaggedRight\arraybackslash}p{50mm} p{30mm} @{\hspace{8pt}} P{\cellszsm}P{\cellszsm}P{\cellszsm}P{\cellszsm}P{\cellszsm}P{\cellszsm} @{\hspace{8pt}} P{\cellszsm}P{\cellszsm}P{\cellszsm} @{\hspace{8pt}} P{\cellszsm}P{\cellszsm}P{\cellszsm}P{\cellszsm}P{\cellszsm}P{\cellszsm}}
\toprule
\textbf{Year} & \textbf{Method} & \textbf{Family} &
\rot{Instr.} & \rot{Text} & \rot{Mask/Box} & \rot{Point/Traj.} & \rot{Pose} & \rot{Image/Style} &
\rot{Motion} & \rot{Appearance} & \rot{Inpaint} &
\rot{Tuning-free} & \rot{Fine-tune} & \rot{Attention} & \rot{Latent} & \rot{Cond. Branch} & \rot{Canonical} \\
\midrule
2024 & \sysName{VideoGrain}~\citep{yang2024videograin} & Attention injection & \cellno & \cellyes & \cellno & \cellno & \cellno & \cellno & \cellyes & \cellyes & \cellno & \cellyes & \cellno & \cellyes & \cellno & \cellno & \cellno \\
2024 & \sysName{AnyV2V}~\citep{ku2024anyv2v} & Attention injection & \cellyes & \cellyes & \cellyes & \cellno & \cellno & \cellyes & \cellyes & \cellyes & \cellyes & \cellyes & \cellno & \cellyes & \cellno & \cellno & \cellno \\
2024 & \sysName{CoCoCo}~\citep{zi2024cococo} & Attention injection & \cellyes & \cellyes & \cellyes & \cellno & \cellno & \cellno & \cellno & \cellyes & \cellyes & \cellyes & \cellno & \cellyes & \cellno & \cellno & \cellno \\
2024 & \sysName{Object-Centric Diffusion}~\citep{kahatapitiya2024objectcentric} & Attention injection & \cellyes & \cellyes & \cellyes & \cellno & \cellno & \cellno & \cellno & \cellyes & \cellno & \cellyes & \cellno & \cellyes & \cellno & \cellno & \cellno \\
2024 & \sysName{UniEdit}~\citep{bai2024uniedit} & Attention injection & \cellyes & \cellyes & \cellno & \cellno & \cellno & \cellno & \cellyes & \cellyes & \cellno & \cellyes & \cellno & \cellyes & \cellno & \cellno & \cellno \\
2023 & \sysName{Make-A-Protagonist}~\citep{zhao2023makeaprotagonist} & Attention injection & \cellyes & \cellyes & \cellyes & \cellno & \cellno & \cellyes & \cellyes & \cellyes & \cellno & \cellno & \cellyes & \cellyes & \cellno & \cellno & \cellno \\
2023 & \sysName{Zero-Shot Video Editing}~\citep{wang2023zeroshotv2v} & Attention injection & \cellno & \cellyes & \cellno & \cellno & \cellno & \cellno & \cellno & \cellyes & \cellno & \cellyes & \cellno & \cellyes & \cellno & \cellno & \cellno \\
2023 & \sysName{FateZero}~\citep{qi2023fatezero} & Attention injection & \cellno & \cellyes & \cellno & \cellno & \cellno & \cellno & \cellno & \cellyes & \cellno & \cellyes & \cellno & \cellyes & \cellno & \cellno & \cellno \\
2023 & \sysName{Edit-A-Video}~\citep{shin2023editavideo} & Attention injection & \cellno & \cellyes & \cellyes & \cellno & \cellno & \cellno & \cellno & \cellyes & \cellno & \cellno & \cellyes & \cellyes & \cellno & \cellno & \cellno \\
2023 & \sysName{Video-P2P}~\citep{liu2023videop2p} & Attention injection & \cellno & \cellyes & \cellno & \cellno & \cellno & \cellno & \cellno & \cellyes & \cellno & \cellyes & \cellno & \cellyes & \cellno & \cellno & \cellno \\
2024 & \sysName{FRESCO}~\citep{yang2024fresco} & Motion feature injection & \cellno & \cellyes & \cellno & \cellno & \cellno & \cellno & \cellyes & \cellyes & \cellno & \cellyes & \cellno & \cellyes & \cellno & \cellno & \cellno \\
2024 & \sysName{FLATTEN}~\citep{cong2024flatten} & Motion feature injection & \cellno & \cellyes & \cellno & \cellno & \cellno & \cellno & \cellyes & \cellyes & \cellno & \cellyes & \cellno & \cellyes & \cellno & \cellno & \cellno \\
2024 & \sysName{TokenFlow}~\citep{geyer2024tokenflow} & Motion feature injection & \cellno & \cellyes & \cellno & \cellno & \cellno & \cellno & \cellyes & \cellyes & \cellno & \cellyes & \cellno & \cellyes & \cellno & \cellno & \cellno \\
2024 & \sysName{STEM-Inv}~\citep{li2024stem} & Latent manipulation & \cellno & \cellyes & \cellno & \cellno & \cellno & \cellno & \cellno & \cellyes & \cellno & \cellyes & \cellno & \cellno & \cellyes & \cellno & \cellno \\
2023 & \sysName{Video ControlNet}~\citep{chu2023videocontrolnet} & Latent manipulation & \cellno & \cellyes & \cellno & \cellno & \cellno & \cellno & \cellyes & \cellyes & \cellno & \cellno & \cellyes & \cellno & \cellyes & \cellno & \cellno \\
2023 & \sysName{Control-A-Video}~\citep{chen2023controlavideo} & Latent manipulation & \cellno & \cellyes & \cellno & \cellno & \cellno & \cellno & \cellyes & \cellyes & \cellno & \cellyes & \cellno & \cellno & \cellyes & \cellno & \cellno \\
2023 & \sysName{Text2Video-Zero}~\citep{khachatryan2023text2videozero} & Latent manipulation & \cellno & \cellyes & \cellno & \cellno & \cellno & \cellno & \cellyes & \cellyes & \cellno & \cellyes & \cellno & \cellno & \cellyes & \cellno & \cellno \\
2024 & \sysName{FRAG}~\citep{yoon2024frag} & Latent manipulation & \cellno & \cellyes & \cellno & \cellno & \cellno & \cellno & \cellno & \cellyes & \cellno & \cellyes & \cellno & \cellno & \cellyes & \cellno & \cellno \\
2024 & \sysName{GenVideo}~\citep{harsha2024genvideo} & Latent manipulation & \cellno & \cellyes & \cellno & \cellno & \cellno & \cellyes & \cellyes & \cellyes & \cellno & \cellno & \cellyes & \cellno & \cellyes & \cellno & \cellno \\
2024 & \sysName{MotionClone}~\citep{ling2024motionclone} & Latent manipulation & \cellno & \cellyes & \cellno & \cellno & \cellno & \cellno & \cellyes & \cellno & \cellno & \cellyes & \cellno & \cellno & \cellyes & \cellno & \cellno \\
2024 & \sysName{RAVE}~\citep{kara2024rave} & Latent manipulation & \cellno & \cellyes & \cellno & \cellno & \cellno & \cellno & \cellno & \cellyes & \cellno & \cellyes & \cellno & \cellno & \cellyes & \cellno & \cellno \\
2024 & \sysName{Space-Time Diffusion Features}~\citep{yatim2024spacetime} & Latent manipulation & \cellno & \cellyes & \cellno & \cellno & \cellno & \cellno & \cellyes & \cellno & \cellno & \cellyes & \cellno & \cellno & \cellyes & \cellno & \cellno \\
2023 & \sysName{DiffSynth}~\citep{duan2023diffsynth} & Latent manipulation & \cellno & \cellyes & \cellno & \cellno & \cellno & \cellno & \cellno & \cellyes & \cellno & \cellyes & \cellno & \cellno & \cellyes & \cellno & \cellno \\
2023 & \sysName{Rerender-A-Video}~\citep{yang2023rerender} & Latent manipulation & \cellno & \cellyes & \cellno & \cellno & \cellno & \cellno & \cellno & \cellyes & \cellno & \cellyes & \cellno & \cellno & \cellyes & \cellno & \cellno \\
2023 & \sysName{ControlVideo}~\citep{zhang2023controlvideo} & Latent manipulation & \cellno & \cellyes & \cellno & \cellno & \cellno & \cellno & \cellyes & \cellyes & \cellno & \cellyes & \cellno & \cellno & \cellyes & \cellno & \cellno \\
2023 & \sysName{Pix2Video}~\citep{ceylan2023pix2video} & Latent manipulation & \cellno & \cellyes & \cellno & \cellno & \cellno & \cellyes & \cellno & \cellyes & \cellno & \cellno & \cellyes & \cellno & \cellyes & \cellno & \cellno \\
2023 & \sysName{Neural Video Fields Editing}~\citep{yang2023nvfediting} & Canonical representation & \cellno & \cellyes & \cellno & \cellno & \cellno & \cellno & \cellno & \cellyes & \cellno & \cellno & \cellyes & \cellno & \cellno & \cellno & \cellyes \\
2023 & \sysName{DiffusionAtlas}~\citep{chang2023diffusionatlas} & Canonical representation & \cellno & \cellyes & \cellno & \cellno & \cellno & \cellno & \cellno & \cellyes & \cellno & \cellno & \cellyes & \cellno & \cellno & \cellno & \cellyes \\
2023 & \sysName{StableVideo}~\citep{chai2023stablevideo} & Canonical representation & \cellno & \cellyes & \cellno & \cellno & \cellno & \cellno & \cellno & \cellyes & \cellno & \cellno & \cellyes & \cellno & \cellno & \cellno & \cellyes \\
2024 & \sysName{CoDeF}~\citep{ouyang2024codef} & Canonical representation & \cellno & \cellyes & \cellno & \cellno & \cellno & \cellno & \cellno & \cellyes & \cellno & \cellno & \cellyes & \cellno & \cellno & \cellno & \cellyes \\
2024 & \sysName{VidEdit}~\citep{couairon2024videdit} & Canonical representation & \cellno & \cellyes & \cellyes & \cellno & \cellno & \cellno & \cellno & \cellyes & \cellno & \cellyes & \cellno & \cellno & \cellno & \cellno & \cellyes \\
2023 & \sysName{Layered Video Editing}~\citep{lee2023layered} & Canonical representation & \cellno & \cellyes & \cellyes & \cellno & \cellno & \cellno & \cellno & \cellyes & \cellno & \cellno & \cellyes & \cellno & \cellno & \cellno & \cellyes \\
2023 & \sysName{Neural Video Deflickering}~\citep{lei2023deflickering} & Canonical representation & \cellno & \cellno & \cellno & \cellno & \cellno & \cellno & \cellno & \cellyes & \cellno & \cellno & \cellyes & \cellno & \cellno & \cellno & \cellyes \\
2024 & \sysName{MotionCtrl}~\citep{wang2024motionctrl} & Point/pose conditioning & \cellno & \cellyes & \cellno & \cellyes & \cellno & \cellno & \cellyes & \cellno & \cellno & \cellno & \cellyes & \cellno & \cellno & \cellyes & \cellno \\
2023 & \sysName{Drag-A-Video}~\citep{teng2023dragavideo} & Point/pose conditioning & \cellyes & \cellyes & \cellno & \cellyes & \cellno & \cellno & \cellyes & \cellyes & \cellno & \cellno & \cellyes & \cellno & \cellno & \cellyes & \cellno \\
2024 & \sysName{DragVideo}~\citep{deng2024dragvideo} & Point/pose conditioning & \cellyes & \cellyes & \cellno & \cellyes & \cellno & \cellno & \cellyes & \cellyes & \cellno & \cellno & \cellyes & \cellno & \cellno & \cellyes & \cellno \\
2024 & \sysName{VideoSwap}~\citep{guo2024videoswap} & Point/pose conditioning & \cellyes & \cellyes & \cellyes & \cellyes & \cellno & \cellyes & \cellno & \cellyes & \cellno & \cellno & \cellyes & \cellno & \cellno & \cellyes & \cellno \\
2023 & \sysName{Animate Anyone}~\citep{hu2023animateanyone} & Human animation & \cellno & \cellno & \cellno & \cellno & \cellyes & \cellyes & \cellyes & \cellno & \cellno & \cellno & \cellyes & \cellno & \cellno & \cellyes & \cellno \\
2023 & \sysName{MagicAnimate}~\citep{xu2023magicanimate} & Human animation & \cellno & \cellno & \cellno & \cellno & \cellyes & \cellyes & \cellyes & \cellno & \cellno & \cellno & \cellyes & \cellno & \cellno & \cellyes & \cellno \\
2023 & \sysName{DreamPose}~\citep{karras2023dreampose} & Human animation & \cellno & \cellno & \cellno & \cellno & \cellyes & \cellyes & \cellyes & \cellno & \cellno & \cellno & \cellyes & \cellno & \cellno & \cellyes & \cellno \\
2024 & \sysName{DisCo}~\citep{wang2024disco} & Human animation & \cellno & \cellno & \cellno & \cellno & \cellyes & \cellyes & \cellyes & \cellno & \cellno & \cellno & \cellyes & \cellno & \cellno & \cellyes & \cellno \\
2024 & \sysName{MagicPose}~\citep{chang2024magicpose} & Human animation & \cellno & \cellno & \cellno & \cellno & \cellyes & \cellyes & \cellyes & \cellno & \cellno & \cellno & \cellyes & \cellno & \cellno & \cellyes & \cellno \\
2024 & \sysName{Fashion-VDM}~\citep{karras2024fashionvdm} & Human animation & \cellno & \cellno & \cellyes & \cellno & \cellyes & \cellyes & \cellyes & \cellyes & \cellno & \cellno & \cellyes & \cellno & \cellno & \cellyes & \cellno \\
2024 & \sysName{Animate-X}~\citep{tan2024animatex} & Human animation & \cellno & \cellno & \cellno & \cellno & \cellyes & \cellyes & \cellyes & \cellno & \cellno & \cellno & \cellyes & \cellno & \cellno & \cellyes & \cellno \\
2024 & \sysName{StableAnimator}~\citep{tu2024stableanimator} & Human animation & \cellno & \cellno & \cellno & \cellno & \cellyes & \cellyes & \cellyes & \cellno & \cellno & \cellno & \cellyes & \cellno & \cellno & \cellyes & \cellno \\
2024 & \sysName{AnchorCrafter}~\citep{xu2024anchorcrafter} & Human-object animation & \cellno & \cellyes & \cellno & \cellno & \cellyes & \cellyes & \cellyes & \cellno & \cellno & \cellno & \cellyes & \cellno & \cellno & \cellyes & \cellno \\
\bottomrule
\end{tabular}
\vspace{-1mm}
\end{table*}
\begin{table*}[t!]
\centering
\renewcommand{\arraystretch}{1.12}
\scriptsize
\caption{
\textbf{Commercial and foundation-system capabilities for video generation with audio and audio-for-video editing.}
These systems are included because they materially define current user-facing capabilities, even when model details or weights are not fully released.
}
\label{tab:commercial-av-systems}
\begin{tabular}{P{6mm} l @{\hspace{8pt}} P{\cellszmd}P{\cellszmd}P{\cellszmd}P{\cellszmd} P{\cellszmd}P{\cellszmd}P{\cellszmd} l}
\toprule
\textbf{Year} & \textbf{System} & \rot{T2AV} & \rot{V2A} & \rot{A2V} & \rot{Joint} & \rot{Video Edit} & \rot{Speech/SFX} & \rot{Music/Amb.} & \textbf{Notes} \\
\midrule
2024 & \sysName{Movie Gen}~\citep{polyak2024moviegen} & \cellyes & \cellyes & \cellno & \cellyes & \cellyes & \cellyes & \cellyes & research model; video, audio, personalization, editing \\
2024 & \sysName{Google V2A}~\citep{deepmind2024v2a} & \cellno & \cellyes & \cellno & \cellno & \cellyes & \cellyes & \cellyes & research system; video pixels + optional audio prompt \\
2025 & \sysName{Veo 3/3.1}~\citep{deepmind2025veo3} & \cellyes & \cellyes & \cellno & \cellyes & \cellyes & \cellyes & \cellyes & commercial/product; native audio \\
2025 & \sysName{Sora 2}~\citep{openai2025sora2} & \cellyes & \cellno & \cellno & \cellyes & \cellyes & \cellyes & \cellyes & commercial/product; synchronized dialogue/SFX \\
2025 & \sysName{Adobe Firefly Audio}~\citep{adobe2025fireflyaudio} & \cellno & \cellyes & \cellno & \cellno & \cellyes & \cellyes & \cellyes & commercial creative tools; sound effects, soundtrack, speech \\
2026 & \sysName{Seedance 2.0}~\citep{bytedance2026seedance2} & \cellyes & \cellno & \cellno & \cellyes & \cellyes & \cellyes & \cellyes & commercial/product; text/image/video/audio prompts \\
\bottomrule
\end{tabular}
\vspace{-1mm}
\end{table*}
}
\input{code-availability-table-arxiv}
\end{document}

%% file: preamble-arxiv.tex
\usepackage{xcolor}
\definecolor{thedarkblue}{RGB}{0,0,120} %104} % 180
\definecolor{mydarkblue}{rgb}{0,0.08,0.45} %ICML dark blue
\definecolor{darkblue}{rgb}{0,0.08,180}
\colorlet{TufteRed}{red!80!black}
\definecolor{theblue}{RGB}{0,0,180}
\colorlet{thered}{TufteRed}
      
\usepackage{microtype}
\usepackage{balance}

\usepackage{booktabs}
\usepackage{tabularx}

\usepackage{amsmath,amssymb,amsthm}

\newcommand{\eat}[1]{\ignorespaces}
\usepackage{comment}

\usepackage{tikz}
\usepackage{verbatim}
\usetikzlibrary{arrows}
\usetikzlibrary{arrows.meta}% modern arrowheads (Latex[...])
\usetikzlibrary{shapes,snakes}
\usetikzlibrary{decorations.pathmorphing} % noisy shapes
\usetikzlibrary{fit}					% fitting shapes to coordinates
\usetikzlibrary{backgrounds}	% drawing the background after the foreground

\usepackage{ragged2e}
\usepackage{multirow}
\usepackage{microtype}
\usepackage{balance}
\usepackage{setspace}

\graphicspath{{./}{./graphics/}}
\newcolumntype{H}{>{\setbox0=\hbox\bgroup}c<{\egroup}@{}}

\newcolumntype{R}[1]{>{\RaggedLeft\arraybackslash}} %p{#1}}
\newcolumntype{L}[1]{>{\RaggedRight\arraybackslash}} %p{#1}}

\AtBeginEnvironment{pmatrix}{\setlength{\arraycolsep}{2pt}}

\DeclareMathOperator{\hugeE}{\mbox{\huge\raise-0.3ex\hbox{E}}}
\DeclareMathOperator{\p}{\mathbb{P}}
\DeclareMathOperator{\hugep}{\mbox{\huge\raise-0.3ex\hbox{$\p$}}}

%% file: code-availability-table-arxiv.tex
\begin{table*}[t!]
\centering
\scriptsize
\renewcommand{\arraystretch}{1.2}
\caption{\textbf{Code and artifact availability for the methods we cover} (checked July 2026).
\cmark: official code released, repository listed; \xmark: no functional official release ---
repositories that exist but hold no code or weights (announcement placeholders, samples-only or
dataset-only repositories) count as \xmark\ and are annotated. Availability changes quickly;
each row was checked individually at the date above.}
\label{tab:code-availability}
\setlength{\tabcolsep}{6pt}
\begin{tabular}{@{}lllp{7.6cm}@{}}
\toprule
\textbf{Method} & \textbf{Family} & \textbf{Code} & \textbf{Repository / Weights} \\
\midrule
MM-Diffusion~\cite{ruan2023mmdiffusion} & Joint gen. & \cmark & \url{https://github.com/researchmm/MM-Diffusion} \\
CoDi~\cite{tang2023codi} & Joint gen. & \cmark & \url{https://github.com/microsoft/i-Code} (i-Code-V3) \\
Seeing-and-Hearing~\cite{xing2024seeing} & Joint gen. & \cmark & \url{https://github.com/yzxing87/Seeing-and-Hearing} (V2A released; other tasks pending) \\
AV-DiT~\cite{wang2024avdit} & Joint gen. & \xmark & --- \\
MM-LDM~\cite{sun2024mmldm} & Joint gen. & \xmark & placeholder repository only (no code or weights released) \\
Movie Gen~\cite{polyak2024moviegen} & Joint gen. & \xmark & --- (benchmark data only) \\
SVG~\cite{ishii2024svg} & Joint gen. & \xmark & --- \\
MMDisCo~\cite{hayakawa2025mmdisco} & Joint gen. & \cmark & \url{https://github.com/SonyResearch/MMDisCo} \\
SyncFlow~\cite{liu2024syncflow} & Joint gen. & \xmark & --- \\
JavisDiT~\cite{liu2025javisdit} & Joint gen. & \cmark & \url{https://github.com/JavisVerse/JavisDiT} \\
JavisDiT++~\cite{liu2026javisditpp} & Joint gen. & \cmark & \url{https://github.com/JavisVerse/JavisDiT} (shared repository) \\
BridgeDiT~\cite{guan2025bridgedit} & Joint gen. & \cmark & \url{https://github.com/guankaisi/BridgeDiT} \\
ALIVE~\cite{guo2026alive} & Joint gen. & \xmark & placeholder repository only (no code or weights released) \\
Ovi~\cite{low2025ovi} & Joint gen. & \cmark & \url{https://github.com/character-ai/Ovi} (inference and weights) \\
UniAVGen~\cite{zhang2025uniavgen} & Joint gen. & \cmark & \url{https://github.com/MCG-NJU/Sora2-mini} \\
Animate-and-Sound~\cite{wang2025jointdit} & Joint gen. & \xmark & --- \\
CCL~\cite{ma2026ccl} & Joint gen. & \xmark & --- \\
Hallo-Live~\cite{li2026hallolive} & Joint gen. & \cmark & \url{https://github.com/fudan-generative-vision/Hallo-Live} \\
UniForm~\cite{zhao2025uniform} & Joint gen. & \xmark & --- \\
Wan 2.5~\cite{wan25_2025} & Joint gen. & \xmark & API-only; no public weights (Wan 2.1/2.2 repositories exclude 2.5) \\
LTX-2~\cite{hacohen2026ltx2} & Joint gen. & \cmark & \url{https://github.com/Lightricks/LTX-2} \\
MOVA~\cite{mova2026} & Joint gen. & \cmark & \url{https://github.com/OpenMOSS/MOVA} \\
Apollo~\cite{wang2026klear} & Joint gen. & \xmark & --- \\
3MDiT~\cite{li2025threemdit} & Joint gen. & \xmark & --- \\
OmniForcing~\cite{su2026omniforcing} & Joint gen. & \cmark & \url{https://github.com/OmniForcing/OmniForcing} \\
\midrule
Diff-Foley~\cite{luo2023difffoley} & Cross-modal & \cmark & \url{https://github.com/luosiallen/Diff-Foley} \\
Foley Analogies~\cite{iyer2023foleyanalogies} & Cross-modal & \cmark & \url{https://github.com/XYPB/CondFoleyGen} \\
V2A-Mapper~\cite{lin2024v2amapper} & Cross-modal & \xmark & samples-only repository (no model code) \\
Video-Foley~\cite{lee2024videofoley} & Cross-modal & \cmark & \url{https://github.com/jnwnlee/video-foley} \\
FoleyCrafter~\cite{zhang2024foleycrafter} & Cross-modal & \cmark & \url{https://github.com/open-mmlab/FoleyCrafter} \\
Frieren~\cite{wang2024frieren} & Cross-modal & \cmark & \url{https://github.com/cyanbx/Frieren-V2A} \\
MaskVAT~\cite{garoufis2024maskvat} & Cross-modal & \xmark & --- \\
STA-V2A~\cite{chen2024stav2a} & Cross-modal & \cmark & \url{https://github.com/PolyPerceiver-Lab/STAV2A} \\
VATT~\cite{liu2024vatt} & Cross-modal & \cmark & \url{https://github.com/DragonLiu1995/video-to-audio-through-text} \\
Google V2A~\cite{deepmind2024v2a} & Cross-modal & \xmark & --- (closed research system) \\
MMAudio~\cite{cheng2025mmaudio} & Cross-modal & \cmark & \url{https://github.com/hkchengrex/MMAudio} \\
Mel-QCD~\cite{wang2025melqcd} & Cross-modal & \cmark & \url{https://github.com/wjc2830/MelQCD-main} (training code pending) \\
VAFlow~\cite{wang2025vaflow} & Cross-modal & \xmark & --- \\
Foley-Flow~\cite{mo2025foleyflow} & Cross-modal & \xmark & --- \\
MultiFoley~\cite{chen2025multifoley} & Cross-modal & \xmark & --- \\
TARO~\cite{zhang2025taro} & Cross-modal & \cmark & \url{https://github.com/triton99/TARO} \\
ThinkSound~\cite{liu2025thinksound} & Cross-modal & \cmark & \url{https://github.com/FunAudioLLM/ThinkSound} \\
Hear-Your-Click~\cite{guo2025hearyourclick} & Cross-modal & \cmark & \url{https://github.com/SynapGrid/Hear-Your-Click} \\
SelVA~\cite{anonymous2025selva} & Cross-modal & \cmark & \url{https://github.com/jnwnlee/selva} (training code pending) \\
SoundReactor~\cite{saito2025soundreactor} & Cross-modal & \xmark & main model unreleased (official eval toolkit and VAE repositories only) \\
Foley-Omni~\cite{tao2026foleyomni} & Cross-modal & \cmark & \url{https://github.com/NJU-Speech/Foley-Omni} \\
AV-Link~\cite{hajiali2025avlink} & Cross-modal & \xmark & repository is a project-page template (no model code) \\
\midrule
Lang.-Guided AV Edit~\cite{liang2024avedit} & Editing & \xmark & dataset-only repository (no implementation code) \\
AvED~\cite{lin2026aved} & Editing & \cmark & \url{https://github.com/GenjiB/AVED} \\
EdiDub~\cite{manela2025edidub} & Editing & \xmark & results-only repository (no model code) \\
Object-AVEdit~\cite{fu2025objectavedit} & Editing & \xmark & --- (project page only) \\
AV-Edit~\cite{guo2026avedit} & Editing & \xmark & placeholder repository only (no code or weights released) \\
JUST-DUB-IT~\cite{chen2026justdubit} & Editing & \cmark & \url{https://github.com/justdubit/just-dub-it} (archived; folded into LTX-2 pipeline) \\
EditYourself~\cite{flynn2026edityourself} & Editing & \xmark & --- \\
\bottomrule
\end{tabular}
\end{table*}